\documentclass{article}

\PassOptionsToPackage{numbers, compress}{natbib}
\usepackage[final]{neurips_2025}

\usepackage[utf8]{inputenc} 
\usepackage[T1]{fontenc}    
\usepackage{hyperref}       
\usepackage{url}            
\usepackage{booktabs}       
\usepackage{amsfonts}       
\usepackage{nicefrac}       
\usepackage{microtype}      
\usepackage{xcolor}         
\usepackage{xspace}
\usepackage{graphicx}
\usepackage{colortbl}
\usepackage{tcolorbox}
\usepackage{amsthm}
\usepackage{amsmath}
\usepackage{multirow}
\usepackage{makecell}
\usepackage{bbding}
\usepackage{pifont}
\usepackage{subfigure}
\usepackage{subcaption}
\definecolor{sectionblue}{HTML}{E0E8FF}
\definecolor{lightyellow}{RGB}{255, 255, 204}
\newcommand{\datasetname}{UVE-Bench\xspace}

\newcommand*{\affaddr}[1]{#1}
\newcommand*{\affmark}[1][*]{\textsuperscript{#1}}

\definecolor{my_green}{RGB}{51,102,0}
\definecolor{my_red}{RGB}{204, 0, 0}
\renewcommand{\checkmark}{\textcolor{my_green}{\ding{51}}} 
\newcommand{\crossmark}{\textcolor{my_red}{\ding{55}}} 

\newcommand{\huggingface}{\raisebox{-1.5pt}{\includegraphics[height=1.05em]{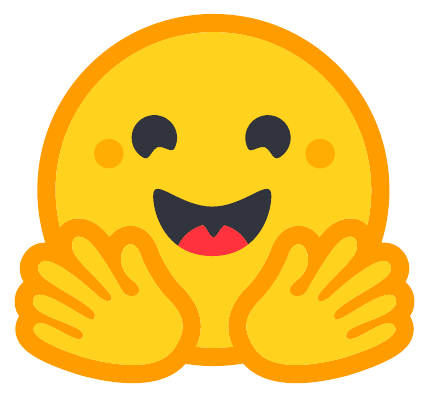}}\xspace}
\newcommand{\github}{\raisebox{-1.5pt}{\includegraphics[height=1.05em]{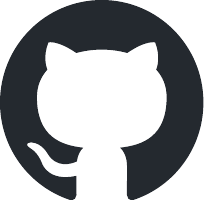}}\xspace}

\title{UVE: Are MLLMs Unified Evaluators for AI-Generated Videos?}

\author{
  Yuanxin Liu\affmark[$\S$] \quad
  Rui Zhu\affmark[\ddag] \quad
  Shuhuai Ren\affmark[$\S$] \quad
  Jiacong Wang\affmark[\P] \quad \\
  \textbf{Haoyuan Guo}\affmark[\ddag] \quad
  \textbf{Xu Sun}\affmark[$\S$] \quad
  \textbf{Lu Jiang}\affmark[\ddag]  \\
  \affaddr{\affmark[\S] State Key Laboratory of Multimedia Information Processing,\\
      School of Computer Science, Peking University} \\
  \affaddr{\affmark[\ddag] ByteDance Seed} \\
  \affaddr{\affmark[\P] School of Artificial Intelligence, University of Chinese Academy of Sciences} \\
  \texttt{liuyuanxin@stu.pku.edu.cn} \quad \\ \texttt{guohaoyuan@bytedance.com} \quad \texttt{xusun@pku.edu.cn} \\
}

\begin{document}

\maketitle

\begin{abstract}
With the rapid growth of video generative models (VGMs), it is essential to develop reliable and comprehensive automatic metrics for AI-generated videos (AIGVs). Existing methods either use off-the-shelf models optimized for other tasks or rely on human assessment data to train specialized evaluators. These approaches are constrained to specific evaluation aspects and are difficult to scale with the increasing demands for finer-grained and more comprehensive evaluations. To address this issue, this work investigates the feasibility of using multimodal large language models (MLLMs) as a unified evaluator for AIGVs, leveraging their strong visual perception and language understanding capabilities. To evaluate the performance of automatic metrics in unified AIGV evaluation, we introduce a benchmark called \datasetname. \datasetname collects videos generated by state-of-the-art VGMs and provides pairwise human preference annotations across 15 evaluation aspects. Using \datasetname, we extensively evaluate 18 MLLMs. Our empirical results suggest that while advanced MLLMs (e.g., Qwen2VL-72B and InternVL2.5-78B) still lag behind human evaluators, they demonstrate promising ability in unified AIGV evaluation, significantly surpassing existing specialized evaluation methods. Additionally, we conduct an in-depth analysis of key design choices that impact the performance of MLLM-driven evaluators, offering valuable insights for future research on AIGV evaluation. 
\github \textbf{Code}: \url{https://github.com/bytedance/UVE}, \huggingface \textbf{Data}: \url{https://huggingface.co/datasets/lyx97/UVE-Bench}.
\end{abstract}

\section{Introduction}
\label{sec:intro}
Video generative models (VGMs) have rapidly evolved in recent years, greatly aiding visual content creation in numerous fields. However, even current state-of-the-art (SOTA) VGMs~\cite{sora,moviegen,hunyuanvideo,kling} still suffer from issues like incorrect subject structure, unnatural motion, and imperfect alignment with the text prompt. Consequently, it is crucial to develop automatic metrics that can effectively identify these imperfections in modern AI-generated videos (AIGVs), so as to facilitate the advancement of VGMs and enhance the application of AIGVs in real-world scenarios.

Existing AIGV evaluation metrics fall into two categories. The first employs off-the-shelf models (e.g., CLIP~\cite{CLIP} and DINO~\cite{DINO}) and designs heuristic rules to assess various specific aspects of AIGVs~\cite{vbench,evalcrafter}. The second category involves collecting human assessments for specific aspects of interests and training models to imitate these assessments~\cite{videoscore,videophy,ugvq,q-align,AIGV-Assessor}. Such specifically trained automatic evaluators have shown better correlation with human evaluations compared to the off-the-shelf models. However, with the rapid development of VGMs and AIGV applications, there is an urgent need for finer-grained and more comprehensive evaluations. Continuously collecting human assessments and training models to accommodate emerging evaluation aspects and frequently changing standards is cost-intensive and difficult to scale.

Unlike existing automatic metrics, we humans can assess any aspect of AIGVs as long as a proper guideline is provided. This ability stems from our robust visual perception and language understanding. Leveraging the knowledge learned from vast amounts of visual and language data, the latest multimodal large language models (MLLMs)~\cite{qwen2vl,GPT4,internvl2.5} also exhibit strong ability in joint vision-language understanding. This progress naturally raises a question: ``\textbf{Can MLLMs be utilized as a unified AIGV evaluator like humans?}''

To address this question, we propose an approach to unify AIGV evaluations by prompting pre-trained MLLMs and mapping their outputs to evaluation results (\S \ref{sec:uve_framework}). This method enables zero-shot evaluation of any aspect of AIGV by simply modifying the prompts. It supports both single video ratings and video pair comparisons. While using MLLMs for AIGV evaluation is a straightforward concept and has been explored in previous works~\cite{videoscore,videophy,AIGV-Assessor}, these studies are limited by evaluating only a few aspects and relying on human-annotated ratings to train the MLLMs.

To evaluate the capability of MLLMs as unified AIGV evaluators, we require a benchmark that (1) encompasses a broad range of AIGV aspects, (2) provides accurate human evaluations as references, and (3) includes AIGVs that highlight the weaknesses of state-of-the-art VGMs. Since no existing AIGV dataset meets all these criteria, we introduce a new benchmark called \textbf{\datasetname} (short for \textbf{U}nified \textbf{V}ideo \textbf{E}valuation; see \S \ref{sec:uve_bench}). \datasetname has three key features: \textbf{First}, it covers a wide range of 15 evaluation aspects. \textbf{Second}, it provides human annotation in the form of pairwise video preference, avoiding the inconsistent standards of absolute ratings between humans~\cite{co-instruct}, and can be used to evaluate both single video ratings and video pair comparisons. \textbf{Third}, the videos in \datasetname are generated by the latest VGMs (e.g., MovieGenVideo~\cite{moviegen} and HunyuanVideo~\cite{hunyuanvideo}), challenging the evaluators to identify the weaknesses of such videos.

Based on \datasetname, we conduct an extensive evaluation of 18 MLLMs. Our results indicate that unified evaluator powered by advanced MLLMs, such as Qwen2-VL-72B~\cite{qwen2vl} and InternVL2.5-78B~\cite{internvl2.5}, indeed show promising abilities in evaluating various AIGV aspects, outperforming existing approaches that focus on specific aspects. However, there remains a significant gap between these MLLMs and human evaluators, particularly in aspects requiring a fine-grained understanding of temporal dynamics in videos. Additionally, we perform a series of analytical studies on the design choices that impact the performance of our MLLM-driven AIGV evaluation framework, providing valuable insights for future research in the field.

The contributions of this work are: \textbf{(1)} We introduce a unified approach to evaluate any aspect of AIGV using pre-trained MLLMs. \textbf{(2)} We propose \datasetname, a comprehensive benchmark to assess the capability of unified AIGV evaluation. \textbf{(3)} We conduct in-depth analysis on the pros and cons of MLLMs in unified AIGV evaluation and the key design choices that impact their performance.

\section{Unifying AIGV Evaluation with MLLMs}
\label{sec:uve_framework}
As shown in Fig.~\ref{fig:uve_framework}, the unified evaluator is designed to handle all aspects of AIGV evaluation, eliminating the need for specialized models tailored to individual aspects. In this section, we describe how this framework can be applied to both single video rating and video pair comparison.

\subsection{Problem Formulation}
\paragraph{Single Video Rating.} Given a generated video $\mathcal{V}$, a textual evaluation guideline $\mathcal{G}_{a}$ and the text-to-video (T2V) prompt $\mathcal{T}$ used to generate the video, the objective of single video rating is to predict a numerical score $\mathcal{S}$ measuring the quality of $\mathcal{V}$ based on the aspect $a$ specified by $\mathcal{G}_{a}$. 

\paragraph{Video Pair Comparison.} Given a pair of generated videos $\mathcal{V}_1, \mathcal{V}_2$, the corresponding T2V prompts $\mathcal{T}_1, \mathcal{T}_2$ used to generate them, and a textual evaluation guideline $\mathcal{G}_a$, the objective of video pair comparison is to make a choice $\mathcal{C}$ from one of the four options: $\mathbf{O}=\{``\mathcal{V}_1 \text{better}", ``\mathcal{V}_2 \text{better}", ``\text{same good}", ``\text{same bad}"\}$, according to the aspect $a$ described by $\mathcal{G}_a$.

\begin{minipage}{0.55\textwidth}
    \centering
    \includegraphics[width=\textwidth]{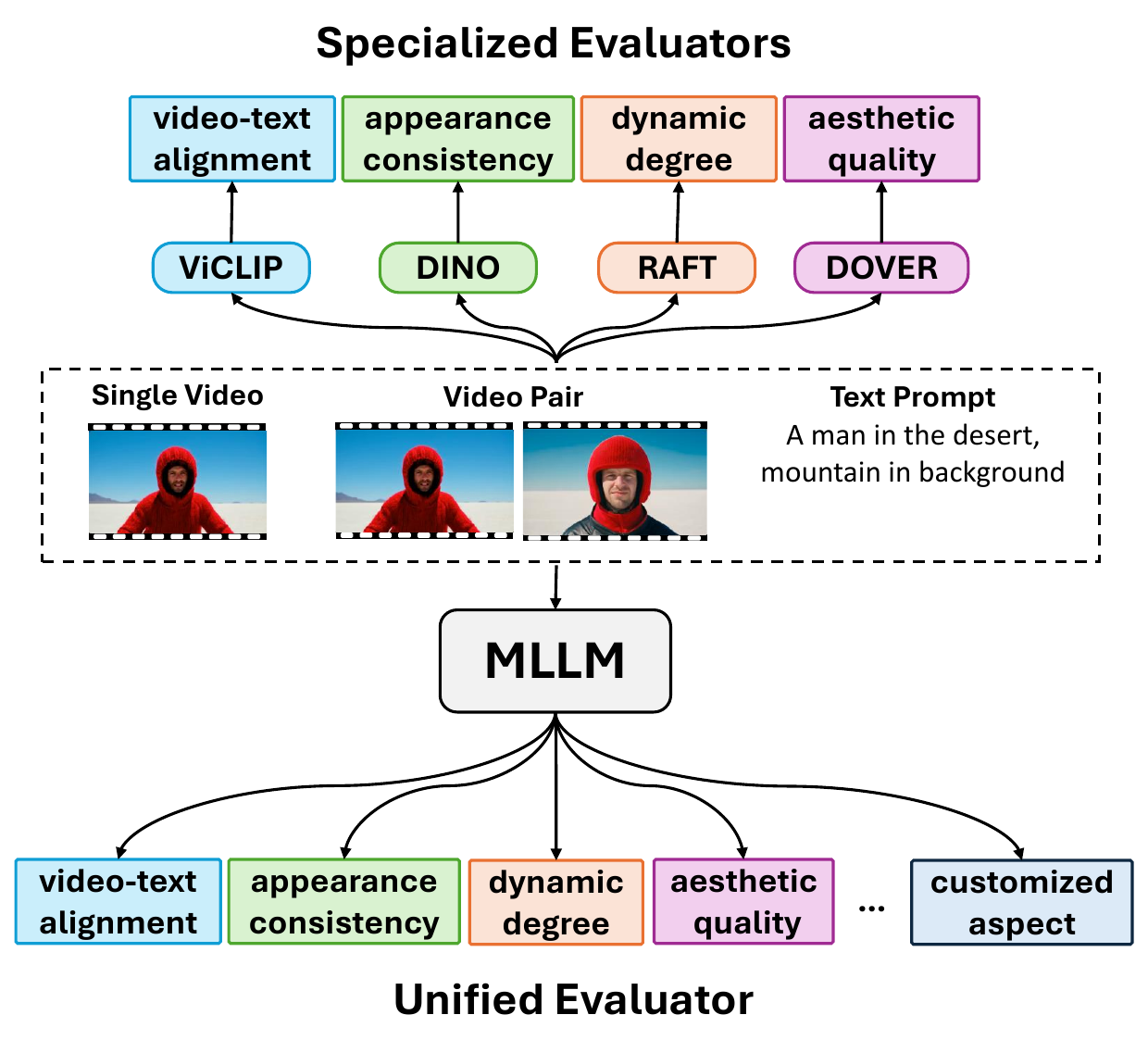}
    \captionof{figure}{Illustration of MLLM-based unified evaluator and specialized evaluators.}
    \label{fig:uve_framework}
\end{minipage}
\hfill
\begin{minipage}{0.44\textwidth}
        \centering
    \captionof{table}{MLLM prompting templates for unified AIGV evaluation.}
    \resizebox{0.8\linewidth}{!}{
    \begin{tcolorbox}  
        \textbf{Single Video Rating} \\
        \textless video\textgreater \\
        Watch the above frames of an AI-generated video and evaluate {\color[HTML]{00CC00} \textless aspect-specific description\textgreater}\\
        
        Complete your evaluation by answering this question:\\
        {\color[HTML]{CC0000} \textless aspect-specific question\textgreater}?\\
        {\color[HTML]{0000CC} \textless answer prompt\textgreater} \\
        \\
        \textbf{Video Pair Comparison}\\
        The first video: \textless video\textgreater \\
        The second video: \textless video\textgreater \\
        Watch the above two AI-generated videos and evaluate {\color[HTML]{00CC00} \textless aspect-specific description\textgreater}\\

        Complete your evaluation by answering this question:\\
        Which video is {\color[HTML]{CC0000} \textless aspect-specific question\textgreater}?\\

        You should make your judgment based on the following rules:\\
        {\color[HTML]{0000CC} \textless instructions on how to make the choice\textgreater}\\
        Now give your judgment:
    \end{tcolorbox}
    }
    \label{tab:prompt_template}
\end{minipage}

\subsection{Method}
\paragraph{Single Video Rating.} Research on image quality evaluation \cite{qbench,vqascore} has demonstrated that, given an appropriate prompt, the generative likelihood of certain tokens (e..g, \textit{yes/no} or \textit{good/poor}) by MLLMs can be used an effective indicator of image visual quality and image-text alignment. Motivated by this, we formulate single video rating in a unified manner as:
\begin{equation}
\label{eq:single_video_rating}
\mathcal{S} = \frac{P_{\mathbf{\theta}}( t_{\text{pos}} | \mathcal{V}, \mathcal{T}, \mathcal{G}_{a})}{P_{\mathbf{\theta}}( t_{\text{pos}} | \mathcal{V}, \mathcal{T}, \mathcal{G}_{a}) + P_{\mathbf{\theta}}( t_{\text{neg}} | \mathcal{V}, \mathcal{T}, \mathcal{G}_{a})}
\end{equation}
where $t_{\text{pos}}$ and $t_{\text{neg}}$ are predefined positive/negative scoring tokens. $P_{\mathbf{\theta}}(t)$ is the probability of generating token $t$ using MLLM $\mathbf{\theta}$. $\mathcal{G}_a$ ends with a question related to aspect $a$ and an answer prompt encouraging the model to generate the scoring tokens (e.g., \textit{Please directly answer yes or no:}).

\paragraph{Video Pair Comparison.} A straightforward way to perform video pair comparison is directly feeding the video pair into MLLMs, as the latest MLLMs \cite{qwen2vl,llavaonevision,internvl2.5} already support interleaved vision-language input with multiple videos. Specifically, we feed the video pair, along with the corresponding T2V prompts if needed, into the MLLM and prompts it to make a choice from $\mathbf{O}$:
\begin{equation}
\label{eq:eq_pair_compare}
\mathcal{C} = f_{\mathbf{\theta}}(\mathcal{V}_1, \mathcal{V}_2, \mathcal{T}_1, \mathcal{T}_2, \mathcal{G}_a) \in \mathbf{O}
\end{equation}
Tab.~\ref{tab:prompt_template} illustrates the prompting templates for single video rating and video pair comparison, which can be adapted to any evaluation aspect by modifying the aspect-specific prompt. The specific prompts being used for different aspects are provided in the supplementary material.

\begin{figure*}[t]
\centering
\includegraphics[width=1.\textwidth]{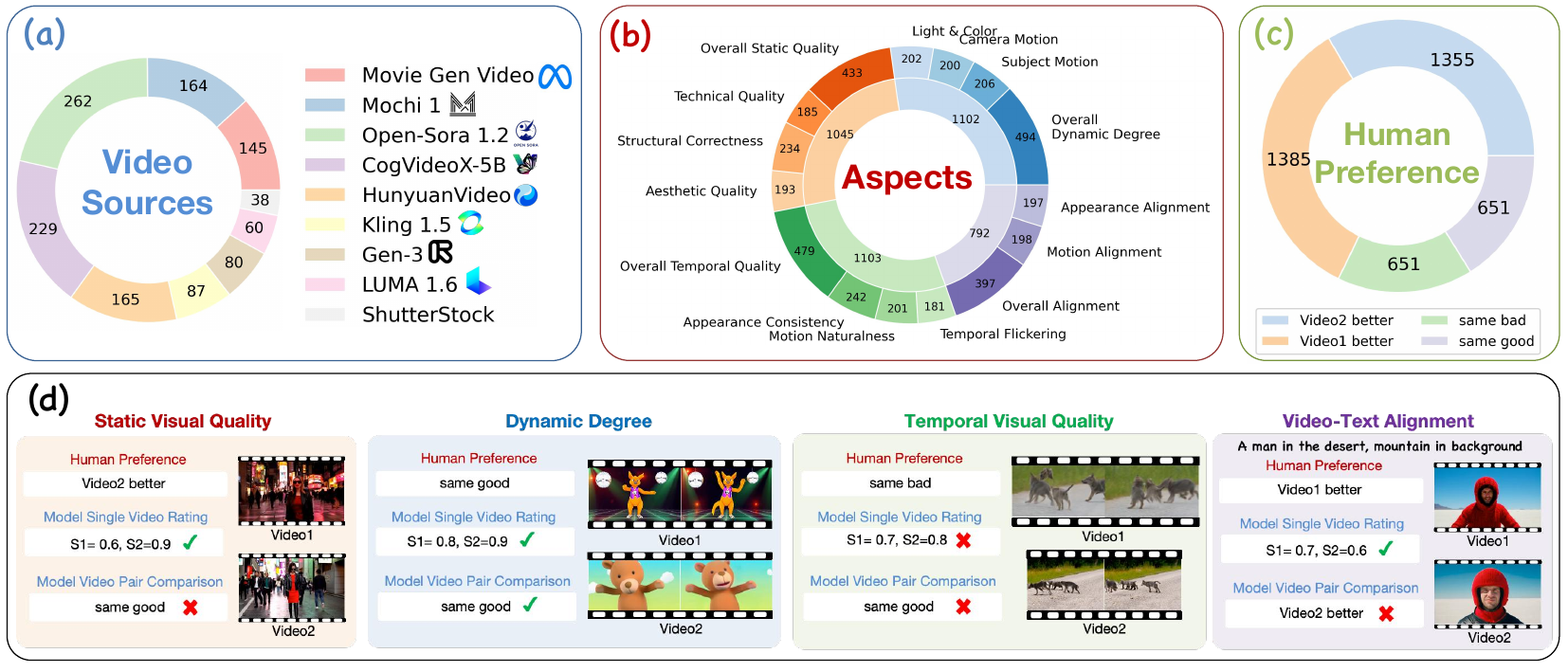}
\caption{Overview of \datasetname. (a) The distribution of video sources. (b) The distribution of data example over 15 fine-grained AIGV evaluation aspects. (c) The distribution of human preference over the four categories. (d) Data examples illustrating how to evaluate both single video rating and video pair comparison using the human preference annotations. More examples can be found in Appendix \ref{app:case_study}.}
\label{fig:overview}
\end{figure*}

\section{\datasetname}
\label{sec:uve_bench}
As illustrated in Fig.~\ref{fig:overview}, \datasetname is designed to assess AIGV automatic evaluation methods. This section details the process of video collection (\S\ref{sec:video_collect}), the evaluation aspects covered by \datasetname (\S\ref{sec:eval_aspects}), the criterion for assessing the performance of automatic AIGV evaluators using human preferences (\S\ref{sec:eval_score}), the data annotation procedure (\S\ref{sec:annotation}) and a comparison with existing AIGV datasets (\S\ref{sec:compare_with_other_dataset}).

\begin{table}[t]
    \centering
    \caption{Information of the videos in UVE-Bench.}
    \resizebox{0.7\textwidth}{!}{$
    \begin{tabular}{lcccccc}
         \toprule
         \multirow{2}{*}{\makecell{Video\\Source}} & \multirow{2}{*}{\makecell{Prompt\\Source}} & \multirow{2}{*}{Resolution} & \multirow{2}{*}{\makecell{Duration}} & \multirow{2}{*}{\makecell{FPS}} & \multirow{2}{*}{\makecell{Release\\Date}} & \multirow{2}{*}{\makecell{Open\\Source}} \\ \\
         \midrule
         Gen-3 ~\cite{gen3}            & VideoGen-Eval & 1280$\times$768  & 5.3   & 24 & 2024.06 & \crossmark  \\
         Kling 1.5 ~\cite{kling}       & VideoGen-Eval & 1920$\times$1080 & 5.1   & 30 & 2024.09 & \crossmark   \\
         LUMA 1.6 ~\cite{luma1.6}        & VideoGen-Eval & 1360$\times$752  & 5.0  & 24 &2024.09 & \crossmark   \\
         Movie Gen Video ~\cite{moviegen}  & Movie Gen & 848$\times$480   & 5.0  & 24 & 2024.10 & \crossmark   \\
         Open-Sora 1.2 ~\cite{opensora}    & \multirow{3}{*}{\makecell{VideoGen-Eval\\Movie Gen\\ShutterStock}} & \multirow{2}{*}{\makecell{1280$\times$720\\848$\times$480}} & \multirow{2}{*}{\makecell{4.3\\5.3\\}}   & 24 & 2024.06 & \checkmark  \\ \\ \\
         CogVideoX-5B~\cite{yang2024cogvideox}     & \multirow{3}{*}{\makecell{VideoGen-Eval\\Movie Gen\\ShutterStock}} & 720$\times$480   & 6.1   & 8  & 2024.08 & \checkmark  \\ \\ \\
         HunyuanVideo~\cite{hunyuanvideo}     & \multirow{2}{*}{\makecell{Movie Gen\\ShutterStock}} & 960$\times$544   & 5.4   & 24  & 2024.12& \checkmark \\ \\
         Mochi 1~\cite{genmo2024mochi}          & \multirow{2}{*}{\makecell{Movie Gen\\ShutterStock}} & 848$\times$480    & 5.4   & 30  & 2024.10& \checkmark  \\ \\
         ShutterStock     & ShutterStock & \multirow{3}{*}{\makecell{600$\times$316\\898$\times$506\\596$\times$336}} & 5.0 & \multirow{2}{*}{\makecell{24, 25\\30, 60}} & - & -  \\ \\ \\
         \bottomrule
    \end{tabular}
    $}
    \label{tab:video_generative_models}
\end{table}
\subsection{Video Collection}
\label{sec:video_collect}
The videos in \datasetname are created using eight of the latest video generative models via text-to-video generation. To ensure diverse video content, we select text prompts from three sources: (1) VideoGen-Eval \cite{videogen-eval} provides over 700 prompts, targeting various application scenarios (e.g., animation and advertisement) and key capabilities (e.g., text alignment and motion diversity) of video generation. (2) Movie Gen Video Bench \cite{moviegen} offers 1,003 prompts covering various categories of video content (e.g., human activity, animals, and scenery). (3) We also modify a small portion of video captions from the ShutterStock\footnote{\url{https://www.shutterstock.com/}} platform to generate videos with different degrees of \textit{Light \& Color} change (an evaluation aspect that will be discussed in \S\ref{sec:eval_aspects}).

For VideoGen-Eval, we directly collect videos generated by Gen-3, Kling 1.5, LUMA 1.6, Open-Sora 1.2, and CogVideoX-5B, as released by the authors of \citet{videogen-eval}. For Movie Gen Video Bench and ShutterStock, we generate videos using Open-Sora 1.2, CogVideoX-5B, Mochi 1 and HunyuanVideo. For the Movie Gen Video model, we directly collect the generated videos released by \citet{moviegen} and select the first five seconds of the videos. Additionally, we collect some non-AIGV videos related to \textit{Light \& Color} from ShutterStock. As shown in Tab.~\ref{tab:video_generative_models}, the videos have a duration of around 5.0 seconds, with resolution ranging from $596\times336$ (360p level) to $1280\times720$ (720p level) and $1920\times1080$ (1080p level).

\subsection{Evaluation Aspects}
\label{sec:eval_aspects}
Inspired by previous work in AIGV evaluation \cite{vbench,fetv,evalcrafter}, we identify four fundamental evaluation aspects to build our benchmark: \textit{Static Quality}, \textit{Temporal Quality}, \textit{Dynamic Degree} and \textit{Video-Text Alignment}, each of them is further divided into several subaspects.

\paragraph{Static Quality.} This aspect evaluates the visual quality of individual video frames. It is broken down into four subaspects: (1) \textit{Aesthetic Quality}, which assesses the aesthetic elements, including frame layout, lighting, and color harmony. (2) \textit{Technical Quality}, which examines the presence of unwanted noise, blur, and distortion. (3) \textit{Structural Correctness}, which checks for abnormal subject structures that contradict common sense (e.g., a person with three eyes). (4) \textit{Overall Static Quality}, which jointly considers the above three subaspects.

\paragraph{Temporal Quality.} This aspect assesses visual quality from a temporal perspective, focusing on four subaspects: (1) \textit{Appearance Consistency}, which evaluates whether the appearance and identity of subjects or backgrounds remain consistent across frames. (2) \textit{Temporal Flickering}, which identifies undesirable flickering and jittering that degrade visual quality. (3) \textit{Motion Naturalness}, which determines if subject motions and interactions appear natural and adhere to physical laws. (4) \textit{Overall Temporal Quality}, which comprehensively assesses the temporal visual quality by integrating these three subaspects.

\paragraph{Dynamic Degree.} This aspect evaluates the degree of dynamic in the generated video. It includes four subaspects: (1) \textit{Subject Motion}, which focuses on the motion degree of subjects. (2) \textit{Camera Motion}, which assesses the motion degree of the camera. (3) \textit{Light \& Color}, which considers changes in lighting conditions and color. (4) \textit{Overall Dynamic Degree}, which combines these three subaspects to provide a comprehensive measure of the video's dynamic degree.

\paragraph{Video-Text Alignment.} This aspect evaluates how well the generated video aligns with the given text prompt, divided into three subaspects: (1) \textit{Appearance Alignment}, which focuses on the alignment of subject or scene appearance with the text. (2) \textit{Motion Alignment}, which assesses the alignment of subject or camera motions with the text. (3) \textit{Overall Alignment}, which considers how faithfully the video content reflects the entire text prompt.

\subsection{Evaluation Criterion}
\label{sec:eval_score}
As shown in Fig.~\ref{fig:overview}, each example in UVE-Bench consists of three elements: a pair of videos $\{\mathcal{V}_1, \mathcal{V}_2\}$, the evaluation aspect $a$ and corresponding human preference annotation $\mathcal{P} \in \mathbf{O}=\{``\mathcal{V}_1 \text{better}", ``\mathcal{V}_2 \text{better}", ``\text{same good}", ``\text{same bad}"\}$. 

For single video rating, the automatic evaluator is required to predict numerical rating scores $\mathcal{S}_1,\mathcal{S}_2\in[0,1]$ for each video, respectively. Then, we assess the correctness of rating according to the following criteria:
\begin{equation}
\label{eq:eq_criteria}
\mathcal{A}^{\text{single}} = 
\begin{cases} 
\mathbf{1}(\mathcal{S}_1>\mathcal{S}_2) & \text{if } \mathcal{P}=``\mathcal{V}_1 \text{better}" \\
\mathbf{1}(\mathcal{S}_1<\mathcal{S}_2) & \text{elif } \mathcal{P}=``\mathcal{V}_2 \text{better}" \\
f_c(\mathcal{S}_1|\beta)\cdot f_c(\mathcal{S}_2|\beta) & \text{elif } \mathcal{P}=``\text{same good}" \\
f^{'}_c(\mathcal{S}_1|\alpha)\cdot f_c(\mathcal{S}_2|\alpha) & \text{elif } \mathcal{P}=``\text{same bad}" \\
\end{cases}
\end{equation}
Here, $\mathbf{1}(\cdot)\in \{0,1\}$ is a binary indicator function and $\alpha<\beta$ are the thresholds for ``bad'' and ``good'', respectively. $f_c(\mathcal{S}|\beta)$ is a piecewise function that equals to 1 when $\mathcal{S}\in[\beta,1]$, indicating that the model evaluation aligns with the human judgment of ``same good''. When $\mathcal{S} \in [0,\beta)$, $f_c(\mathcal{S}|\beta)$ exponentially decays from 1 to 0, indicating a gradual deviation from ``same good''. Similarly, $f^{'}_c(\mathcal{S}|\alpha)$  is constructed in a reversed manner to assess whether $\mathcal{S}$ agrees with the human judgment of ``same bad''. Details of $f_c(\mathcal{S}|\alpha)$ and $f^{'}_c(\mathcal{S}|\beta)$ can be found in Appendix \ref{app:eval_criteria}.

When it comes to video pair comparison, the automatic evaluator is asked to make a choice $\mathcal{C}$ from $\mathbf{O}$ in the same way as humans. Therefore, we directly adopt accuracy as the evaluation criteria: $\mathcal{A}^{pair}=\mathbf{1}(\mathcal{C}=\mathcal{P})$.

\subsection{Data Annotation and Quality Review}
\label{sec:annotation}
With the collected videos, the authors of this paper, who have rich research experience in VGMs and MLLMs and are proficient in English, annotate the evaluation aspects and pairwise video preferences. Specifically, videos generated by the same text prompt are presented in a pairwise manner to the annotators, who are asked to determine a preference choice from $\mathbf{O}$ and identify the aspect that influenced their decision. To minimize subjectivity, annotators are instructed to only annotate video pairs for which they could confidently make a preference judgment, and the remaining video pairs are discarded. Following this procedure, we obtained 4,042 pairwise preference annotations and retained 1,230 individual videos. The distributions of the videos and preference annotations are illustrated in Fig.~\ref{fig:overview}.

To verify the quality of annotations, we randomly sample a subset of the dataset, consisting of 50 samples for each subaspect and assign a different group of three annotators to label pairwise video preferences. Including the original annotations, the inter-annotator agreement, as measured by Fleiss' Kappa, is 0.803. This indicates a high level of agreement among the human annotators.

\begin{table}[t]
    \centering
    \caption{Comparison between UVE-Bench and existing AIGV datasets. ``Single'' denotes single video rating. ``Pair'' denotes video pair comparison. ``Understanding'' denotes general video quality understanding via open-ended question answering.}
    \resizebox{0.7\textwidth}{!}{$
    \begin{tabular}{lccc}
         \toprule
         Dataset & Aspects & Eval Task &  Latest VGM \\
         \midrule
         FETV \cite{fetv}               & 9  & Single  & ZeroScope (2023.06)   \\
         VBench \cite{vbench}             & 16 & Pair    & VideoCrafter1 (2023.10)   \\
         EvalCrafter \cite{evalcrafter}        & 4  & Single  & Gen-2 (2023.12)   \\
         T2VQA-DB \cite{kou2024subjective}           & 1  & Single  & AnimateDiff (2023.12)   \\
         VideoFeedback \cite{videoscore}      & 5  & Single  & SORA (2024.02)   \\
         LGVQ   \cite{ugvq}            & 3  & Single  & Gen-2 (2023.12)   \\
         TVGE \cite{tvge}   & 2  & Single  & Gen-2 (2023.12) \\
         GAIA   \cite{gaia}             & 3  & Single  & Mora (2024.03) \\
         Q-Bench-Video \cite{Q-Bench-Video}       & 4  & Understanding   & SORA (2024.02) \\
         AIGV-Assessor \cite{AIGV-Assessor}      & 4  & Single\&Pair    & SORA (2024.02) \\
         UVE-Bench (Ours)       & 15 & Single\&Pair    & HunyuanVideo (2024.12)   \\
         \bottomrule
    \end{tabular}
    $}
    \label{tab:aigv_datasets}
\end{table}
\subsection{Comparison with Existing AIGV Datasets}
\label{sec:compare_with_other_dataset}
Tab.~\ref{tab:aigv_datasets} compares \datasetname with existing AIGV datasets across three dimensions. \textbf{First}, current datasets typically focus on a few basic aspects (such as video-text alignment, static and temporal visual quality), making them not suitable for assessing MLLMs as unified evaluators across various aspects. \textbf{Second}, existing datasets predominantly provide human annotations in the form of  single-video rating, which not only precludes video pair comparison analysis but also introduces greater subjectivity compared to pairwise preference annotation~\cite{co-instruct,qbench}. \textbf{Third}, the VGMs employed in existing datasets are relatively outdated, failing to represent the current SOTA performance of AIGVs. Identifying the weaknesses of such videos is both important and challenging for automatic AIGV evaluators. \datasetname addresses these limitations and provides a more comprehensive and accurate assessment for automatic AIGV evaluation models.

\section{Experiments}
\subsection{Experimental Settings}
\label{sec:exp_set}
\paragraph{Evaluated Systems.}
We conduct zero-shot evaluations on 18 MLLMs, including 15 open-sourced models along with three proprietary models: GPT-4o \cite{gpt4o}, Seed1.5-VL \cite{seed1.5-vl} and Gemini2.5-Flash \cite{gemini-2.5}. Additionally, we assess five evaluation systems specialized in particular aspects: VBench \cite{vbench}, VideoScore \cite{videoscore}, UMTScore \cite{fetv}, VIDEOCON-PHYSICS \cite{videophy} and DOVER \cite{dover}. Details of these automatic evaluators are presented in Appendix \ref{app:evaluated_systems}. To establish a human baseline, we engage three annotators to perform video pair comparisons on a subset of \datasetname, evaluating 50 samples for each subaspect.

\paragraph{Single Video Rating.} We sample 16 frames per video for our evaluation, except for Video-LLaVA, which has an 8-frame limitation. Unless otherwise specified, we adopt \textit{yes/no} as the default scoring tokens for single video rating.

\paragraph{Video Pair Comparison.} In this evaluation mode, we utilize 12 frames per video. As shown in Fig.~\ref{fig:overview} (c), the original \datasetname contains more ``$\mathcal{V}_{1/2} \text{better}$'' compared to ``same good/bad''. While this imbalance does not compromise the fairness of single video ratings (since $\mathcal{A}^{\text{single}}$ handles ``better'' and ``same'' independently) it introduces a bias toward ``$\mathcal{V}_{1/2} \text{better}$'' when computing four-way selection accuracy for video pair comparisons. To mitigate this bias, we create a subset of 2,411 samples, with a more balanced distribution across the four preference categories, in evaluating video pair comparison.

\begin{table*}[t!]
\centering
\caption{Performance of single video rating measured by $\mathcal{A}^{\text{single}}$ in Eq. \ref{eq:eq_criteria}. The best and second-best results are highlighted with \colorbox{yellow}{yellow} and \colorbox{lightyellow}{light yellow}, respectively. Abbreviations: SM (subject motion), CM (camera motion), LC (light\&color), TQ (technical quality), SC (structural correctness), AQ (aesthetic quality), AC (appearance consistency), MN (motion naturalness), TF (temporal flickering), MA (motion alignment), AA (appearance alignment).}
\resizebox{\textwidth}{!}{$
\begin{tabular}{l|c|cccc|cccc|cccc|ccc|c}
\toprule
{\makecell{\textbf{Method}}} & {\makecell{\textbf{Model}\\\textbf{Size}}} &{\makecell{\textbf{Overall}\\\textbf{Dynamic}}} & {\makecell{\textbf{SM}}} & {\makecell{\textbf{CM}}} & {\makecell{\textbf{LC}}} & {\makecell{\textbf{Overall}\\\textbf{Static}}} & {\makecell{\textbf{TQ}}} & {\makecell{\textbf{SC}}} & {\makecell{\textbf{AQ}}} & {\makecell{\textbf{Overall}\\\textbf{Temporal}}} & {\makecell{\textbf{AC}}} & {\makecell{\textbf{MN}}} & {\makecell{\textbf{TF}}} & {\makecell{\textbf{Overall}\\\textbf{Alignment}}} & {\makecell{\textbf{MA}}} & {\makecell{\textbf{AA}}} & {\makecell{\textbf{AVG}}} \\
\midrule
Random & - & 48.8 & 49.5 & 48.9 & 49.8 & 49.2 & 47.6 & 49.0 & 48.1 & 49.0 & 48.6 & 48.2 & 46.2 & 48.4 & 48.6 & 48.3 & 48.6 \\
\midrule
\multicolumn{18}{@{}l@{}}
{\cellcolor{sectionblue}\textbf{Specialized Evaluators}} \\
VideoScore-v1.1 & 8B & 57.4 & - & - & - & 40.1 & 30.4 & 47.7 & 41.9 & - & 43.1 & 36.1 & - & 38.9 & - & - & - \\
VBench & - & \cellcolor{yellow}\textbf{87.8} & - & - & - & - & 62.5 & - & 75.6 & - & 54.0 & 50.2 & - & 54.4 & - & - & -\\
UMTScore & - & - & - & - & - & - & - & - & - & - & - & - & -& 66.4 & - & - & - \\
VIDEOCON-PHYSICS & 7B & - & - & - & - & - & - & - & - & - & - & 53.4 & -& 68.7 & - & - & -\\
DOVER & 58M & - & - & - & - & - & 69.2 & - & 80.3 & - & - & - & -& - & - & - & -\\
\midrule
\multicolumn{18}{@{}l@{}}
{\cellcolor{sectionblue}\textbf{Unified Evaluators}} \\
Video-LLaVA & 7B & 52.4 & 74.8 & 63.4 & 47.6 & 47.8 & 66.1 & 44.5 & 63.4 & 44.1 & 51.3 & 55.5 & 48.6 & 59.4 & 54.9 & 66.8 & 54.5 \\
LongVA-DPO & 7B & 59.8 & 76.2 & 69.9 & 57.5 & 62.4 & 74.9 & 56.2 & 72.0 & 56.0 & 47.3 & 40.7 & 54.3 & 68.7 & 63.9 & 71.6 & 61.6 \\
ShareGPT4Video & 8B & 77.5 & 81.0 & 77.1 & 82.5 & 59.4 & 68.7 & 54.1 & 69.4 & 54.4 & 48.1 & 42.6 & 60.9 & 62.7 & 54.3 & 70.3 & 63.9 \\
VideoLLaMA2.1 & 7B & 72.1 & 80.5 & 67.1 & 77.2 & 61.4 & 78.7 & 47.5 & 72.6 & 46.1 & 50.9 & 50.2 & 61.8 & 72.6 & 65.7 & 77.7 & 64.4 \\
mPLUG-Owl3 & 7B & 78.6 & 84.8 & 77.8 & 79.9 & \cellcolor{lightyellow}76.0 & \cellcolor{lightyellow}83.1 & 55.6 & 80.0 & 59.8 & 59.5 & 42.0 & 72.0 & 80.4 & 75.2 & 87.6 & 72.6 \\
VideoChat2-Mistral & 7B & 83.1 & \cellcolor{lightyellow}92.2 & 89.6 & 74.9 & 68.1 & 76.2 & 53.8 & 74.1 & 58.7 & 58.3 & 52.8 & 85.6 & 75.6 & 78.0 & 80.9 & 72.6 \\
MiniCPM-V-2.6 & 8B & 81.4 & 86.3 & 80.3 & 88.8 & 70.9 & 75.0 & 52.9 & 80.6 & 61.1 & 59.4 & 51.7 & 70.9 & 82.1 & 74.3 & 90.8 & 73.4 \\
LLaVA-OneVision & 7B & 81.0 & 87.6 & 83.2 & 84.9 & 70.7 & 78.2 & 50.6 & 83.1 & 62.9 & 60.4 & 41.5 & 85.9 & 79.3 & 66.9 & 86.7 & 73.0 \\
LLaVA-OneVision & 72B & 82.3 & 87.8 & 78.4 & 88.2 & 71.3 & 77.6 & 60.2 & 81.5 & 64.6 & 61.9 & 39.9 & \cellcolor{lightyellow}86.8 & 84.4 & 71.7 & 93.5 & 75.0 \\
LLaVA-Video & 7B & 80.4 & 85.5 & 80.9 & 81.5 & 66.2 & 74.2 & 49.5 & 75.7 & 58.3 & 58.2 & 39.3 & 82.3 & 80.5 & 69.9 & 90.0 & 71.0 \\
LLaVA-Video & 72B & 82.8 & 86.1 & 82.9 & 86.9 & 70.2 & 80.0 & 55.4 & 77.7 & 60.1 & 59.2 & 40.4 & 83.5 & 84.8 & 73.7 & 94.6 & 74.0 \\
Qwen2-VL & 7B & 84.6 & 89.7 & \cellcolor{yellow}\textbf{94.2} & 79.7 & 64.6 & 67.3 & 50.7 & 70.6 & 51.1 & 51.3 & 48.0 & 62.7 & 85.4 & 78.7 & 92.2 & 70.9 \\
Qwen2-VL & 72B & \cellcolor{lightyellow}86.5 & \cellcolor{yellow}\textbf{92.6} & \cellcolor{lightyellow}92.7 & 86.0 & 70.6 & 76.9 & 60.2 & 83.4 & 52.5 & 58.0 & 48.9 & 71.0 & \cellcolor{yellow}\textbf{89.0} & \cellcolor{yellow}\textbf{81.8} & \cellcolor{lightyellow}95.0 & 75.4 \\
InternVL-2.5-MPO & 8B & 81.3 & 86.1 & 80.4 & 88.0 & 68.1 & 77.9 & 53.6 & 77.5 & 60.9 & 54.9 & 50.9 & 72.5 & 80.6 & 73.2 & 90.5 & 72.6 \\
InternVL-2.5-MPO & 78B & 84.3 & 86.6 & 82.4 & \cellcolor{yellow}\textbf{91.6} & 72.8 & \cellcolor{yellow}\textbf{84.0} & \cellcolor{yellow}\textbf{67.2} & 82.3 & 61.5 & 65.2 & \cellcolor{yellow}\textbf{61.8} & \cellcolor{yellow}\textbf{88.4} & \cellcolor{lightyellow}87.4 & \cellcolor{lightyellow}79.3 & \cellcolor{yellow}\textbf{95.5} & \cellcolor{lightyellow}78.2 \\
GPT-4o & - & 79.0 & 84.3 & 74.9 & 81.8 & 74.0 & 81.6 & 65.2 & \cellcolor{lightyellow}84.2 & \cellcolor{lightyellow}70.0 & \cellcolor{yellow}\textbf{79.8} & 54.0 & 58.6 & 80.8 & 77.2 & 89.6 & 75.7 \\
Seed1.5-VL & 20B Act.& 83.2 & 91.2 & 83.7 & \cellcolor{lightyellow}89.5 & \cellcolor{yellow}\textbf{82.4} & 82.9 & \cellcolor{lightyellow}66.8 & \cellcolor{yellow}\textbf{88.8} & \cellcolor{yellow}\textbf{70.5} & \cellcolor{lightyellow}78.6 & \cellcolor{lightyellow}59.5 & 70.1 & 84.2 & 79.0 & 94.2 & \cellcolor{yellow}\textbf{80.0} \\
\bottomrule
\end{tabular}
$}
\label{tab:main_single}
\end{table*}
\begin{table*}[t!]
\centering
\caption{Performance of video pair comparison measured by accuracy. The best and second-best results are highlighted with \colorbox{yellow}{yellow} and \colorbox{lightyellow}{light yellow}, respectively.}
\resizebox{\textwidth}{!}{$
\begin{tabular}{l|c|cccc|cccc|cccc|ccc|c}
\toprule
{\makecell{\textbf{Method}}} & {\makecell{\textbf{Model}\\\textbf{Size}}} &{\makecell{\textbf{Overall}\\\textbf{Dynamic}}} & {\makecell{\textbf{SM}}} & {\makecell{\textbf{CM}}} & {\makecell{\textbf{LC}}} & {\makecell{\textbf{Overall}\\\textbf{Static}}} & {\makecell{\textbf{TQ}}} & {\makecell{\textbf{SC}}} & {\makecell{\textbf{AQ}}} & {\makecell{\textbf{Overall}\\\textbf{Temporal}}} & {\makecell{\textbf{AC}}} & {\makecell{\textbf{MN}}} & {\makecell{\textbf{TF}}} & {\makecell{\textbf{Overall}\\\textbf{Alignment}}} & {\makecell{\textbf{MA}}} & {\makecell{\textbf{AA}}} & {\makecell{\textbf{AVG}}} \\
\midrule
Random & - & 25.0 & 25.0 & 25.0 & 25.0 & 25.0 & 25.0 & 25.0 & 25.0 & 25.0 & 25.0 & 25.0 & 25.0 & 25.0 & 25.0 & 25.0 & 25.0 \\
\midrule
\multicolumn{18}{@{}l@{}}
{\cellcolor{sectionblue}\textbf{Unified Evaluators}} \\
LLaVA-OneVision & 7B & 44.2 & 52.6 & 40.0 & 38.5 & 35.8 & 35.2 & 25.7 & 51.0 & 37.4 & 23.9 & 29.1 & 52.0 & 43.1 & 39.7 & 43.4 & 38.6 \\
LLaVA-OneVision & 72B & 36.5 & 55.1 & 40.0 & 53.8 & 37.0 & 51.4 & 22.2 & 54.1 & 25.6 & \cellcolor{lightyellow}43.3 & 34.5 & 45.7 & 63.0 & 58.9 & 69.7 & 44.3 \\
LLaVA-Video & 7B & 37.0 & 52.6 & 37.5 & 44.9 & 44.4 & 43.8 & 38.9 & 62.2 & 33.7 & 28.9 & 31.1 & 57.5 & 43.1 & 39.7 & 46.9 & 41.1 \\
LLaVA-Video & 72B & 42.5 & 57.7 & 32.5 & 56.4 & 41.6 & \cellcolor{lightyellow}55.2 & 31.2 & 60.2 & 32.6 & 41.1 & 27.7 & 55.9 & 60.3 & 53.0 & 66.2 & 46.1 \\
Qwen2-VL & 7B & 46.4 & 56.4 & \cellcolor{lightyellow}43.8 & 39.7 & 42.8 & 38.1 & 20.8 & 54.1 & 29.8 & 24.4 & 29.7 & 42.5 & 54.9 & 50.3 & 57.9 & 41.1 \\
Qwen2-VL & 72B & 51.4 & \cellcolor{yellow}\textbf{66.7} & \cellcolor{yellow}\textbf{71.2} & 53.8 & 47.7 & \cellcolor{yellow}\textbf{62.9} & 19.4 & 60.2 & 41.0 & 40.0 & 31.8 & 40.2 & \cellcolor{yellow}\textbf{69.7} & \cellcolor{yellow}\textbf{62.9} & \cellcolor{yellow}\textbf{78.6} & 51.6 \\
InternVL-2.5-MPO & 8B & 42.5 & 51.3 & 33.8 & 43.6 & 38.3 & 31.4 & 40.3 & 55.1 & 35.1 & 32.2 & 27.0 & 44.9 & 53.5 & 50.3 & 53.8 & 41.8 \\
InternVL-2.5-MPO & 78B & 43.1 & 57.7 & 41.2 & \cellcolor{yellow}\textbf{61.5} & \cellcolor{lightyellow}54.7 & \cellcolor{yellow}\textbf{62.9} & 27.1 & \cellcolor{yellow}\textbf{71.4} & 45.2 & \cellcolor{yellow}\textbf{47.8} & 33.8 & \cellcolor{yellow}\textbf{70.9} & \cellcolor{lightyellow}67.7 & 55.6 & \cellcolor{lightyellow}76.6 & \cellcolor{lightyellow}53.7 \\
GPT-4o & - & 42.0 & 48.7 & 38.8 & \cellcolor{lightyellow}59.0 & 53.5 & 54.3 & 41.0 & \cellcolor{yellow}\textbf{71.4} & 44.9 & 38.9 & 31.8 & 59.8 & 58.9 & 57.0 & 61.4 & 50.2 \\
Seed1.5-VL & 20B Act.& \cellcolor{lightyellow}52.5 & 56.4 & 47.5 & \cellcolor{yellow}\textbf{61.5} & 51.0 & 52.4 & \cellcolor{yellow}\textbf{44.4} & 62.2 & \cellcolor{lightyellow}48.6 & 36.7 & \cellcolor{lightyellow}35.8 & \cellcolor{lightyellow}65.4 & 56.9 & 53.6 & 61.4 & 51.6 \\
Gemini2.5-Flash & - & \cellcolor{yellow}\textbf{55.8} & \cellcolor{lightyellow}60.3 & 45.0 & \cellcolor{lightyellow}59.0 & \cellcolor{yellow}\textbf{55.6} & 50.5 & \cellcolor{lightyellow}43.1 & \cellcolor{lightyellow}64.3 & \cellcolor{yellow}\textbf{50.8} & 41.7 & \cellcolor{yellow}\textbf{38.5} & 60.6 & 65.0 & \cellcolor{lightyellow}62.3 & 66.2 & \cellcolor{yellow}\textbf{54.6} \\
\midrule
Human & - & 87.3 & 85.3 & 87.3 & 85.3 & 90.0 & 92.0 & 88.0 & 91.3 & 88.0 & 90.0 & 78.7 & 92.0 & 88.0 & 84.7 & 92.0 & 88.0 \\
\hline
\end{tabular}
$}
\label{tab:main_pair}
\end{table*}
\begin{figure*}[t]
    \parbox{0.70\linewidth}{
       \centering
        \subfigure[Single Video Rating.]{
            \label{fig:exp_prompt_strategy_single_internvl_78b}
            \includegraphics[width=.45\linewidth]{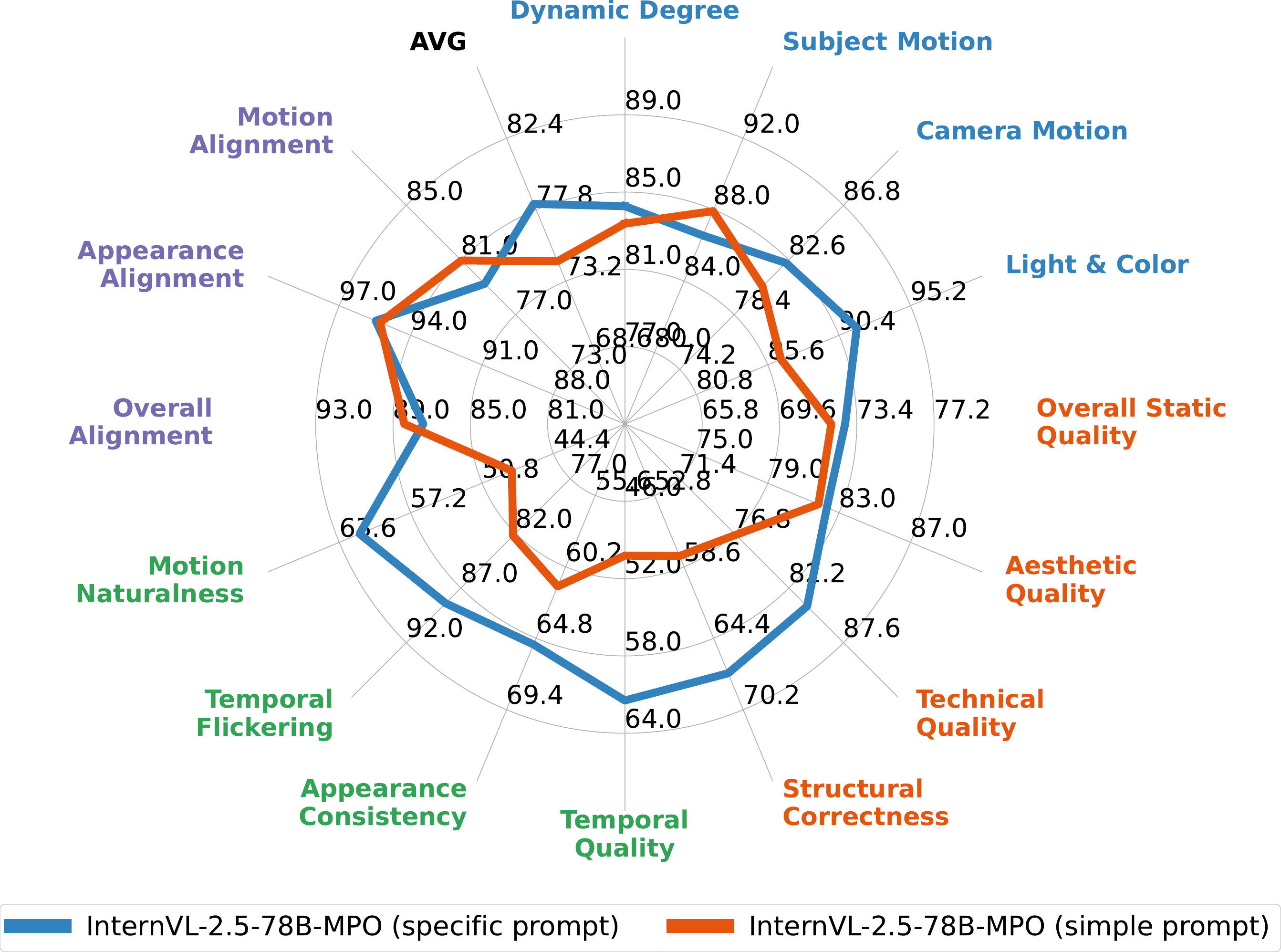}}
        \subfigure[Video Pair Comparison.]{
            \label{fig:exp_prompt_strategy_pair_internvl_78b}
            \includegraphics[width=.45\linewidth]{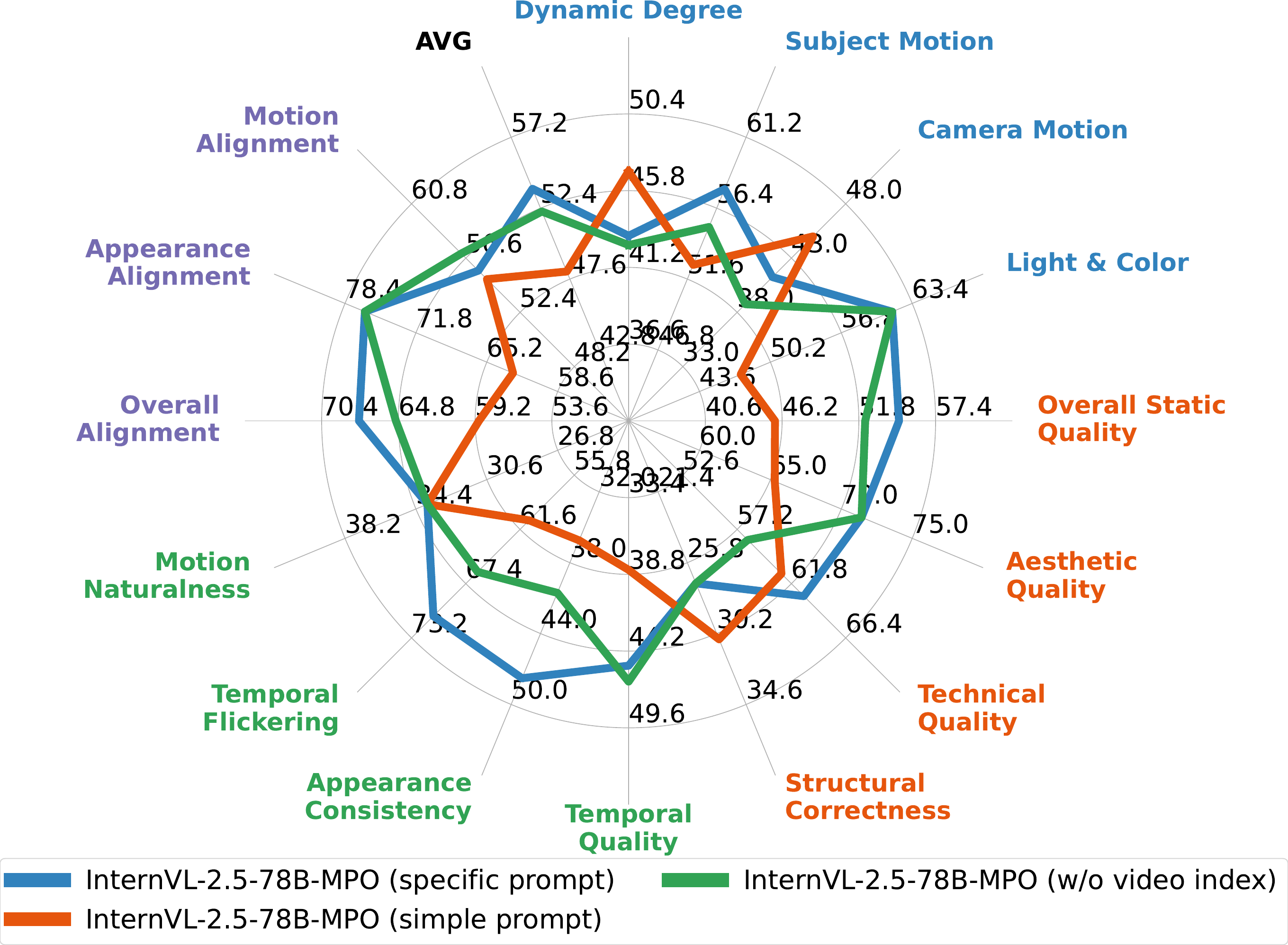}}
        \caption{Results of different prompting strategies.}
      \label{fig:exp_prompt_strategy_internvl}
    }
    \hfill
    \parbox{0.29\linewidth}{
       \centering
        \includegraphics[width=1.\linewidth]{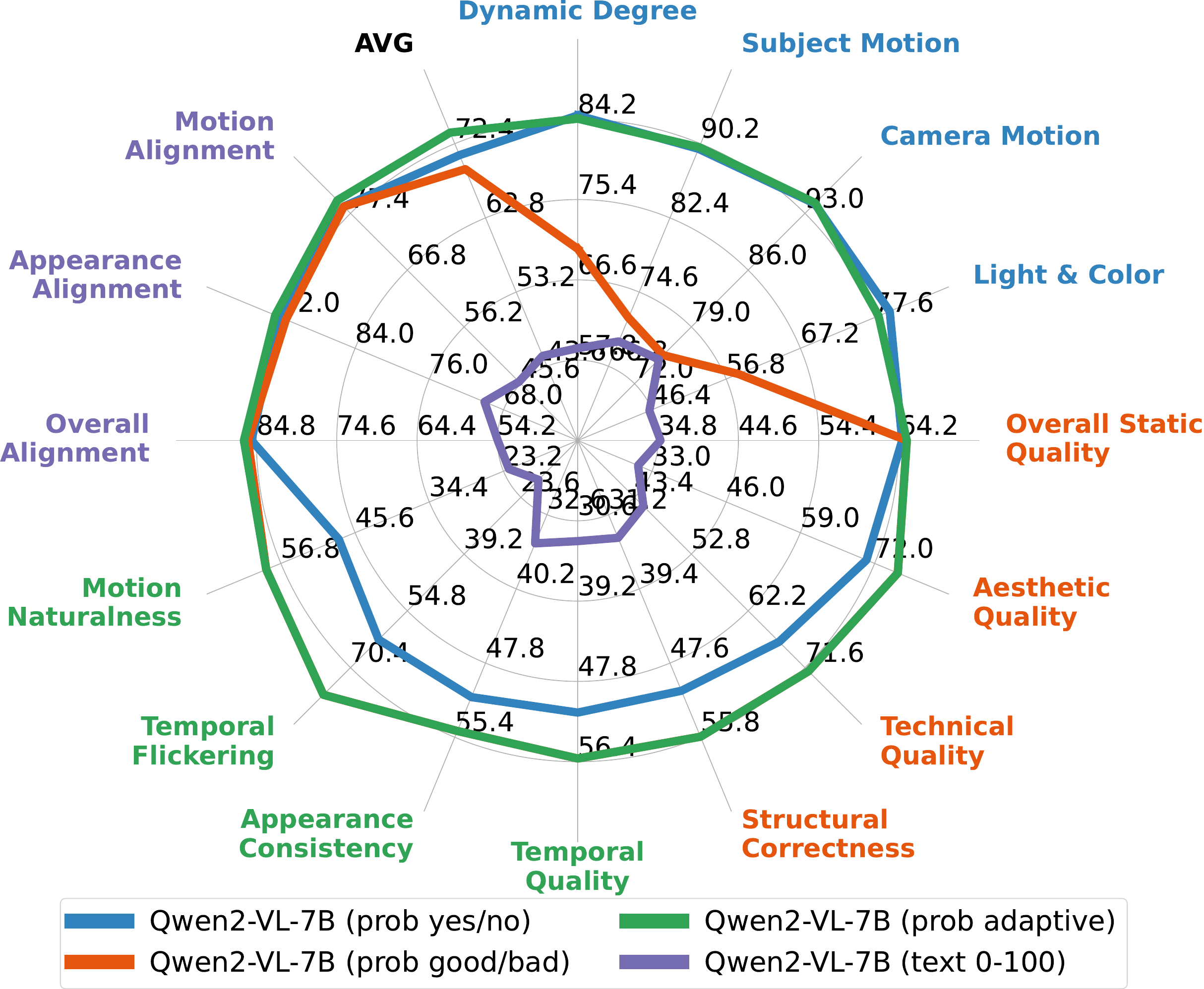}
        \caption{Results of different scoring strategies with Qwen2-VL-7B for single video rating.}
        \label{fig:exp_scoring_strategy}
    }
\end{figure*}

\subsection{Performance of MLLMs as Unified Evaluator}
\paragraph{Single Video Rating.}
Tab.~\ref{tab:main_single} presents the results of MLLMs in single video rating, from which we can obtain the following findings: \textbf{(1)} SOTA MLLMs demonstrate strong capabilities in assessing \textit{Dynamic Degree} and \textit{Video-Text Alignment}, with Qwen2-VL-72B achieving over $80$ $\mathcal{A}^{\text{single}}$ score across all eight subaspects. \textbf{(2)} current MLLMs show limitations in evaluating \textit{Temporal Quality} aspects, which require a nuanced understanding of video temporal dynamics. Additionally, their performance in \textit{Motion Alignment} lags behind that of \textit{Appearance Alignment}. These observations echo with previous research \cite{tempcompass,tomato} which reveal the limitation of MLLMs in fine-grained video temporal understanding. Interestingly, Video-LLaVA and VideoChat2-Mistral, despite their lower average $\mathcal{A}^{\text{single}}$, perform well in assessing motion naturalness, surpassing the 72B Qwen2-VL and LLaVA-OneVision. We attribute this to their use of native video encoders, rather than the frame-by-frame processing approach used by most MLLMs. \textbf{(3)} While advanced MLLMs can effectively assess \textit{Technical Quality} and \textit{Aesthetic Quality}, they still struggle to identify incorrect subject structures (e.g., human hands with six fingers). \textbf{(4)} MLLMs' performance in single video rating strongly correlates with their general multimodal understanding capability: Larger models, such as Qwen2-VL-72B, InternVL2.5-78B-MPO, GPT-4o and Seed1.5-VL demonstrate superior performance compared to the smaller ones at 7B scale. \textbf{(5)} The unified evaluators consistently outperform or match specialized methods across all evaluation aspects. This indicates that utilizing MLLMs for unified AIGV evaluation is a promising direction, which showcases both versatility and effectiveness.

\paragraph{Video Pair Comparison.}
As shown in Tab.~\ref{tab:main_pair}, the MLLMs' performance in video pair comparison exhibits a similar pattern as single video rating: They achieve relative higher accuracy when evaluating \textit{Dynamic Degree} subaspects, \textit{Technical Quality}, \textit{Aesthetic Quality} and \textit{Appearance Alignment}, while showing weakness in assessing \textit{Structural Correctness}, \textit{Motion Naturalness} and \textit{Motion Alignment}. Additionally, the significant performance gap of over 30 points between SOTA MLLMs and human evaluators indicates that current MLLMs have not yet achieved reliable pairwise video quality assessment capabilities.

\subsection{Factors Influencing Unified Evaluator Performance}
\subsubsection{Impact of Prompting Strategy}
\label{sec:effect_prompt_strategy}
We investigate how different prompting strategies affect model performance by modifying our original prompt templates (Tab.~\ref{tab:prompt_template}). We test two variations: (1) removing the aspect-specific descriptions while maintaining only the question and answer instructions, and (2) for video pair comparisons, eliminating video order indicators (``The first/second video''). The latter modification was motivated by \citet{QualityAssessmentSurvey}, which emphasizes the importance of specifying relative order in image quality comparisons. The modified prompts can be found in the supplementary material.

\paragraph{Single Video Rating.} The results in Fig.~\ref{fig:exp_prompt_strategy_internvl} (a) demonstrate that using simplified prompts (without aspect-specific descriptions) generally led to decreased performance in InternVL-2.5-78B-MPO. The performance decline was particularly pronounced in several categories: \textit{Temporal Quality} subaspects, \textit{Structural Correctness}, \textit{Technical Quality}, and \textit{Light \& Color}. However, other aspects maintain relatively stable performance levels. This phenomenon suggests that it is generally beneficial to provide a detailed description of the aspect to evaluate. Results of other models also showcase an advantage of detailed aspect description on the average performance (Appendix \ref{app:effect_prompting_strategy}).

\begin{figure*}[t]
    \parbox{0.54\linewidth}{
       \centering
        \subfigure[Single Video Rating.]{
            \label{fig:exp_video_frame_single}
            \includegraphics[width=.54\textwidth]{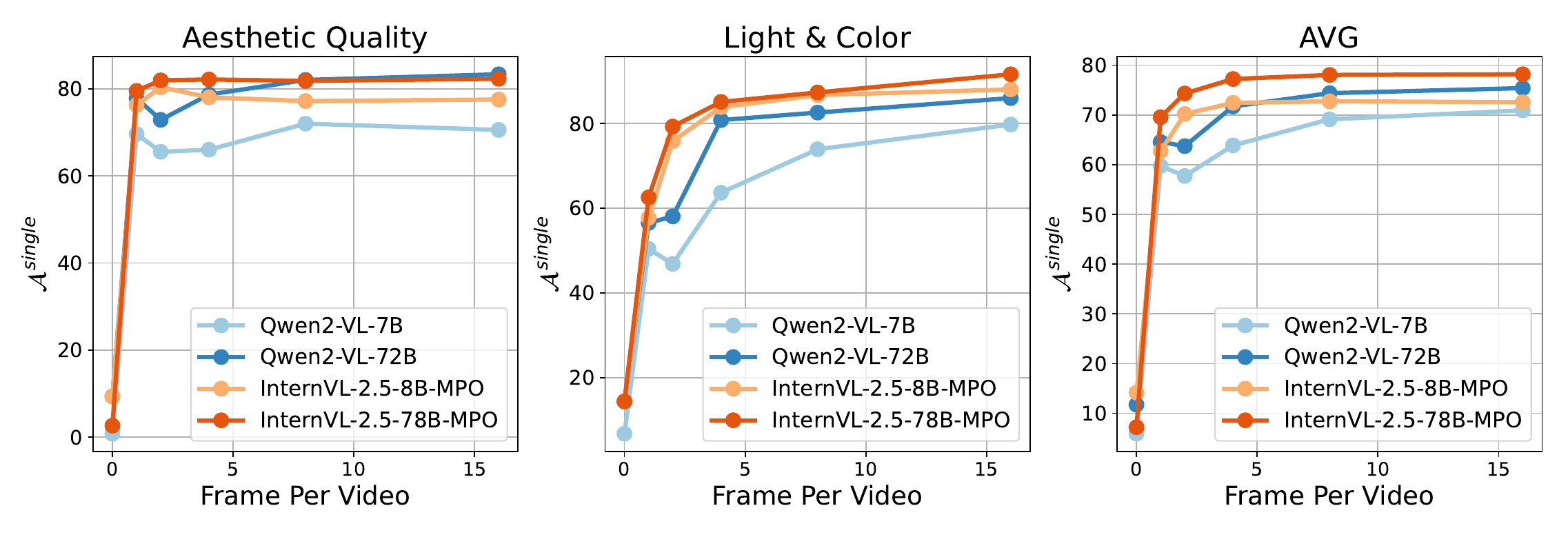}}
        \subfigure[Video Pair Comparison.]{
            \label{fig:exp_video_frame_pair}
            \includegraphics[width=.54\textwidth]{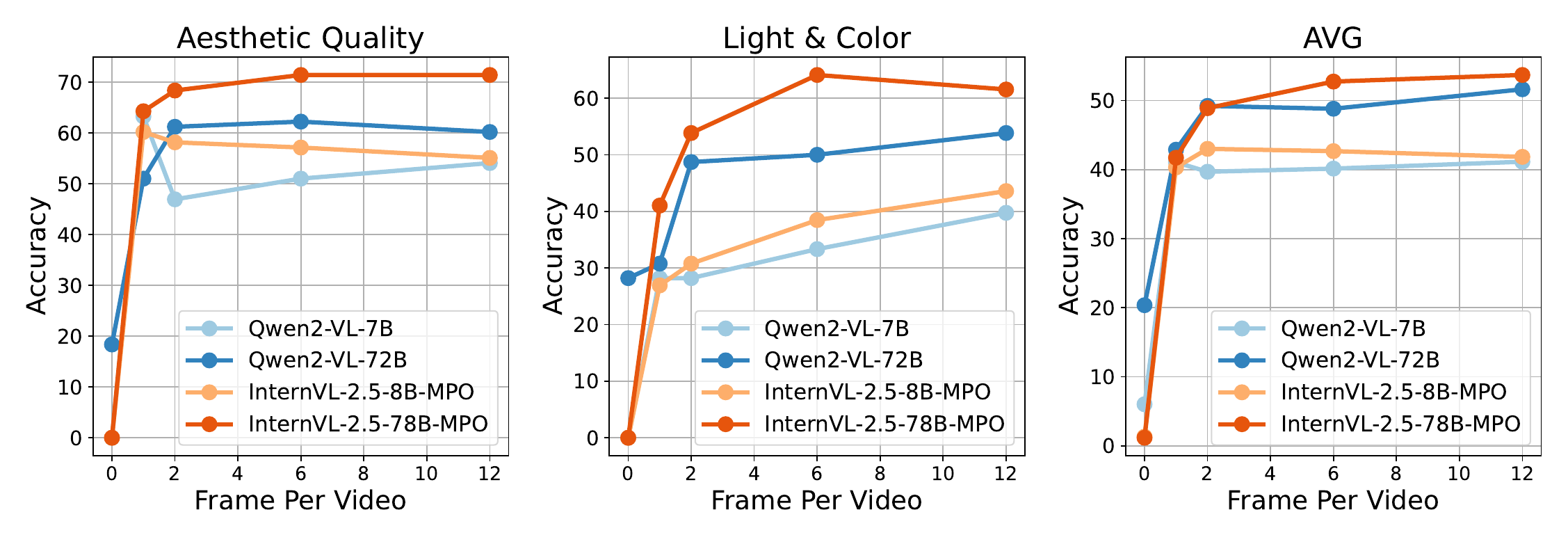}}
        \caption{Results with varying numbers of video frames.}
      \label{fig:exp_video_frame}
    }
    \hfill
    \parbox{0.44\linewidth}{
       \centering
        \includegraphics[width=1.\linewidth]{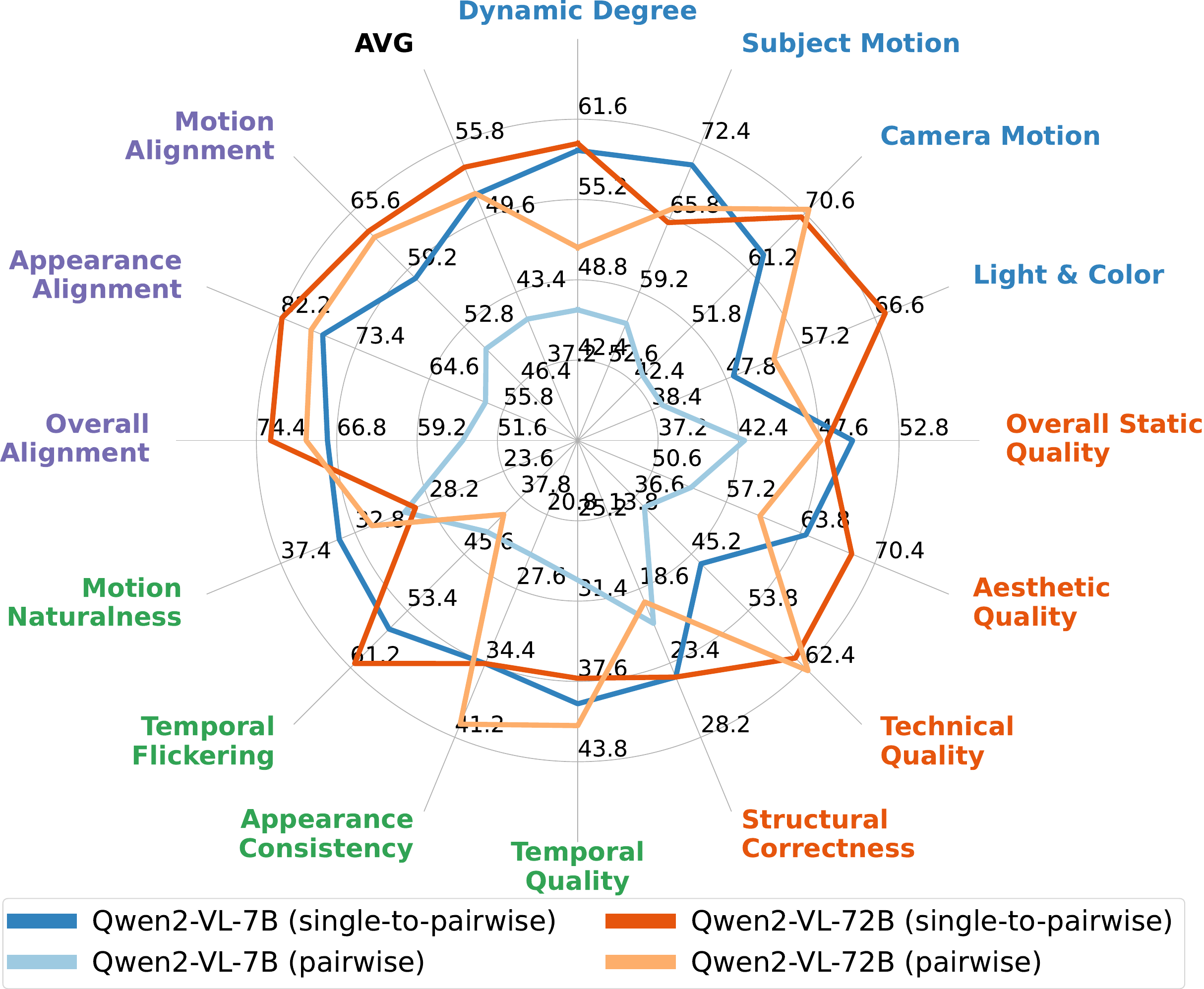}
        \caption{Results of direct video pair comparison versus adapting single video rating to pairwise comparison.}
        \label{fig:exp_qwen2vl_s2p}
    }
\end{figure*}

\paragraph{Video Pair Comparison.} Similarly, Fig.~\ref{fig:exp_prompt_strategy_internvl} (b) shows that simplified prompts result in reduced performance for video pair comparisons, reinforcing the value of detailed aspect descriptions in the prompts. Interestingly, contrary to established findings in image quality comparison \cite{QualityAssessmentSurvey}, removing video order indicators does not significantly impact performance. We attribute this robustness to the enhanced temporal order understanding capabilities of recent MLLMs.

\subsubsection{Impact of Scoring Strategy}
\label{sec:effect_scoring_strategy}
In the previous experiments, we adopt \textit{yes/no} as the default scoring tokens and calculate single video ratings using the \textit{yes} probability according to Eq. \ref{eq:single_video_rating}. Here, we explore three different scoring strategies: (1) using \textit{good/bad} as scoring tokens, (2) an adaptive approach that alternates between \textit{yes/no} and \textit{good/bad} tokens, and (3) prompting the MLLM to directly generate rating scores as texts, ranging from 0 to 100. The detailed prompts are shown in the supplementary material.

The results of Qwen2-VL-7B, presented in Fig.~\ref{fig:exp_scoring_strategy}, reveal that: (1) \textit{good/bad} demonstrates superior performance for \textit{Temporal Quality} and \textit{Static Quality} assessments, while \textit{yes/no} performs better for \textit{Dynamic Degree} evaluation. (2) The adaptive strategy yields the best overall results for Qwen2-VL-7B. However, for other advanced MLLMs like InternVL-2.5-78B-MPO, the improvement from adaptive scoring is minimal (see Appendix \ref{app:effect_scoring_strategy}). (3) Directly prompting MLLMs to generate rating scores performs significantly worse than probability-based scoring methods. These findings suggest that using \textit{yes/no} as unified scoring tokens strikes a good balance between simplicity and effectiveness.

\subsubsection{Adapting Single Ratings for Pairwise Comparison}
\label{sec:s2p}
While our previous experiments involve direct pairwise video comparisons by simultaneously feeding two videos into MLLMs, we now explore an alternative approach. This method first independently rates each video using Eq. \ref{eq:single_video_rating} and then converts these individual ratings into a four-way selection from set $\mathbf{O}$. The detailed conversion methodology is discussed in Appendix \ref{app:s2p_adaptation_method}.

As shown in Fig.~\ref{fig:exp_qwen2vl_s2p}, adaptation from single video rating brings substantial improvement to Qwen2-VL-7B compared to directly performing video pair comparison. However, Qwen2-VL-72B shows only minimal improvements under the same approach. Similar patterns were observed with InternVL-2.5-MPO, as shown in Appendix \ref{app:s2p_results}. We hypothesize that smaller-scale models, like the 7B variant, have inherently weaker pairwise comparison capabilities, and adapting from single video rating bypasses this limitation. Based on these findings, we recommend implementing the single-to-pairwise adaptation strategy particularly for MLLMs with limited pairwise comparison capabilities.

\subsubsection{Impact of Frame Number}
Fig.~\ref{fig:exp_video_frame} illustrates the impact of varying video frame numbers on model performance. We can see that: (1) In the zero-frame scenario (pure-text input), MLLMs perform poorly as their outputs remain constant regardless of visual information, confirming that language shortcuts cannot be exploited to achieve good performance in \datasetname. (2) As the frame count increases, aspects requiring temporal understanding (\textit{Light \& Color}) show more consistent improvement compared to aspects focusing on individual frame information (\textit{Aesthetic Quality}). (3) The 72B-scale models demonstrate more consistent improvement in average accuracy with increasing frame numbers compared to their 7B-scale counterparts. This suggests that larger models possess superior capabilities in processing and integrating information from additional video frames.

\begin{table}[t]
    \centering
    \caption{Automatic and human rating of video generative models. The human prefer rate is computed as the ratio of ``better'' plus ``same good'' examples, based on video pair preference annotation.}
    \resizebox{1.\textwidth}{!}{$
    \begin{tabular}{lccccc}
         \toprule
          &\textbf{Open-Sora 1.2} & \textbf{CogVideoX-5B}  & \textbf{Movie Gen Video} & \textbf{HunyuanVideo} & \textbf{Real Videos}  \\
         \midrule
         \multicolumn{6}{@{}l@{}}
            {\cellcolor{sectionblue}\textbf{Static Quality}} \\
        Auto (Qwen2-VL-72B)   & 63.2 & 70.0 & 84.3 & 84.9 & \textbf{85.7}  \\
         Human (Prefer Rate) & 3.3 & 21.6 & 96.4 & 94.7 & - \\
         \midrule
         \multicolumn{6}{@{}l@{}}
            {\cellcolor{sectionblue}\textbf{Temporal Quality}} \\
         Auto (Qwen2-VL-72B) & 59.7 & 67.3 & 74.7 & 77.1 & \textbf{78.0}   \\
         Human (Prefer Rate) & 1.5 & 42.4 & 96.6 & 86.6 & - \\
         \bottomrule
    \end{tabular}
    $}
    \label{tab:vgm_comparison}
\end{table}
\subsection{Evaluation of Real versus Generated Videos}
To examine the difference when applying MLLMs to evaluate real videos instead of AI-generated videos, we collect 204 real videos from the TempCompass \cite{tempcompass} dataset and utilize Qwen2-VL-72B to perform single video rating on two aspects: \textit{Static Quality} and \textit{Temporal Quality}. For comparison, we also evaluate videos generated by Open-Sora 1.2, CogVideoX-5B, Movie Gen Video and HunyuanVideo. As shown in Tab.~\ref{tab:vgm_comparison}, real videos consistently receive higher scores than generated ones across both quality aspects. Furthermore, the relative ranking of VGMs basically aligns with human perceptual judgments: the more recent models (Movie Gen Video and HunyuanVideo) outperform earlier ones (Open-Sora 1.2 and CogVideoX-5B) in both human and automatic evaluations. However, Qwen2-VL-72B and humans disagree slightly on the rating of Movie Gen Video and HunyuanVideo, highlighting the need for more advanced evaluators capable of finer-grained comparisons.

\section{Conclusion and Future Work}
\label{sec:conclusion}
This study investigates the potential of MLLMs to serve as unified evaluators for AI-generated videos. To answer this question, we introduce the \datasetname, a benchmark designed to assess the capability of unified AIGV evaluation. Compared to existing AIGV datasets, \datasetname highlights (1) comprehensive evaluation aspects, (2) videos produced by SOTA video generative models and (3) reliable evaluation of both single video rating and video pair comparison. Based on \datasetname, we extensively evaluates 18 MLLMs. Our findings reveal that while SOTA MLLMs cannot fully replace human evaluators, they significantly outperform existing specialized evaluation methods, showcasing promising potential as unified AIGV evaluators. Our analytical studies highlight several design choices, including: providing detailed aspect-specific description in the prompts (\S \ref{sec:effect_prompt_strategy}), using \textit{yes/no} as the scoring tokens for single video rating (\S \ref{sec:effect_scoring_strategy}) and adapting single video rating for pairwise comparison when using 7B-scale MLLMs (\S \ref{sec:s2p}).

Looking ahead, we identify two main directions for future work: \textbf{First}, we aim to extend the benchmark to cover broader AIGV evaluation scenarios, including safety, ethics, and bias assessments, as well as more diverse generation settings such as image-to-video and video-to-video tasks. \textbf{Second}, we plan to enhance MLLM performance in AIGV evaluation through three complementary directions: (1) \textbf{Advancing video encoding techniques} to capture richer temporal information without compromising efficiency, which is essential to the temporally sensitive evaluation aspects. (2) \textbf{Automatically curating high- and low-quality video pairs} to train evaluators. This can be achieved by leveraging VGMs at different training stages or intentionally varying prompts or other hyperparameters during inference to generate controlled degradations in videos. (3) \textbf{Incorporating visual generative loss} into MLLM training—alongside language generation loss—to strengthen their understanding of real-world visual distributions and underlying physical dynamics.

\begin{ack}
We thank all the anonymous reviewers for their constructive comments. This work is supported in
part by ByteDance Seed and National Natural Science Foundation of China (No. 62176002). Haoyuan Guo and Xu Sun are the corresponding authors of this paper.
\end{ack}

\bibliography{neurips_2025}

\begin{thebibliography}{80}
\providecommand{\natexlab}[1]{#1}
\providecommand{\url}[1]{\texttt{#1}}
\expandafter\ifx\csname urlstyle\endcsname\relax
  \providecommand{\doi}[1]{doi: #1}\else
  \providecommand{\doi}{doi: \begingroup \urlstyle{rm}\Url}\fi

\bibitem[Bansal et~al.(2024{\natexlab{a}})Bansal, Bitton, Szpektor, Chang, and Grover]{VideoCon}
H.~Bansal, Y.~Bitton, I.~Szpektor, K.~Chang, and A.~Grover.
\newblock Videocon: Robust video-language alignment via contrast captions.
\newblock In \emph{{CVPR}}, pages 13927--13937. {IEEE}, 2024{\natexlab{a}}.

\bibitem[Bansal et~al.(2024{\natexlab{b}})Bansal, Lin, Xie, Zong, Yarom, Bitton, Jiang, Sun, Chang, and Grover]{videophy}
H.~Bansal, Z.~Lin, T.~Xie, Z.~Zong, M.~Yarom, Y.~Bitton, C.~Jiang, Y.~Sun, K.-W. Chang, and A.~Grover.
\newblock Videophy: Evaluating physical commonsense for video generation.
\newblock \emph{arXiv preprint arXiv:2406.03520}, 2024{\natexlab{b}}.

\bibitem[Brooks et~al.(2024)Brooks, Peebles, Holmes, DePue, Guo, Jing, Schnurr, Taylor, Luhman, Luhman, Ng, Wang, and Ramesh]{sora}
T.~Brooks, B.~Peebles, C.~Holmes, W.~DePue, Y.~Guo, L.~Jing, D.~Schnurr, J.~Taylor, T.~Luhman, E.~Luhman, C.~Ng, R.~Wang, and A.~Ramesh.
\newblock Video generation models as world simulators.
\newblock 2024.

\bibitem[Cai et~al.(2024)Cai, Cao, Chen, Chen, Chen, Chen, Chen, Chen, Chen, Chu, Dong, Duan, Fan, Fei, Gao, Ge, Gu, Gu, Gui, Guo, Guo, He, Hu, Huang, Jiang, Jiao, Jin, Lei, Li, Li, Li, Li, Li, Li, Liu, Liu, Hong, Liu, Liu, Liu, Lv, Lv, Lv, Ma, Ma, Ma, Ning, Ouyang, Qiu, Qu, Shang, Shao, Song, Song, Sui, Sun, Sun, Tang, Wang, Wang, Wang, Wang, Wang, Wang, Wang, Wei, Weng, Wu, Xiong, Zhao, and et~al.]{internlm2.5}
Z.~Cai, M.~Cao, H.~Chen, K.~Chen, K.~Chen, X.~Chen, X.~Chen, Z.~Chen, Z.~Chen, P.~Chu, X.~Dong, H.~Duan, Q.~Fan, Z.~Fei, Y.~Gao, J.~Ge, C.~Gu, Y.~Gu, T.~Gui, A.~Guo, Q.~Guo, C.~He, Y.~Hu, T.~Huang, T.~Jiang, P.~Jiao, Z.~Jin, Z.~Lei, J.~Li, J.~Li, L.~Li, S.~Li, W.~Li, Y.~Li, H.~Liu, J.~Liu, J.~Hong, K.~Liu, K.~Liu, X.~Liu, C.~Lv, H.~Lv, K.~Lv, L.~Ma, R.~Ma, Z.~Ma, W.~Ning, L.~Ouyang, J.~Qiu, Y.~Qu, F.~Shang, Y.~Shao, D.~Song, Z.~Song, Z.~Sui, P.~Sun, Y.~Sun, H.~Tang, B.~Wang, G.~Wang, J.~Wang, J.~Wang, R.~Wang, Y.~Wang, Z.~Wang, X.~Wei, Q.~Weng, F.~Wu, Y.~Xiong, X.~Zhao, and et~al.
\newblock Internlm2 technical report.
\newblock \emph{CoRR}, abs/2403.17297, 2024.

\bibitem[Caron et~al.(2021)Caron, Touvron, Misra, J{\'{e}}gou, Mairal, Bojanowski, and Joulin]{DINO}
M.~Caron, H.~Touvron, I.~Misra, H.~J{\'{e}}gou, J.~Mairal, P.~Bojanowski, and A.~Joulin.
\newblock Emerging properties in self-supervised vision transformers.
\newblock In \emph{{ICCV}}, pages 9630--9640. {IEEE}, 2021.

\bibitem[Chen et~al.(2023{\natexlab{a}})Chen, Xia, He, Zhang, Cun, Yang, Xing, Liu, Chen, Wang, et~al.]{chen2023videocrafter1}
H.~Chen, M.~Xia, Y.~He, Y.~Zhang, X.~Cun, S.~Yang, J.~Xing, Y.~Liu, Q.~Chen, X.~Wang, et~al.
\newblock Videocrafter1: Open diffusion models for high-quality video generation.
\newblock \emph{arXiv preprint arXiv:2310.19512}, 2023{\natexlab{a}}.

\bibitem[Chen et~al.(2024{\natexlab{a}})Chen, Wei, Li, Dong, Zhang, Zang, Chen, Duan, Lin, Tang, Yuan, Qiao, Lin, Zhao, and Wang]{ShareGPT4Video}
L.~Chen, X.~Wei, J.~Li, X.~Dong, P.~Zhang, Y.~Zang, Z.~Chen, H.~Duan, B.~Lin, Z.~Tang, L.~Yuan, Y.~Qiao, D.~Lin, F.~Zhao, and J.~Wang.
\newblock Sharegpt4video: Improving video understanding and generation with better captions.
\newblock \emph{ArXiv preprint}, abs/2406.04325, 2024{\natexlab{a}}.

\bibitem[Chen et~al.(2023{\natexlab{b}})Chen, Wu, Wang, Su, Chen, Xing, Zhong, Zhang, Zhu, Lu, Li, Luo, Lu, Qiao, and Dai]{internvl}
Z.~Chen, J.~Wu, W.~Wang, W.~Su, G.~Chen, S.~Xing, M.~Zhong, Q.~Zhang, X.~Zhu, L.~Lu, B.~Li, P.~Luo, T.~Lu, Y.~Qiao, and J.~Dai.
\newblock Internvl: Scaling up vision foundation models and aligning for generic visual-linguistic tasks.
\newblock \emph{CoRR}, abs/2312.14238, 2023{\natexlab{b}}.

\bibitem[Chen et~al.(2024{\natexlab{b}})Chen, Wang, Cao, Liu, Gao, Cui, Zhu, Ye, Tian, Liu, et~al.]{internvl2.5}
Z.~Chen, W.~Wang, Y.~Cao, Y.~Liu, Z.~Gao, E.~Cui, J.~Zhu, S.~Ye, H.~Tian, Z.~Liu, et~al.
\newblock Expanding performance boundaries of open-source multimodal models with model, data, and test-time scaling.
\newblock \emph{arXiv preprint arXiv:2412.05271}, 2024{\natexlab{b}}.

\bibitem[Cheng et~al.(2024)Cheng, Leng, Zhang, Xin, Li, Chen, Zhu, Zhang, Luo, Zhao, and Bing]{videollama2}
Z.~Cheng, S.~Leng, H.~Zhang, Y.~Xin, X.~Li, G.~Chen, Y.~Zhu, W.~Zhang, Z.~Luo, D.~Zhao, and L.~Bing.
\newblock Videollama 2: Advancing spatial-temporal modeling and audio understanding in video-llms.
\newblock \emph{CoRR}, abs/2406.07476, 2024.

\bibitem[Chiang et~al.(2023)Chiang, Li, Lin, Sheng, Wu, Zhang, Zheng, Zhuang, Zhuang, Gonzalez, Stoica, and Xing]{Vicuna}
W.-L. Chiang, Z.~Li, Z.~Lin, Y.~Sheng, Z.~Wu, H.~Zhang, L.~Zheng, S.~Zhuang, Y.~Zhuang, J.~E. Gonzalez, I.~Stoica, and E.~P. Xing.
\newblock Vicuna: An open-source chatbot impressing gpt-4 with 90\%* chatgpt quality, March 2023.
\newblock URL \url{https://lmsys.org/blog/2023-03-30-vicuna/}.

\bibitem[Dubey et~al.(2024)Dubey, Jauhri, Pandey, Kadian, Al{-}Dahle, Letman, Mathur, Schelten, Yang, Fan, Goyal, Hartshorn, Yang, Mitra, Sravankumar, Korenev, Hinsvark, Rao, Zhang, Rodriguez, Gregerson, Spataru, Rozi{\`{e}}re, Biron, Tang, Chern, Caucheteux, Nayak, Bi, Marra, McConnell, Keller, Touret, Wu, Wong, Ferrer, Nikolaidis, Allonsius, Song, Pintz, Livshits, Esiobu, Choudhary, Mahajan, Garcia{-}Olano, Perino, Hupkes, Lakomkin, AlBadawy, Lobanova, Dinan, Smith, Radenovic, Zhang, Synnaeve, Lee, Anderson, Nail, Mialon, Pang, Cucurell, Nguyen, Korevaar, Xu, Touvron, Zarov, Ibarra, Kloumann, Misra, Evtimov, Copet, Lee, Geffert, Vranes, Park, Mahadeokar, Shah, van~der Linde, Billock, Hong, Lee, Fu, Chi, Huang, Liu, Wang, Yu, Bitton, Spisak, Park, Rocca, Johnstun, Saxe, Jia, Alwala, Upasani, Plawiak, Li, Heafield, Stone, and et~al.]{llama3}
A.~Dubey, A.~Jauhri, A.~Pandey, A.~Kadian, A.~Al{-}Dahle, A.~Letman, A.~Mathur, A.~Schelten, A.~Yang, A.~Fan, A.~Goyal, A.~Hartshorn, A.~Yang, A.~Mitra, A.~Sravankumar, A.~Korenev, A.~Hinsvark, A.~Rao, A.~Zhang, A.~Rodriguez, A.~Gregerson, A.~Spataru, B.~Rozi{\`{e}}re, B.~Biron, B.~Tang, B.~Chern, C.~Caucheteux, C.~Nayak, C.~Bi, C.~Marra, C.~McConnell, C.~Keller, C.~Touret, C.~Wu, C.~Wong, C.~C. Ferrer, C.~Nikolaidis, D.~Allonsius, D.~Song, D.~Pintz, D.~Livshits, D.~Esiobu, D.~Choudhary, D.~Mahajan, D.~Garcia{-}Olano, D.~Perino, D.~Hupkes, E.~Lakomkin, E.~AlBadawy, E.~Lobanova, E.~Dinan, E.~M. Smith, F.~Radenovic, F.~Zhang, G.~Synnaeve, G.~Lee, G.~L. Anderson, G.~Nail, G.~Mialon, G.~Pang, G.~Cucurell, H.~Nguyen, H.~Korevaar, H.~Xu, H.~Touvron, I.~Zarov, I.~A. Ibarra, I.~M. Kloumann, I.~Misra, I.~Evtimov, J.~Copet, J.~Lee, J.~Geffert, J.~Vranes, J.~Park, J.~Mahadeokar, J.~Shah, J.~van~der Linde, J.~Billock, J.~Hong, J.~Lee, J.~Fu, J.~Chi, J.~Huang, J.~Liu, J.~Wang, J.~Yu, J.~Bitton, J.~Spisak, J.~Park,
  J.~Rocca, J.~Johnstun, J.~Saxe, J.~Jia, K.~V. Alwala, K.~Upasani, K.~Plawiak, K.~Li, K.~Heafield, K.~Stone, and et~al.
\newblock The llama 3 herd of models.
\newblock \emph{CoRR}, abs/2407.21783, 2024.

\bibitem[Gao et~al.(2024)Gao, Zhang, Liu, Qiu, Huang, Lin, Zhao, Geng, Lin, Jin, Zhang, Shao, Xu, He, He, Shao, Lu, Li, and Qiao]{SPHINX-X}
P.~Gao, R.~Zhang, C.~Liu, L.~Qiu, S.~Huang, W.~Lin, S.~Zhao, S.~Geng, Z.~Lin, P.~Jin, K.~Zhang, W.~Shao, C.~Xu, C.~He, J.~He, H.~Shao, P.~Lu, H.~Li, and Y.~Qiao.
\newblock Sphinx-x: Scaling data and parameters for a family of multi-modal large language models.
\newblock \emph{ArXiv}, abs/2402.05935, 2024.
\newblock URL \url{https://api.semanticscholar.org/CorpusID:267547619}.

\bibitem[Google and DeepMind(2025)]{gemini-2.5}
Google and DeepMind.
\newblock Gemini 2.5: Our most intelligent ai model, 2025.

\bibitem[Guo et~al.(2024)Guo, Wu, Zhu, Leng, Shi, Chen, Fan, Wang, Jiang, Wang, Chen, Huang, Lei, Yuan, Luo, Liu, Ye, Qian, Yan, Zhao, Peng, Li, Yuan, Wu, Cheng, Liu, Wang, Zeng, Liu, Qin, Ding, Xiao, Zhang, Zhang, Xiong, Peng, Chen, Li, Hu, Lin, Hu, Zhang, Wu, Li, Liu, Ling, Qin, Wang, He, Zhang, Yi, Liao, Huang, Zhang, Deng, Deng, Lin, Yuan, Li, Gou, Lou, Wei, Liu, Li, Zhu, Zhong, Li, Zhang, Wu, Li, Xiao, Lin, Yang, Wang, Ji, Hao, Shen, Li, Li, Wu, Zhu, Jiao, Feng, Chen, Duan, Liu, Zeng, Tang, Sun, Chen, Long, Feng, Zhan, Fang, Lu, Hua, Liu, Shen, Zhang, Shen, Wang, Pan, Zhang, Li, Li, Li, Shi, Han, Xiang, Chen, Chen, Li, Yan, Chi, Liu, Du, Wang, Pan, Chen, Chen, Wu, Yuan, Shuai, Tao, Zheng, Zhang, Zhang, Wang, Yang, Zhao, Xu, Liang, Yan, Zhong, Cao, Wu, Liu, Chang, Cai, Ao, Yang, Zhang, Zhong, Jia, Weng, Yu, Huang, Zhu, Yang, Wang, Long, Yin, Li, Zhu, Jia, Zhang, Liu, Zhang, Yang, Luo, Chen, Zhong, Xiao, Li, Wu, Wen, Du, Zhang, Ye, Wu, Liu, Yue, Zhou, Yuan, Xu, Yang, Zhang, Fang, Li, Ren, Xiong, Hong,
  Wang, Sun, Wang, Cai, Zha, An, Zhao, Xu, Chen, Wu, Zheng, Wang, Huang, Zhu, and Song]{seed1.5-vl}
D.~Guo, F.~Wu, F.~Zhu, F.~Leng, G.~Shi, H.~Chen, H.~Fan, J.~Wang, J.~Jiang, J.~Wang, J.~Chen, J.~Huang, K.~Lei, L.~Yuan, L.~Luo, P.~Liu, Q.~Ye, R.~Qian, S.~Yan, S.~Zhao, S.~Peng, S.~Li, S.~Yuan, S.~Wu, T.~Cheng, W.~Liu, W.~Wang, X.~Zeng, X.~Liu, X.~Qin, X.~Ding, X.~Xiao, X.~Zhang, X.~Zhang, X.~Xiong, Y.~Peng, Y.~Chen, Y.~Li, Y.~Hu, Y.~Lin, Y.~Hu, Y.~Zhang, Y.~Wu, Y.~Li, Y.~Liu, Y.~Ling, Y.~Qin, Z.~Wang, Z.~He, A.~Zhang, B.~Yi, B.~Liao, C.~Huang, C.~Zhang, C.~Deng, C.~Deng, C.~Lin, C.~Yuan, C.~Li, C.~Gou, C.~Lou, C.~Wei, C.~Liu, C.~Li, D.~Zhu, D.~Zhong, F.~Li, F.~Zhang, G.~Wu, G.~Li, G.~Xiao, H.~Lin, H.~Yang, H.~Wang, H.~Ji, H.~Hao, H.~Shen, H.~Li, J.~Li, J.~Wu, J.~Zhu, J.~Jiao, J.~Feng, J.~Chen, J.~Duan, J.~Liu, J.~Zeng, J.~Tang, J.~Sun, J.~Chen, J.~Long, J.~Feng, J.~Zhan, J.~Fang, J.~Lu, K.~Hua, K.~Liu, K.~Shen, K.~Zhang, K.~Shen, K.~Wang, K.~Pan, K.~Zhang, K.~Li, L.~Li, L.~Li, L.~Shi, L.~Han, L.~Xiang, L.~Chen, L.~Chen, L.~Li, L.~Yan, L.~Chi, L.~Liu, M.~Du, M.~Wang, N.~Pan, P.~Chen, P.~Chen, P.~Wu, Q.~Yuan,
  Q.~Shuai, Q.~Tao, R.~Zheng, R.~Zhang, R.~Zhang, R.~Wang, R.~Yang, R.~Zhao, S.~Xu, S.~Liang, S.~Yan, S.~Zhong, S.~Cao, S.~Wu, S.~Liu, S.~Chang, S.~Cai, T.~Ao, T.~Yang, T.~Zhang, W.~Zhong, W.~Jia, W.~Weng, W.~Yu, W.~Huang, W.~Zhu, W.~Yang, W.~Wang, X.~Long, X.~Yin, X.~Li, X.~Zhu, X.~Jia, X.~Zhang, X.~Liu, X.~Zhang, X.~Yang, X.~Luo, X.~Chen, X.~Zhong, X.~Xiao, X.~Li, Y.~Wu, Y.~Wen, Y.~Du, Y.~Zhang, Y.~Ye, Y.~Wu, Y.~Liu, Y.~Yue, Y.~Zhou, Y.~Yuan, Y.~Xu, Y.~Yang, Y.~Zhang, Y.~Fang, Y.~Li, Y.~Ren, Y.~Xiong, Z.~Hong, Z.~Wang, Z.~Sun, Z.~Wang, Z.~Cai, Z.~Zha, Z.~An, Z.~Zhao, Z.~Xu, Z.~Chen, Z.~Wu, Z.~Zheng, Z.~Wang, Z.~Huang, Z.~Zhu, and Z.~Song.
\newblock Seed1.5-vl technical report.
\newblock \emph{CoRR}, abs/2505.07062, 2024.

\bibitem[Gupta et~al.(2023)Gupta, Yu, Sohn, Gu, Hahn, Fei-Fei, Essa, Jiang, and Lezama]{gupta2023photorealistic}
A.~Gupta, L.~Yu, K.~Sohn, X.~Gu, M.~Hahn, L.~Fei-Fei, I.~Essa, L.~Jiang, and J.~Lezama.
\newblock Photorealistic video generation with diffusion models.
\newblock \emph{arXiv preprint arXiv:2312.06662}, 2023.

\bibitem[He et~al.(2024)He, Jiang, Zhang, Ku, Soni, Siu, Chen, Chandra, Jiang, Arulraj, et~al.]{videoscore}
X.~He, D.~Jiang, G.~Zhang, M.~Ku, A.~Soni, S.~Siu, H.~Chen, A.~Chandra, Z.~Jiang, A.~Arulraj, et~al.
\newblock Videoscore: Building automatic metrics to simulate fine-grained human feedback for video generation.
\newblock \emph{arXiv preprint arXiv:2406.15252}, 2024.

\bibitem[Ho et~al.(2020)Ho, Jain, and Abbeel]{ho2020denoising}
J.~Ho, A.~Jain, and P.~Abbeel.
\newblock Denoising diffusion probabilistic models.
\newblock In \emph{NeurIPS}, 2020.

\bibitem[Ho et~al.(2022{\natexlab{a}})Ho, Chan, Saharia, Whang, Gao, Gritsenko, Kingma, Poole, Norouzi, Fleet, et~al.]{ho2022imagen}
J.~Ho, W.~Chan, C.~Saharia, J.~Whang, R.~Gao, A.~Gritsenko, D.~P. Kingma, B.~Poole, M.~Norouzi, D.~J. Fleet, et~al.
\newblock Imagen video: High definition video generation with diffusion models.
\newblock \emph{arXiv preprint arXiv:2210.02303}, 2022{\natexlab{a}}.

\bibitem[Ho et~al.(2022{\natexlab{b}})Ho, Salimans, Gritsenko, Chan, Norouzi, and Fleet]{vdm}
J.~Ho, T.~Salimans, A.~A. Gritsenko, W.~Chan, M.~Norouzi, and D.~J. Fleet.
\newblock Video diffusion models.
\newblock In \emph{NeurIPS}, 2022{\natexlab{b}}.

\bibitem[Huang et~al.(2024)Huang, He, Yu, Zhang, Si, Jiang, Zhang, Wu, Jin, Chanpaisit, Wang, Chen, Wang, Lin, Qiao, and Liu]{vbench}
Z.~Huang, Y.~He, J.~Yu, F.~Zhang, C.~Si, Y.~Jiang, Y.~Zhang, T.~Wu, Q.~Jin, N.~Chanpaisit, Y.~Wang, X.~Chen, L.~Wang, D.~Lin, Y.~Qiao, and Z.~Liu.
\newblock Vbench: Comprehensive benchmark suite for video generative models.
\newblock In \emph{{CVPR}}, pages 21807--21818. {IEEE}, 2024.

\bibitem[Jiang et~al.(2023)Jiang, Sablayrolles, Mensch, Bamford, Chaplot, de~Las~Casas, Bressand, Lengyel, Lample, Saulnier, Lavaud, Lachaux, Stock, Scao, Lavril, Wang, Lacroix, and Sayed]{mistral}
A.~Q. Jiang, A.~Sablayrolles, A.~Mensch, C.~Bamford, D.~S. Chaplot, D.~de~Las~Casas, F.~Bressand, G.~Lengyel, G.~Lample, L.~Saulnier, L.~R. Lavaud, M.~Lachaux, P.~Stock, T.~L. Scao, T.~Lavril, T.~Wang, T.~Lacroix, and W.~E. Sayed.
\newblock Mistral 7b.
\newblock \emph{CoRR}, abs/2310.06825, 2023.

\bibitem[Jiang et~al.(2024)Jiang, He, Zeng, Wei, Ku, Liu, and Chen]{MANTIS}
D.~Jiang, X.~He, H.~Zeng, C.~Wei, M.~Ku, Q.~Liu, and W.~Chen.
\newblock {MANTIS:} interleaved multi-image instruction tuning.
\newblock \emph{CoRR}, abs/2405.01483, 2024.

\bibitem[Ke et~al.(2021)Ke, Wang, Wang, Milanfar, and Yang]{MUSIQ}
J.~Ke, Q.~Wang, Y.~Wang, P.~Milanfar, and F.~Yang.
\newblock {MUSIQ:} multi-scale image quality transformer.
\newblock In \emph{{ICCV}}, pages 5128--5137. {IEEE}, 2021.

\bibitem[Kong et~al.(2024)Kong, Tian, Zhang, Min, Dai, Zhou, Xiong, Li, Wu, Zhang, et~al.]{hunyuanvideo}
W.~Kong, Q.~Tian, Z.~Zhang, R.~Min, Z.~Dai, J.~Zhou, J.~Xiong, X.~Li, B.~Wu, J.~Zhang, et~al.
\newblock Hunyuanvideo: A systematic framework for large video generative models.
\newblock \emph{arXiv preprint arXiv:2412.03603}, 2024.

\bibitem[Kou et~al.(2024)Kou, Liu, Zhang, Li, Wu, Min, Zhai, and Liu]{kou2024subjective}
T.~Kou, X.~Liu, Z.~Zhang, C.~Li, H.~Wu, X.~Min, G.~Zhai, and N.~Liu.
\newblock Subjective-aligned dateset and metric for text-to-video quality assessment.
\newblock \emph{arXiv preprint arXiv:2403.11956}, 2024.

\bibitem[Kuaishou.(2024)]{kling}
Kuaishou.
\newblock Kling ai.
\newblock \emph{https://klingai.kuaishou.com/}, 2024.

\bibitem[LAION-AI(2022)]{laion_aesthetic}
LAION-AI.
\newblock aesthetic-predictor, 2022.

\bibitem[Li et~al.(2024)Li, Zhang, Guo, Zhang, Li, Zhang, Zhang, Li, Liu, and Li]{llavaonevision}
B.~Li, Y.~Zhang, D.~Guo, R.~Zhang, F.~Li, H.~Zhang, K.~Zhang, Y.~Li, Z.~Liu, and C.~Li.
\newblock Llava-onevision: Easy visual task transfer.
\newblock \emph{ArXiv preprint}, abs/2408.03326, 2024.

\bibitem[Li et~al.(2023{\natexlab{a}})Li, Li, Savarese, and Hoi]{li2023blip}
J.~Li, D.~Li, S.~Savarese, and S.~Hoi.
\newblock Blip-2: Bootstrapping language-image pre-training with frozen image encoders and large language models.
\newblock In \emph{International conference on machine learning}, pages 19730--19742. PMLR, 2023{\natexlab{a}}.

\bibitem[Li et~al.(2023{\natexlab{b}})Li, Wang, He, Li, Wang, Liu, Wang, Xu, Chen, Luo, Wang, and Qiao]{MVBench}
K.~Li, Y.~Wang, Y.~He, Y.~Li, Y.~Wang, Y.~Liu, Z.~Wang, J.~Xu, G.~Chen, P.~Luo, L.~Wang, and Y.~Qiao.
\newblock Mvbench: A comprehensive multi-modal video understanding benchmark.
\newblock \emph{ArXiv}, abs/2311.17005, 2023{\natexlab{b}}.
\newblock URL \url{https://api.semanticscholar.org/CorpusID:265466214}.

\bibitem[Li et~al.(2023{\natexlab{c}})Li, Chu, Wu, Yuan, Liu, Zhang, Li, Feng, Ding, and Wang]{li2023videogen}
X.~Li, W.~Chu, Y.~Wu, W.~Yuan, F.~Liu, Q.~Zhang, F.~Li, H.~Feng, E.~Ding, and J.~Wang.
\newblock Videogen: A reference-guided latent diffusion approach for high definition text-to-video generation.
\newblock \emph{arXiv preprint arXiv:2309.00398}, 2023{\natexlab{c}}.

\bibitem[Li et~al.(2023{\natexlab{d}})Li, Zhu, Han, Hou, Guo, and Cheng]{amt}
Z.~Li, Z.~Zhu, L.~Han, Q.~Hou, C.~Guo, and M.~Cheng.
\newblock {AMT:} all-pairs multi-field transforms for efficient frame interpolation.
\newblock In \emph{{CVPR}}, pages 9801--9810. {IEEE}, 2023{\natexlab{d}}.

\bibitem[Lin et~al.(2023)Lin, Zhu, Ye, Ning, Jin, and Yuan]{Video-LLaVA}
B.~Lin, B.~Zhu, Y.~Ye, M.~Ning, P.~Jin, and L.~Yuan.
\newblock Video-llava: Learning united visual representation by alignment before projection.
\newblock \emph{ArXiv}, abs/2311.10122, 2023.
\newblock URL \url{https://api.semanticscholar.org/CorpusID:265281544}.

\bibitem[Lin et~al.(2024)Lin, Pathak, Li, Li, Xia, Neubig, Zhang, and Ramanan]{vqascore}
Z.~Lin, D.~Pathak, B.~Li, J.~Li, X.~Xia, G.~Neubig, P.~Zhang, and D.~Ramanan.
\newblock Evaluating text-to-visual generation with image-to-text generation.
\newblock In \emph{{ECCV} {(9)}}, volume 15067 of \emph{Lecture Notes in Computer Science}, pages 366--384. Springer, 2024.

\bibitem[Liu et~al.(2023{\natexlab{a}})Liu, Li, Wu, and Lee]{LLaVA}
H.~Liu, C.~Li, Q.~Wu, and Y.~J. Lee.
\newblock Visual instruction tuning.
\newblock \emph{ArXiv}, abs/2304.08485, 2023{\natexlab{a}}.
\newblock URL \url{https://api.semanticscholar.org/CorpusID:258179774}.

\bibitem[Liu et~al.(2024{\natexlab{a}})Liu, Li, Li, Li, Zhang, Shen, and Lee]{llava-next}
H.~Liu, C.~Li, Y.~Li, B.~Li, Y.~Zhang, S.~Shen, and Y.~J. Lee.
\newblock Llava-next: Improved reasoning, ocr, and world knowledge, January 2024{\natexlab{a}}.
\newblock URL \url{https://llava-vl.github.io/blog/2024-01-30-llava-next/}.

\bibitem[Liu et~al.(2022)Liu, Li, Wu, Chen, Shan, and Qie]{UMT}
Y.~Liu, S.~Li, Y.~Wu, C.~W. Chen, Y.~Shan, and X.~Qie.
\newblock Umt: Unified multi-modal transformers for joint video moment retrieval and highlight detection.
\newblock \emph{2022 IEEE/CVF Conference on Computer Vision and Pattern Recognition (CVPR)}, pages 3032--3041, 2022.
\newblock URL \url{https://api.semanticscholar.org/CorpusID:247627801}.

\bibitem[Liu et~al.(2023{\natexlab{b}})Liu, Li, Ren, Gao, Li, Chen, Sun, and Hou]{fetv}
Y.~Liu, L.~Li, S.~Ren, R.~Gao, S.~Li, S.~Chen, X.~Sun, and L.~Hou.
\newblock {FETV:} {A} benchmark for fine-grained evaluation of open-domain text-to-video generation.
\newblock In \emph{NeurIPS}, 2023{\natexlab{b}}.

\bibitem[Liu et~al.(2024{\natexlab{b}})Liu, Cun, Liu, Wang, Zhang, Chen, Liu, Zeng, Chan, and Shan]{evalcrafter}
Y.~Liu, X.~Cun, X.~Liu, X.~Wang, Y.~Zhang, H.~Chen, Y.~Liu, T.~Zeng, R.~Chan, and Y.~Shan.
\newblock Evalcrafter: Benchmarking and evaluating large video generation models.
\newblock In \emph{{CVPR}}, pages 22139--22149. {IEEE}, 2024{\natexlab{b}}.

\bibitem[Liu et~al.(2024{\natexlab{c}})Liu, Li, Liu, Wang, Ren, Li, Chen, Sun, and Hou]{tempcompass}
Y.~Liu, S.~Li, Y.~Liu, Y.~Wang, S.~Ren, L.~Li, S.~Chen, X.~Sun, and L.~Hou.
\newblock {T}emp{C}ompass: Do video {LLM}s really understand videos?
\newblock In L.-W. Ku, A.~Martins, and V.~Srikumar, editors, \emph{Findings of the Association for Computational Linguistics ACL 2024}, pages 8731--8772, 2024{\natexlab{c}}.

\bibitem[LumaLabs.(2024)]{luma1.6}
LumaLabs.
\newblock Dream machine.
\newblock \emph{https://lumalabs.ai/dream-machine,}, 2024.

\bibitem[Mialon et~al.(2023)Mialon, Fourrier, Swift, Wolf, LeCun, and Scialom]{gaia}
G.~Mialon, C.~Fourrier, C.~Swift, T.~Wolf, Y.~LeCun, and T.~Scialom.
\newblock Gaia: a benchmark for general ai assistants.
\newblock \emph{arXiv preprint arXiv:2311.12983}, 2023.

\bibitem[OpenAI(2023)]{GPT4}
OpenAI.
\newblock Gpt-4 technical report.
\newblock \emph{ArXiv}, abs/2303.08774, 2023.
\newblock URL \url{https://api.semanticscholar.org/CorpusID:266362871}.

\bibitem[OpenAI(2024)]{gpt4o}
OpenAI.
\newblock Gpt-4o system card, 2024.

\bibitem[Polyak et~al.(2024)Polyak, Zohar, Brown, Tjandra, Sinha, Lee, Vyas, Shi, Ma, Chuang, et~al.]{moviegen}
A.~Polyak, A.~Zohar, A.~Brown, A.~Tjandra, A.~Sinha, A.~Lee, A.~Vyas, B.~Shi, C.-Y. Ma, C.-Y. Chuang, et~al.
\newblock Movie gen: A cast of media foundation models.
\newblock \emph{arXiv preprint arXiv:2410.13720}, 2024.

\bibitem[Radford et~al.(2021)Radford, Kim, Hallacy, Ramesh, Goh, Agarwal, Sastry, Askell, Mishkin, Clark, Krueger, and Sutskever]{CLIP}
A.~Radford, J.~W. Kim, C.~Hallacy, A.~Ramesh, G.~Goh, S.~Agarwal, G.~Sastry, A.~Askell, P.~Mishkin, J.~Clark, G.~Krueger, and I.~Sutskever.
\newblock Learning transferable visual models from natural language supervision.
\newblock In \emph{{ICML}}, volume 139 of \emph{Proceedings of Machine Learning Research}, pages 8748--8763. {PMLR}, 2021.

\bibitem[Ramesh et~al.(2022)Ramesh, Dhariwal, Nichol, Chu, and Chen]{ramesh2022hierarchical}
A.~Ramesh, P.~Dhariwal, A.~Nichol, C.~Chu, and M.~Chen.
\newblock Hierarchical text-conditional image generation with clip latents.
\newblock \emph{arXiv preprint arXiv:2204.06125}, 2022.

\bibitem[Rombach et~al.(2022)Rombach, Blattmann, Lorenz, Esser, and Ommer]{rombach2022high}
R.~Rombach, A.~Blattmann, D.~Lorenz, P.~Esser, and B.~Ommer.
\newblock High-resolution image synthesis with latent diffusion models.
\newblock In \emph{CVPR}, 2022.

\bibitem[Runway.(2024)]{gen3}
Runway.
\newblock Gen-3.
\newblock \emph{https://runwayml.com/research/introducing-gen-3-alpha}, 2024.

\bibitem[Saharia et~al.(2022)Saharia, Chan, Saxena, Li, Whang, Denton, Ghasemipour, Gontijo~Lopes, Karagol~Ayan, Salimans, et~al.]{saharia2022photorealistic}
C.~Saharia, W.~Chan, S.~Saxena, L.~Li, J.~Whang, E.~L. Denton, K.~Ghasemipour, R.~Gontijo~Lopes, B.~Karagol~Ayan, T.~Salimans, et~al.
\newblock Photorealistic text-to-image diffusion models with deep language understanding.
\newblock In \emph{NeurIPS}, 2022.

\bibitem[Shangguan et~al.(2024)Shangguan, Li, Ding, Zheng, Zhao, Fitzgerald, and Cohan]{tomato}
Z.~Shangguan, C.~Li, Y.~Ding, Y.~Zheng, Y.~Zhao, T.~Fitzgerald, and A.~Cohan.
\newblock Tomato: Assessing visual temporal reasoning capabilities in multimodal foundation models.
\newblock \emph{arXiv preprint arXiv:2410.23266}, 2024.

\bibitem[Song et~al.(2021)Song, Sohl-Dickstein, Kingma, Kumar, Ermon, and Poole]{song2020score}
Y.~Song, J.~Sohl-Dickstein, D.~P. Kingma, A.~Kumar, S.~Ermon, and B.~Poole.
\newblock Score-based generative modeling through stochastic differential equations.
\newblock In \emph{ICLR}, 2021.

\bibitem[Team(2024)]{genmo2024mochi}
G.~Team.
\newblock Mochi 1.
\newblock \url{https://github.com/genmoai/models}, 2024.

\bibitem[Teed and Deng(2020)]{raft}
Z.~Teed and J.~Deng.
\newblock {RAFT:} recurrent all-pairs field transforms for optical flow.
\newblock In \emph{{ECCV} {(2)}}, volume 12347 of \emph{Lecture Notes in Computer Science}, pages 402--419. Springer, 2020.

\bibitem[Wang et~al.(2024{\natexlab{a}})Wang, Duan, Zhai, Wang, and Min]{AIGV-Assessor}
J.~Wang, H.~Duan, G.~Zhai, J.~Wang, and X.~Min.
\newblock Aigv-assessor: Benchmarking and evaluating the perceptual quality of text-to-video generation with lmm.
\newblock \emph{arXiv preprint arXiv:2411.17221}, 2024{\natexlab{a}}.

\bibitem[Wang et~al.(2024{\natexlab{b}})Wang, Wu, Jiang, Xun, Xiao, Guo, and Xiao]{wang2024world}
J.~Wang, B.~Wu, H.~Jiang, Z.~Xun, X.~Xiao, H.~Guo, and J.~Xiao.
\newblock World to code: Multi-modal data generation via self-instructed compositional captioning and filtering.
\newblock In \emph{Proceedings of the 2024 Conference on Empirical Methods in Natural Language Processing}, pages 4608--4623, 2024{\natexlab{b}}.

\bibitem[Wang et~al.(2024{\natexlab{c}})Wang, Bai, Tan, Wang, Fan, Bai, Chen, Liu, Wang, Ge, Fan, Dang, Du, Ren, Men, Liu, Zhou, Zhou, and Lin]{qwen2vl}
P.~Wang, S.~Bai, S.~Tan, S.~Wang, Z.~Fan, J.~Bai, K.~Chen, X.~Liu, J.~Wang, W.~Ge, Y.~Fan, K.~Dang, M.~Du, X.~Ren, R.~Men, D.~Liu, C.~Zhou, J.~Zhou, and J.~Lin.
\newblock Qwen2-vl: Enhancing vision-language model's perception of the world at any resolution.
\newblock \emph{arXiv preprint arXiv:2409.12191}, 2024{\natexlab{c}}.

\bibitem[Wang et~al.(2024{\natexlab{d}})Wang, He, Li, Li, Yu, Ma, Li, Chen, Chen, Wang, Luo, Liu, Wang, Wang, and Qiao]{viclip}
Y.~Wang, Y.~He, Y.~Li, K.~Li, J.~Yu, X.~Ma, X.~Li, G.~Chen, X.~Chen, Y.~Wang, P.~Luo, Z.~Liu, Y.~Wang, L.~Wang, and Y.~Qiao.
\newblock Internvid: {A} large-scale video-text dataset for multimodal understanding and generation.
\newblock In \emph{{ICLR}}. OpenReview.net, 2024{\natexlab{d}}.

\bibitem[Wu et~al.(2023)Wu, Zhang, Liao, Chen, Hou, Wang, Sun, Yan, and Lin]{dover}
H.~Wu, E.~Zhang, L.~Liao, C.~Chen, J.~Hou, A.~Wang, W.~Sun, Q.~Yan, and W.~Lin.
\newblock Exploring video quality assessment on user generated contents from aesthetic and technical perspectives.
\newblock In \emph{{ICCV}}, pages 20087--20097. {IEEE}, 2023.

\bibitem[Wu et~al.(2024{\natexlab{a}})Wu, Zhang, Zhang, Chen, Liao, Li, Gao, Wang, Zhang, Sun, Yan, Min, Zhai, and Lin]{q-align}
H.~Wu, Z.~Zhang, W.~Zhang, C.~Chen, L.~Liao, C.~Li, Y.~Gao, A.~Wang, E.~Zhang, W.~Sun, Q.~Yan, X.~Min, G.~Zhai, and W.~Lin.
\newblock Q-align: Teaching lmms for visual scoring via discrete text-defined levels.
\newblock In \emph{{ICML}}. OpenReview.net, 2024{\natexlab{a}}.

\bibitem[Wu et~al.(2024{\natexlab{b}})Wu, Zhu, Zhang, Zhang, Chen, Liao, Li, Wang, Sun, Yan, Liu, Zhai, Wang, and Lin]{co-instruct}
H.~Wu, H.~Zhu, Z.~Zhang, E.~Zhang, C.~Chen, L.~Liao, C.~Li, A.~Wang, W.~Sun, Q.~Yan, X.~Liu, G.~Zhai, S.~Wang, and W.~Lin.
\newblock Towards open-ended visual quality comparison.
\newblock In \emph{{ECCV} {(3)}}, volume 15061 of \emph{Lecture Notes in Computer Science}, pages 360--377. Springer, 2024{\natexlab{b}}.

\bibitem[Wu et~al.(2024{\natexlab{c}})Wu, Fang, Wu, Wang, Ge, Cun, Zhang, Liu, Gu, Zhao, Lin, Hsu, Shan, and Shou]{tvge}
J.~Z. Wu, G.~Fang, H.~Wu, X.~Wang, Y.~Ge, X.~Cun, D.~J. Zhang, J.~Liu, Y.~Gu, R.~Zhao, W.~Lin, W.~Hsu, Y.~Shan, and M.~Z. Shou.
\newblock Towards {A} better metric for text-to-video generation.
\newblock \emph{CoRR}, abs/2401.07781, 2024{\natexlab{c}}.

\bibitem[Xu et~al.(2016)Xu, Mei, Yao, and Rui]{MSR-VTT}
J.~Xu, T.~Mei, T.~Yao, and Y.~Rui.
\newblock {MSR-VTT:} {A} large video description dataset for bridging video and language.
\newblock In \emph{{CVPR}}, pages 5288--5296. {IEEE} Computer Society, 2016.

\bibitem[Yang et~al.(2024{\natexlab{a}})Yang, Yang, Hui, Zheng, Yu, Zhou, Li, Li, Liu, Huang, Dong, Wei, Lin, Tang, Wang, Yang, Tu, Zhang, Ma, Yang, Xu, Zhou, Bai, He, Lin, Dang, Lu, Chen, Yang, Li, Xue, Ni, Zhang, Wang, Peng, Men, Gao, Lin, Wang, Bai, Tan, Zhu, Li, Liu, Ge, Deng, Zhou, Ren, Zhang, Wei, Ren, Liu, Fan, Yao, Zhang, Wan, Chu, Liu, Cui, Zhang, Guo, and Fan]{qwen2}
A.~Yang, B.~Yang, B.~Hui, B.~Zheng, B.~Yu, C.~Zhou, C.~Li, C.~Li, D.~Liu, F.~Huang, G.~Dong, H.~Wei, H.~Lin, J.~Tang, J.~Wang, J.~Yang, J.~Tu, J.~Zhang, J.~Ma, J.~Yang, J.~Xu, J.~Zhou, J.~Bai, J.~He, J.~Lin, K.~Dang, K.~Lu, K.~Chen, K.~Yang, M.~Li, M.~Xue, N.~Ni, P.~Zhang, P.~Wang, R.~Peng, R.~Men, R.~Gao, R.~Lin, S.~Wang, S.~Bai, S.~Tan, T.~Zhu, T.~Li, T.~Liu, W.~Ge, X.~Deng, X.~Zhou, X.~Ren, X.~Zhang, X.~Wei, X.~Ren, X.~Liu, Y.~Fan, Y.~Yao, Y.~Zhang, Y.~Wan, Y.~Chu, Y.~Liu, Z.~Cui, Z.~Zhang, Z.~Guo, and Z.~Fan.
\newblock Qwen2 technical report.
\newblock \emph{CoRR}, abs/2407.10671, 2024{\natexlab{a}}.

\bibitem[Yang et~al.(2024{\natexlab{b}})Yang, Teng, Zheng, Ding, Huang, Xu, Yang, Hong, Zhang, Feng, et~al.]{yang2024cogvideox}
Z.~Yang, J.~Teng, W.~Zheng, M.~Ding, S.~Huang, J.~Xu, Y.~Yang, W.~Hong, X.~Zhang, G.~Feng, et~al.
\newblock Cogvideox: Text-to-video diffusion models with an expert transformer.
\newblock \emph{arXiv preprint arXiv:2408.06072}, 2024{\natexlab{b}}.

\bibitem[Yao et~al.(2024)Yao, Yu, Zhang, Wang, Cui, Zhu, Cai, Li, Zhao, He, Chen, Zhou, Zou, Zhang, Hu, Zheng, Zhou, Cai, Han, Zeng, Li, Liu, and Sun]{minicpm-v}
Y.~Yao, T.~Yu, A.~Zhang, C.~Wang, J.~Cui, H.~Zhu, T.~Cai, H.~Li, W.~Zhao, Z.~He, Q.~Chen, H.~Zhou, Z.~Zou, H.~Zhang, S.~Hu, Z.~Zheng, J.~Zhou, J.~Cai, X.~Han, G.~Zeng, D.~Li, Z.~Liu, and M.~Sun.
\newblock Minicpm-v: {A} {GPT-4V} level {MLLM} on your phone.
\newblock \emph{CoRR}, abs/2408.01800, 2024.

\bibitem[Ye et~al.(2024)Ye, Xu, Liu, Hu, Yan, Qian, Zhang, Huang, and Zhou]{mPLUG-Owl3}
J.~Ye, H.~Xu, H.~Liu, A.~Hu, M.~Yan, Q.~Qian, J.~Zhang, F.~Huang, and J.~Zhou.
\newblock mplug-owl3: Towards long image-sequence understanding in multi-modal large language models.
\newblock \emph{CoRR}, abs/2408.04840, 2024.

\bibitem[Yu et~al.(2023)Yu, Cheng, Sohn, Lezama, Zhang, Chang, Hauptmann, Yang, Hao, Essa, et~al.]{yu2023magvit}
L.~Yu, Y.~Cheng, K.~Sohn, J.~Lezama, H.~Zhang, H.~Chang, A.~G. Hauptmann, M.-H. Yang, Y.~Hao, I.~Essa, et~al.
\newblock Magvit: Masked generative video transformer.
\newblock In \emph{CVPR}, 2023.

\bibitem[Zeng et~al.(2024)Zeng, Yang, Chen, and Liu]{videogen-eval}
A.~Zeng, Y.~Yang, W.~Chen, and W.~Liu.
\newblock The dawn of video generation: Preliminary explorations with sora-like models.
\newblock \emph{CoRR}, abs/2410.05227, 2024.

\bibitem[Zhai et~al.(2023)Zhai, Mustafa, Kolesnikov, and Beyer]{siglip}
X.~Zhai, B.~Mustafa, A.~Kolesnikov, and L.~Beyer.
\newblock Sigmoid loss for language image pre-training.
\newblock In \emph{{ICCV}}, pages 11941--11952. {IEEE}, 2023.

\bibitem[Zhang et~al.(2024{\natexlab{a}})Zhang, Zhang, Li, Zeng, Yang, Zhang, Wang, Tan, Li, and Liu]{longva}
P.~Zhang, K.~Zhang, B.~Li, G.~Zeng, J.~Yang, Y.~Zhang, Z.~Wang, H.~Tan, C.~Li, and Z.~Liu.
\newblock Long context transfer from language to vision.
\newblock \emph{ArXiv preprint}, abs/2406.16852, 2024{\natexlab{a}}.

\bibitem[Zhang et~al.(2024{\natexlab{b}})Zhang, Wu, Li, Li, Ma, Liu, and Li]{llava-video}
Y.~Zhang, J.~Wu, W.~Li, B.~Li, Z.~Ma, Z.~Liu, and C.~Li.
\newblock Video instruction tuning with synthetic data.
\newblock \emph{CoRR}, abs/2410.02713, 2024{\natexlab{b}}.

\bibitem[Zhang et~al.(2024{\natexlab{c}})Zhang, Jia, Wu, Li, Chen, Zhou, Sun, Liu, Min, Lin, et~al.]{Q-Bench-Video}
Z.~Zhang, Z.~Jia, H.~Wu, C.~Li, Z.~Chen, Y.~Zhou, W.~Sun, X.~Liu, X.~Min, W.~Lin, et~al.
\newblock Q-bench-video: Benchmarking the video quality understanding of lmms.
\newblock \emph{arXiv preprint arXiv:2409.20063}, 2024{\natexlab{c}}.

\bibitem[Zhang et~al.(2024{\natexlab{d}})Zhang, Li, Sun, Jia, Min, Zhang, Li, Chen, Wang, Ji, et~al.]{ugvq}
Z.~Zhang, X.~Li, W.~Sun, J.~Jia, X.~Min, Z.~Zhang, C.~Li, Z.~Chen, P.~Wang, Z.~Ji, et~al.
\newblock Benchmarking aigc video quality assessment: A dataset and unified model.
\newblock \emph{arXiv preprint arXiv:2407.21408}, 2024{\natexlab{d}}.

\bibitem[Zhang et~al.(2024{\natexlab{e}})Zhang, Wu, Zhang, Zhai, and Lin]{qbench}
Z.~Zhang, H.~Wu, E.~Zhang, G.~Zhai, and W.~Lin.
\newblock Q-bench{\textdollar}{\^{}}+{\textdollar}+: {A} benchmark for multi-modal foundation models on low-level vision from single images to pairs.
\newblock \emph{{IEEE} Trans. Pattern Anal. Mach. Intell.}, 46\penalty0 (12):\penalty0 10404--10418, 2024{\natexlab{e}}.

\bibitem[Zhang et~al.(2024{\natexlab{f}})Zhang, Zhou, Li, Zhao, Liu, and Zhai]{QualityAssessmentSurvey}
Z.~Zhang, Y.~Zhou, C.~Li, B.~Zhao, X.~Liu, and G.~Zhai.
\newblock Quality assessment in the era of large models: {A} survey.
\newblock \emph{CoRR}, abs/2409.00031, 2024{\natexlab{f}}.

\bibitem[Zhao et~al.(2024)Zhao, Gu, Wu, Zhang, Liu, Wu, Keppo, and Shou]{MotionDirector}
R.~Zhao, Y.~Gu, J.~Z. Wu, D.~J. Zhang, J.~Liu, W.~Wu, J.~Keppo, and M.~Z. Shou.
\newblock Motiondirector: Motion customization of text-to-video diffusion models.
\newblock In \emph{{ECCV} {(56)}}, volume 15114 of \emph{Lecture Notes in Computer Science}, pages 273--290. Springer, 2024.

\bibitem[Zheng et~al.(2024)Zheng, Peng, Yang, Shen, Li, Liu, Zhou, Li, and You]{opensora}
Z.~Zheng, X.~Peng, T.~Yang, C.~Shen, S.~Li, H.~Liu, Y.~Zhou, T.~Li, and Y.~You.
\newblock Open-sora: Democratizing efficient video production for all.
\newblock \emph{arXiv preprint arXiv:2412.20404}, 2024.

\bibitem[Zhu et~al.(2024)Zhu, Lin, Ning, Yan, Cui, Wang, Pang, Jiang, Zhang, Li, Zhang, Li, Liu, and Yuan]{languagebind}
B.~Zhu, B.~Lin, M.~Ning, Y.~Yan, J.~Cui, H.~Wang, Y.~Pang, W.~Jiang, J.~Zhang, Z.~Li, C.~Zhang, Z.~Li, W.~Liu, and L.~Yuan.
\newblock Languagebind: Extending video-language pretraining to n-modality by language-based semantic alignment.
\newblock In \emph{{ICLR}}. OpenReview.net, 2024.

\end{thebibliography}
\bibliographystyle{abbrvnat}


\clearpage
\appendix

\section{Impact Statement}
\label{app:impact}
The evaluation of video generative models (VGMs) currently relies predominantly on human assessment or qualitative analysis. Our research demonstrates that multimodal large language models (MLLMs) could potentially serve as unified evaluators for AI-generated videos. The transition from human to automated evaluation frameworks would have significant dual implications: On the one hand, this shift would substantially reduce evaluation costs, thereby accelerating the development cycle of VGMs. On the other hand, if VGMs are optimized solely based on model feedback, there is a potential risk of divergence between the automated evaluators’ preferences and human values. Therefore, it essential to carefully scrutinize the correlation between automatic metrics and humans in the future research on AIGV evaluation.

\section{Related Work}
\subsection{Video Generative Models}
The emergence of diffusion models~\cite{ho2020denoising,song2020score} has revolutionized generative applications, spanning from image synthesis~\cite{saharia2022photorealistic,ramesh2022hierarchical,rombach2022high} to video generation~\cite{gupta2023photorealistic,yu2023magvit,ho2022imagen,vdm}. On the one hand, recent VGMs like SORA \cite{sora}, Kling \cite{kling}, and MovieGen \cite{moviegen} have achieved remarkable success in producing highly realistic videos that closely resemble real ones. On the other hand, the condition signal for video generation has been expanded from text (T2V) to RGB images (I2V) \cite{li2023videogen,chen2023videocrafter1} and other videos (V2V) \cite{MotionDirector}. This work specifically focuses on evaluating text-conditioned video generation.

\subsection{Multimodal Large Language Models}
Multimodal large language models \cite{gpt4o,qwen2vl,llavaonevision,li2023blip} integrate LLMs with additional perception modules, enabling them to comprehend information beyond the text modality. With advancements in LLM backbones and the increasing scale of multimodal training, MLLMs are showing promising performance in understanding real-world images \cite{LLaVA,llava-next,SPHINX-X,wang2024world} and videos \cite{llava-video,MVBench,Video-LLaVA}. However, understanding AI-generated visual content presents unique challenges compared to real images, and the application of MLLMs in this field is still in its early stages.

\subsection{AI-Generated Video Evaluation}
Despite the progress in VGMs, they still struggle to consistently generate high-quality videos, which necessitate a comprehensive and fine-grained evaluation. Initial evaluation approaches \cite{vbench,evalcrafter} employ off-the-shelf models to evaluate specific aspects, such as CLIP \cite{CLIP} for video-text alignment and DINO \cite{DINO} for frame-to-frame consistency. With the advancement of MLLMs, recent studies have incorporated these models in AIGV evaluation \cite{videoscore,q-align,AIGV-Assessor,tvge,videophy}. However, they are constrained to a fixed set of evaluation aspects and typically rely on human annotations to further fine-tune the MLLMs for AIGV evaluation. By contrast, in this work explores using MLLMs as unified evaluators for any AIGV aspect by leveraging their inherent vision-language understanding capabilities, eliminating the need for human annotations.

\section{More Details of Experimental Settings}
\label{app:more_exp_set}

\subsection{Evaluated Systems}
\label{app:evaluated_systems}
\paragraph{MLLMs.} We conduct zero-shot evaluations on 18 MLLMs, the detailed information of which is summarized in Tab.~\ref{tab:mllms}.

\paragraph{VideoScore-v1.1 \cite{videoscore}.} This model is a fine-tuned version of Mantis-Idefics2-8B \cite{MANTIS}, trained to imitate human rating scores. In their work, \citet{videoscore} gathered human ratings (on a discrete scale from 1 to 4) for five key evaluation aspects: \textit{Visual Quality}, \textit{Temporal Consistency}, \textit{Dynamic Degree}, \textit{Text-to-Video Alignment}, and \textit{Factual Consistency}. The VideoScore model is trained using a regression approach to predict these human rating scores. According to the definition of these aspects, we associate them with 8 out of the 15 aspects in \datasetname, as detailed in Tab.~\ref{tab:aspect_map}.

\paragraph{VBench \cite{vbench}.} VBench is a benchmark suite that comprehensively evaluates 16 aspects of AIGV using off-the-shelf models and specifically designed rules. As shown in Tab.~\ref{tab:aspect_map}, we identified an overlap of 6 aspects between VBench and our \datasetname, and we adopted the VBench metrics to evaluate these 6 aspects in \datasetname.

\paragraph{UMTScore \cite{fetv}.} UMTScore is designed to assess video–text alignment. It builds on the UMT-L/16 model \cite{UMT}, fine-tuned on the MSR-VTT dataset \cite{MSR-VTT} for video–text retrieval. Within \datasetname, we employ UMTScore to evaluate the \textit{Overall Alignment} aspect.

\paragraph{VIDEOCON-PHYSICS \cite{videophy}.} VIDEOCON-PHYSICS evaluates (1) whether a generated video adheres to physical commonsense and (2) whether it matches the given textual prompt. The model is initialized from VIDEOCON \cite{VideoCon} and fine-tuned on a dataset of AI-generated videos with binary human annotations covering these two aspects. In \datasetname, we use VIDEOCON-PHYSICS to evaluate both \textit{Motion Naturalness} and \textit{Overall Alignment}.

\paragraph{DOVER \cite{dover}.} DOVER assesses video quality from both technical and aesthetic perspectives. It is trained on DIVIDE-3k, a dataset of 3,590 real-world videos annotated with mean opinion scores (MOS). We adopt DOVER in \datasetname to evaluate \textit{Technical Quality} and \textit{Aesthetic Quality}.

\subsection{Evaluation Criteria for Single Video Rating}
\label{app:eval_criteria}
We rewrite the evaluation criteria for single video rating in Eq. \ref{eq:eq_criteria} as follows:
\begin{equation}
\mathcal{A}^{\text{single}} = 
\begin{cases} 
\mathbf{1}(\mathcal{S}_1>\mathcal{S}_2) & \text{if } \mathcal{P}=``\mathcal{V}_1 \text{better}" \\
\mathbf{1}(\mathcal{S}_1<\mathcal{S}_2) & \text{elif } \mathcal{P}=``\mathcal{V}_2 \text{better}" \\
f_c(\mathcal{S}_1|\beta)\cdot f_c(\mathcal{S}_2|\beta) & \text{elif } \mathcal{P}=``\text{same good}" \\
f^{'}_c(\mathcal{S}_1|\alpha)\cdot f_c(\mathcal{S}_2|\alpha) & \text{elif } \mathcal{P}=``\text{same bad}" \\
\end{cases}
\end{equation}
$f_c(\mathcal{S}|\beta)$ and $f'_c(\mathcal{S}|\alpha)$ exponentially decays from 1 to 0 when the predicted rating score $\mathcal{S}$ gradually deviates from ``same good'' and ``same bad'', respectively:
\begin{equation}
f_c(\mathcal{S}|\beta) = 
\begin{cases} 
1 & \text{if } \beta < \mathcal{S} <= 1\\
e^{-s\cdot(\beta-\mathcal{S})} & \text{elif } 0=<\mathcal{S} <= \beta
\end{cases}
\end{equation}

\begin{equation}
f'_c(\mathcal{S}|\alpha) = 
\begin{cases} 
1 & \text{if } 0 <= \mathcal{S} < \alpha\\
e^{-s\cdot(\mathcal{S}-\alpha)} & \text{elif } \alpha=<\mathcal{S} <= 1
\end{cases}
\end{equation}
where $s$ is the coefficient that controls the speed of decaying. The graph of $f_c(\mathcal{S}|\beta)$ and $f'_c(\mathcal{S}|\alpha)$ are illustrated in Fig.~\ref{fig:exp_decay_func}. An analysis of the impact of $\alpha$ and $\beta$ on the final result is conducted in Appendix \ref{app:effect_alpha_beta}.

\subsection{Methodology for Adapting Single Video Rating for Pairwise Comparison}
\label{app:s2p_adaptation_method}
We first independently rate the two videos using Eq. \ref{eq:single_video_rating} to obtain the rating scores $\mathcal{S}_1$ and $\mathcal{S}_2$. Based on these scores, we then make a choice from $\mathbf{O}$:
\begin{equation}
\label{eq:eq_2}
C = 
\begin{cases} 
f_1(\{\mathcal{V}_1 \text{better}, \mathcal{V}_2 \text{better}\}, \mathcal{S}_1, \mathcal{S}_2) & \text{if } |\mathcal{S}_1 - \mathcal{S}_2|>\tau \\
& \quad \text{or } \alpha<\mathcal{S}_1<\beta \\
& \quad \text{or } \alpha<\mathcal{S}_2<\beta \\
f_1(\{\text{same good}, \text{same bad}\}, \mathcal{S}_1, \mathcal{S}_2) & \text{else}
\end{cases}
\end{equation}
where
\begin{equation}
\label{eq:eq_3}
f_1(\{\mathcal{C}_1, \mathcal{C}_2\}, \mathcal{S}_1, \mathcal{S}_2) = 
\begin{cases} 
\mathcal{C}_1 & \text{if } \mathcal{S}_1 > \mathcal{S}_2 \\
\mathcal{C}_2 & \text{else}
\end{cases}
\end{equation}
\begin{equation}
\label{eq:eq_4}
f_2(\{\mathcal{C}_1, \mathcal{C}_2\}, \mathcal{S}_1, \mathcal{S}_2) = 
\begin{cases} 
\mathcal{C}_1 & \text{if } \mathcal{S}_1>\beta \quad\text{and}\quad \mathcal{S}_2>\beta \\
\mathcal{C}_2 & \text{elif} \quad \mathcal{S}_1<\alpha \quad\text{and}\quad \mathcal{S}_2<\alpha
\end{cases}
\end{equation}
In the above equations, $\alpha<\beta$ are the thresholds for ``bad'' and ``good'' videos, respectively. $\tau$ is a threshold of the difference between $\mathcal{S}_1$ and $\mathcal{S}_2$, which controls whether we should select from $\{``\mathcal{V}_1 \text{better}", ``\mathcal{V}_2 \text{better}"\}$ or $\{``\text{same good}", ``\text{same bad}"\}$. By default, we set $\alpha=0.4, \beta=0.8, \tau=0.05$.

\subsection{Annotation Interface}
\begin{figure*}[t]
\centering
\includegraphics[width=0.7\textwidth]{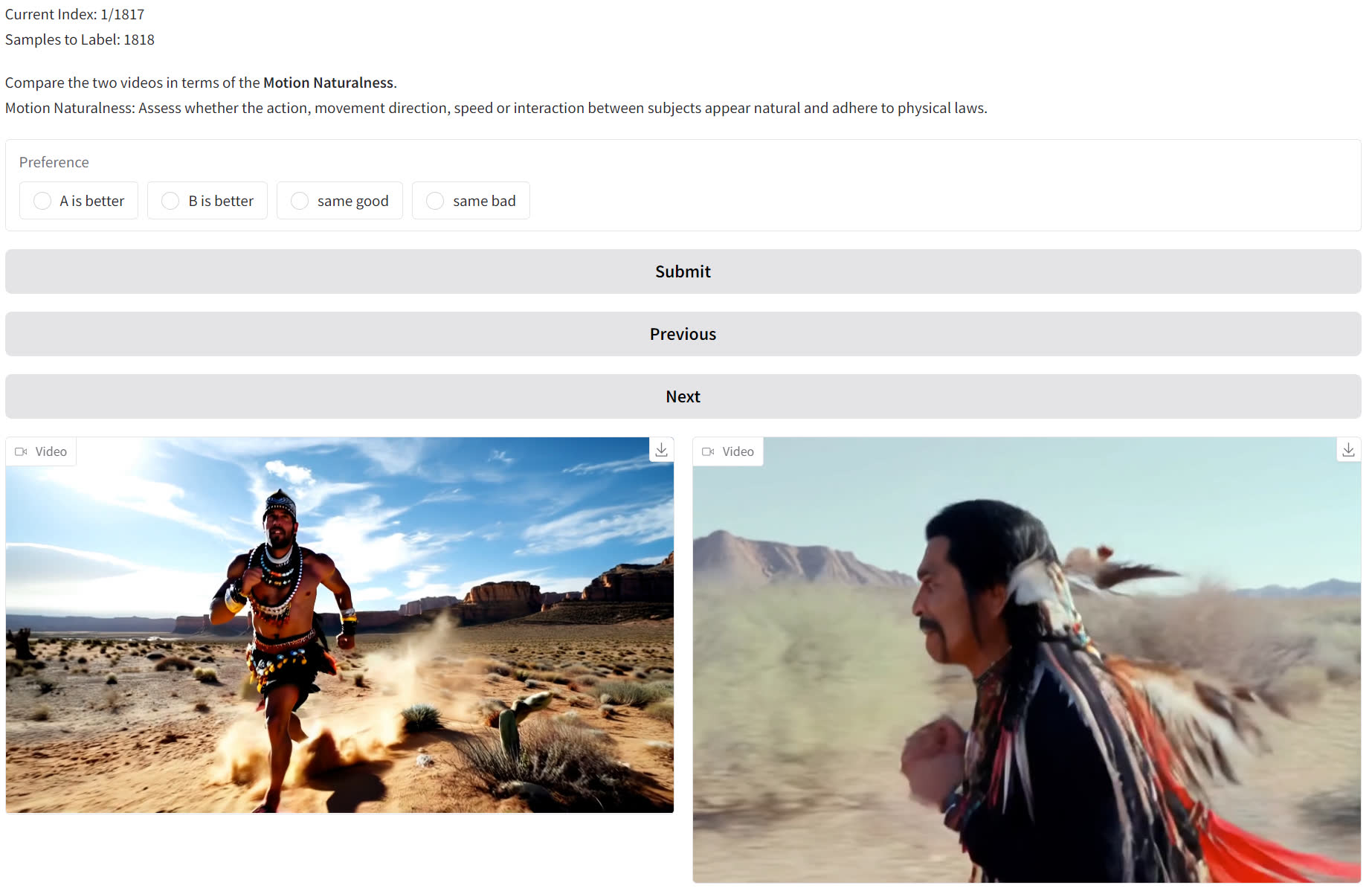}
\caption{Human baseline annotation interface.}
\label{fig:anno_interface}
\end{figure*}

Fig.~\ref{fig:anno_interface} presents the annotation interface used for establishing our human baseline. Annotators are shown a pair of videos along with aspect-specific evaluation instructions. Based on these, they indicate their preference by selecting one of four options: \{``A is better'', ``B is better'', ``same good'', ``same bad''\}.

\section{More Experimental Results}
\subsection{Effect of Prompting Strategy}
\label{app:effect_prompting_strategy}
\begin{figure*}[t]
\centering
  \subfigure[Single Video Rating with InternVL2.5-8B-MPO.]{
    \label{fig:exp_prompt_strategy_single_internvl_8b}
    \includegraphics[width=.45\textwidth]{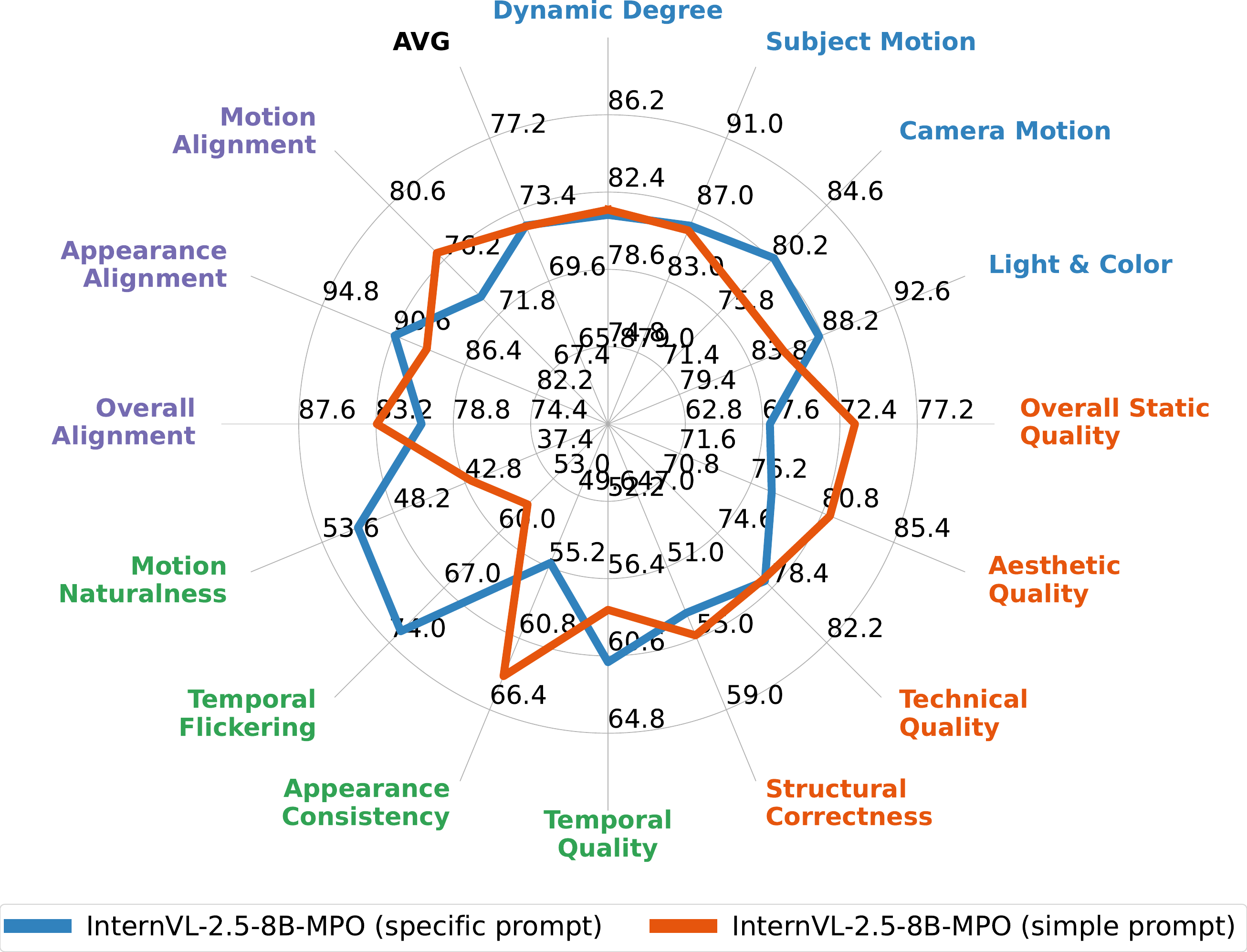}}
  \subfigure[Video Pair Comparison with InternVL2.5-8B-MPO.]{
    \label{fig:exp_prompt_strategy_pair_internvl_8b}
    \includegraphics[width=.45\textwidth]{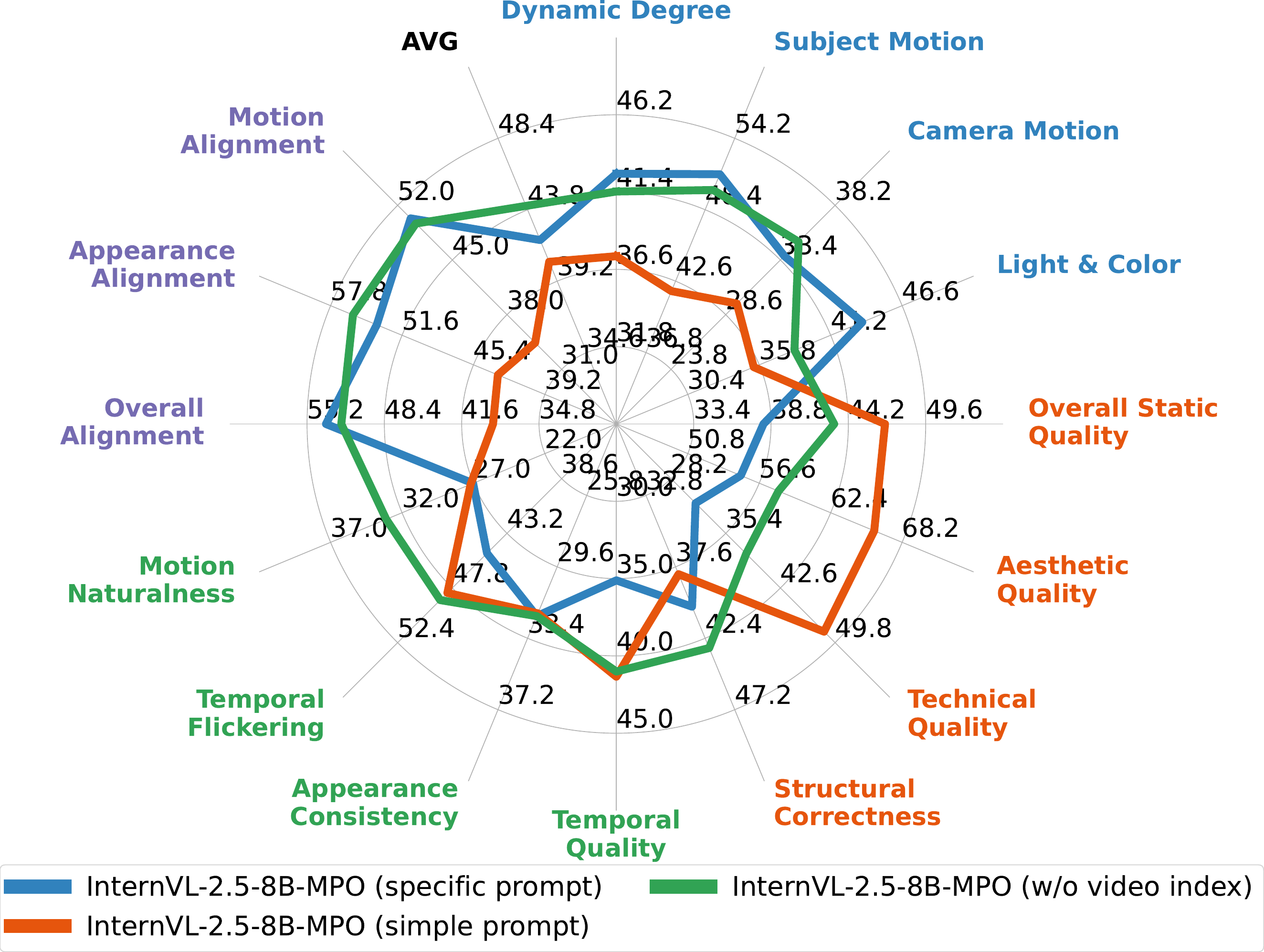}}
  \subfigure[Single Video Rating with Qwen2-VL-7B.]{
    \label{fig:exp_prompt_strategy_single_qwen2vl_7b}
    \includegraphics[width=.45\textwidth]{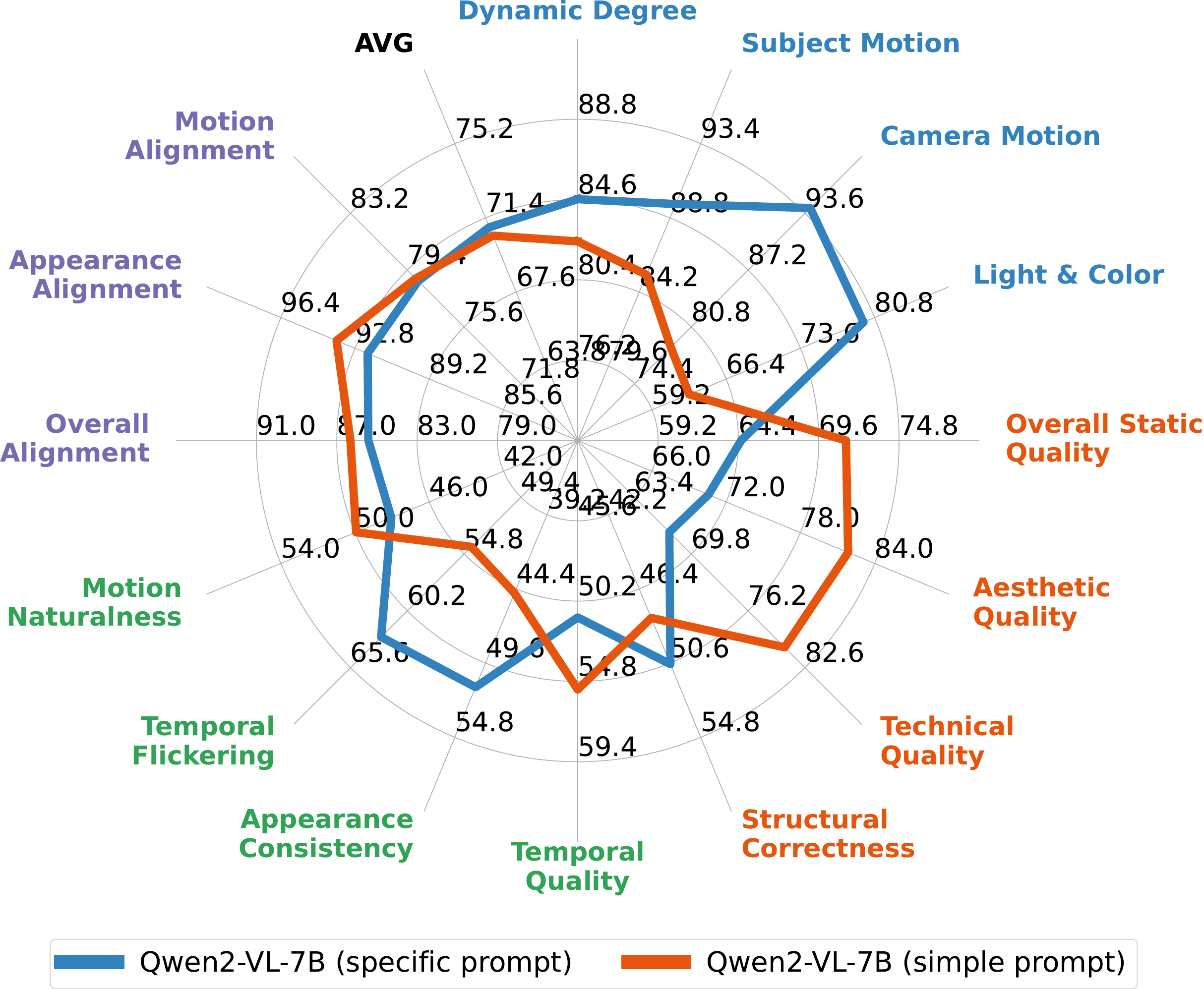}}
  \subfigure[Video Pair Comparison with Qwen2-VL-7B.]{
    \label{fig:exp_prompt_strategy_pair_qwen2vl_7b}
    \includegraphics[width=.45\textwidth]{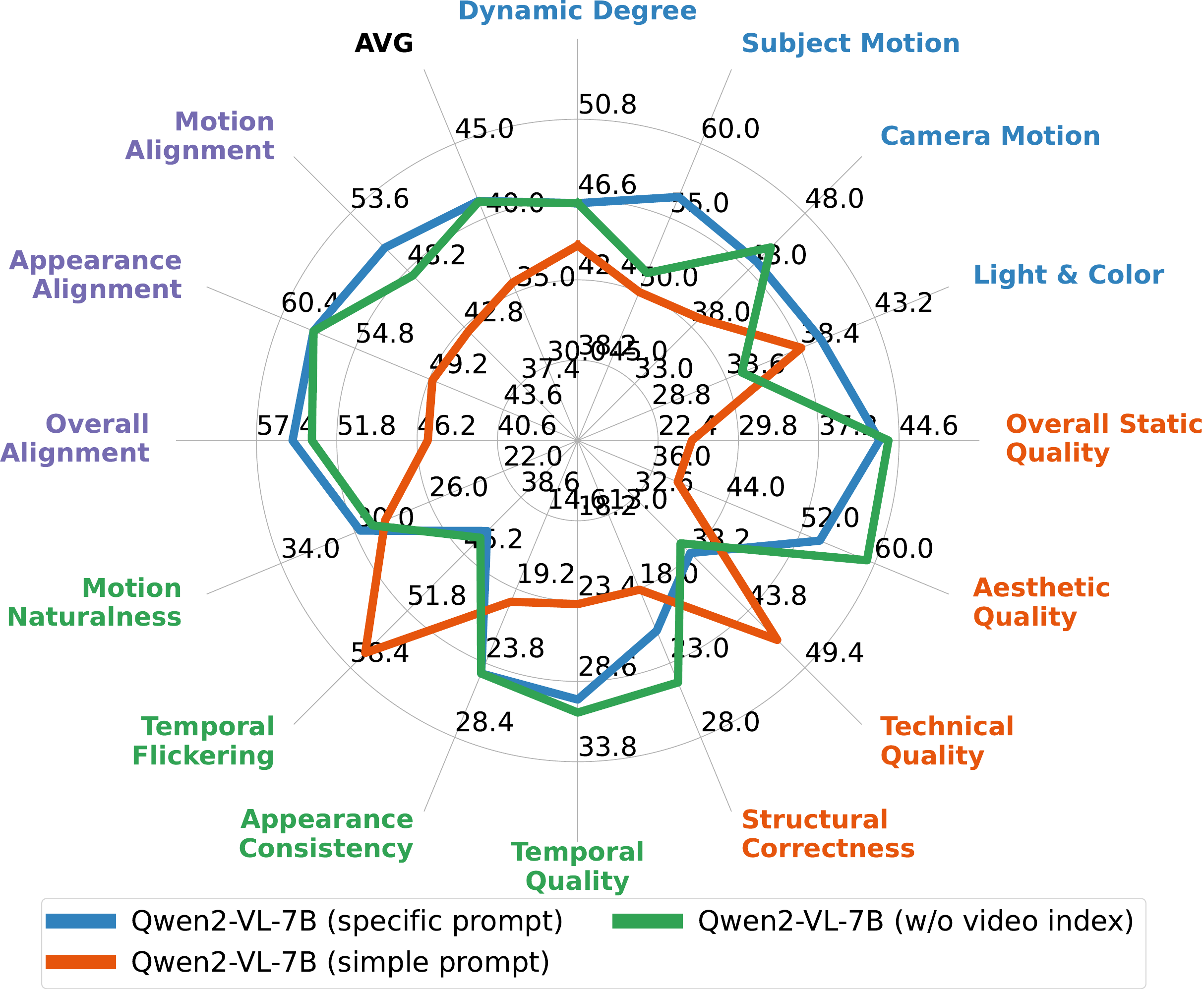}}
  \subfigure[Single Video Rating with Qwen2-VL-72B.]{
    \label{fig:exp_prompt_strategy_single_qwen2vl_72b}
    \includegraphics[width=.45\textwidth]{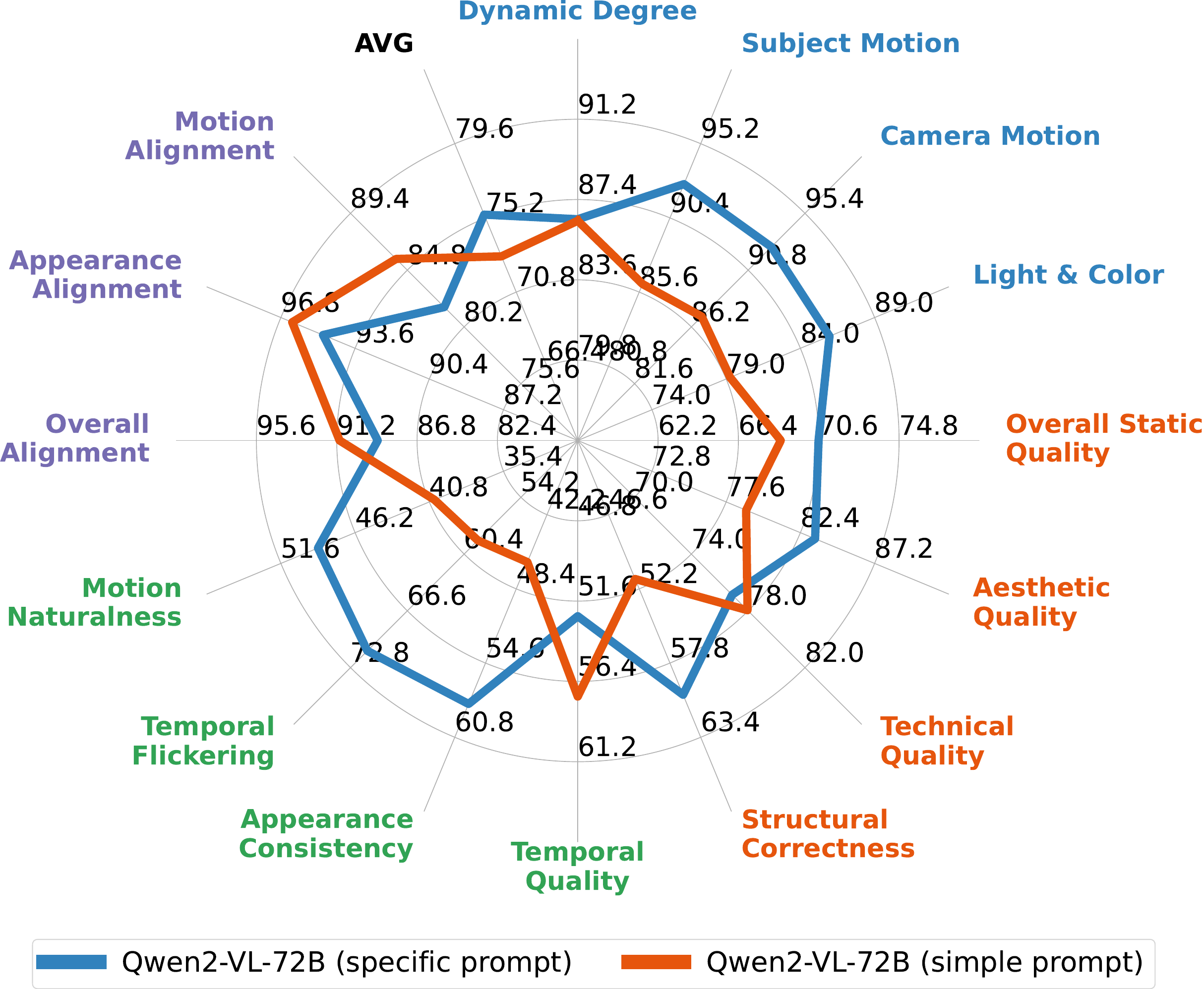}}
  \subfigure[Video Pair Comparison with Qwen2-VL-72B.]{
    \label{fig:exp_prompt_strategy_pair_qwen2vl_72b}
    \includegraphics[width=.45\textwidth]{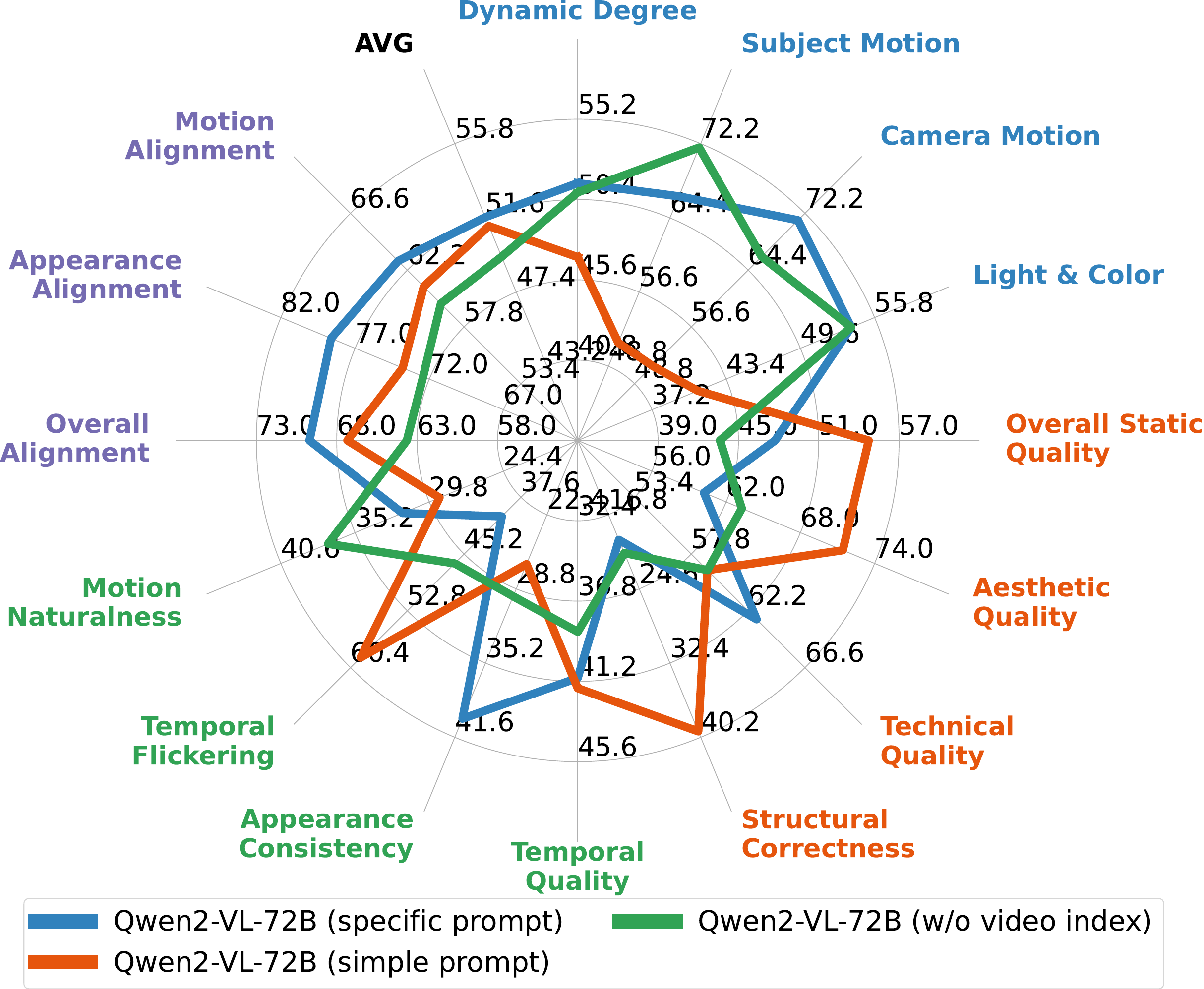}}
  \caption{Results of different prompting strategies with InternVL-2.5-8B-MPO, Qwen2-VL-7B and Qwen2-VL-72B.}
  \label{fig:app_prompt_strategy}
\end{figure*}

Fig.~\ref{fig:app_prompt_strategy} presents the results of different prompting strategy with InternVL-2.5-8B-MPO, Qwen2-VL-7B and Qwen2-VL-72B. Unlike InternVL-2.5-78B-MPO (previously presented in Fig.~\ref{fig:exp_prompt_strategy_internvl}), these three models showed a less pronounced advantage when using detailed aspect-specific descriptions, with the simplified prompt achieving better performance in several aspects. We conjecture that this is because these three models cannot comprehend the detailed description as effectively as InternVL-2.5-78B-MPO. Nevertheless, when considering overall performance, the full prompt consistently maintained a slight improvement over the simplified version across all three models. Consistent with our findings for InternVL-2.5-78B-MPO, removing video order indicator does not result in significant performance change.

\subsection{Effect of Scoring Strategy}
\label{app:effect_scoring_strategy}
\begin{figure*}[t]
  \centering
  \subfigure[Qwen2-VL-72B.]{
    \label{fig:exp_scoring_strategy_qwen2vl_72b}
    \includegraphics[width=.45\textwidth]{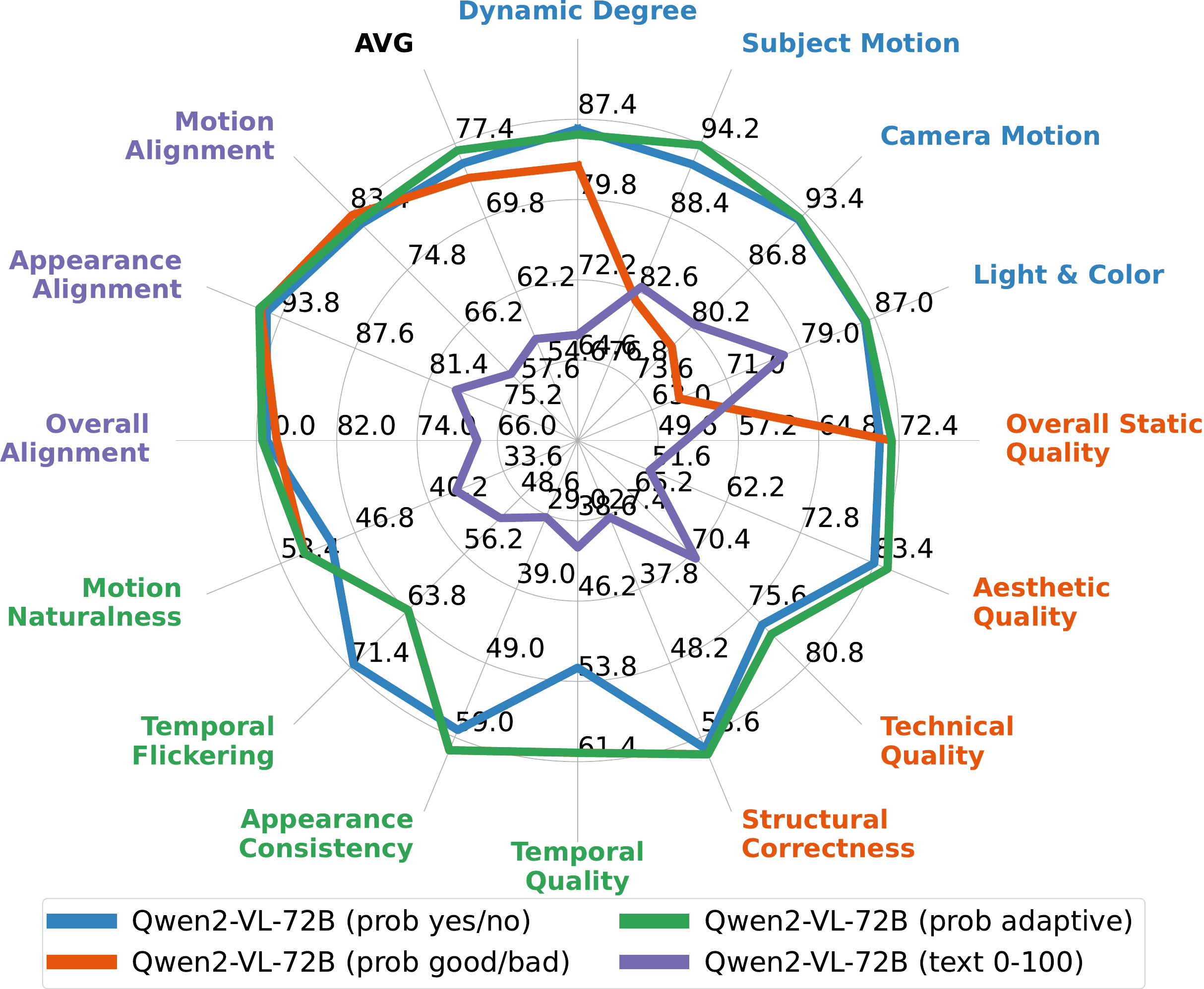}}
  \subfigure[InternVL-2.5-8B-MPO.]{
    \label{fig:exp_scoring_strategy_internvl_8b}
    \includegraphics[width=.45\textwidth]{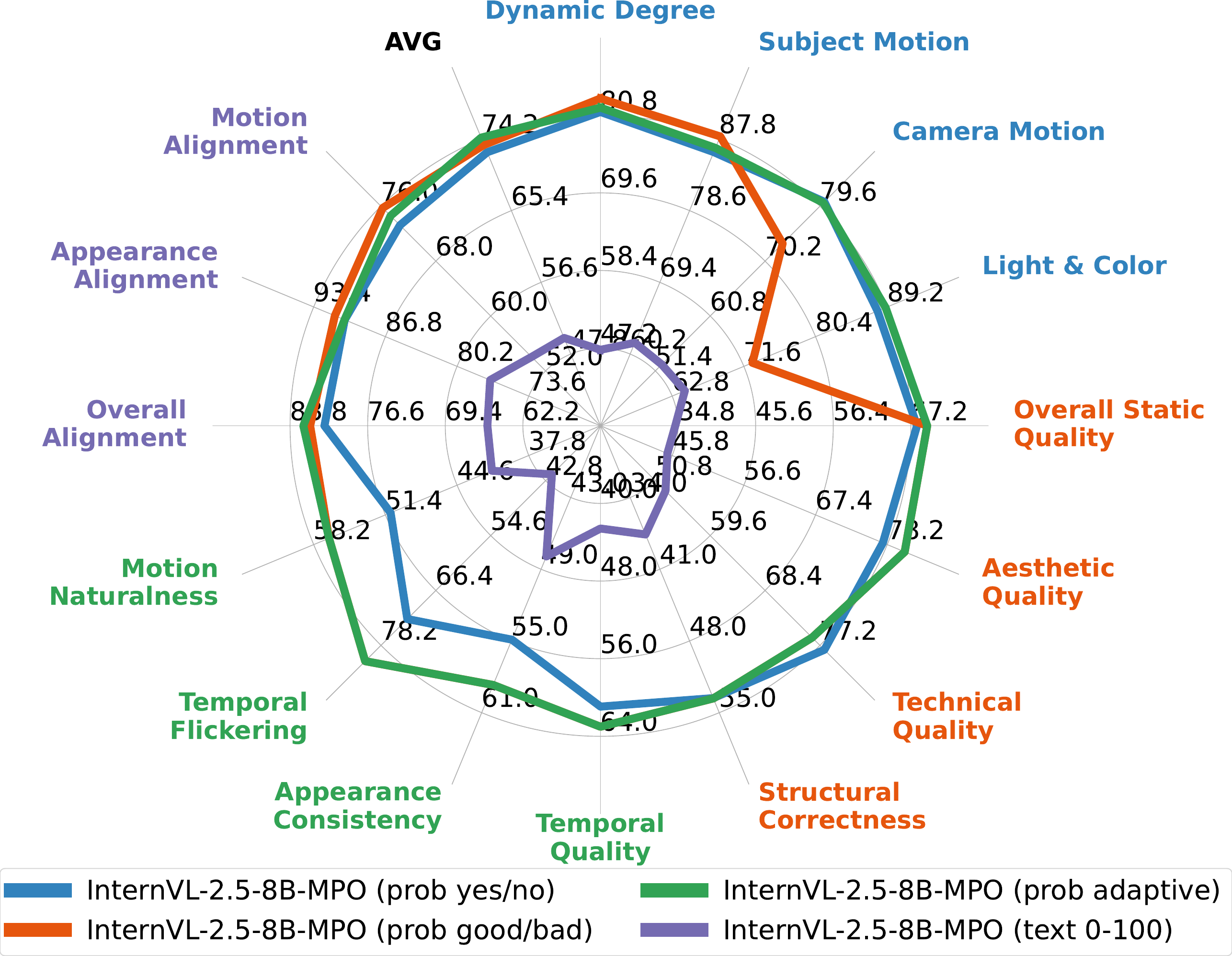}}
  \subfigure[InternVL-2.5-78B-MPO.]{
    \label{fig:exp_scoring_strategy_internvl_78b}
    \includegraphics[width=.45\textwidth]{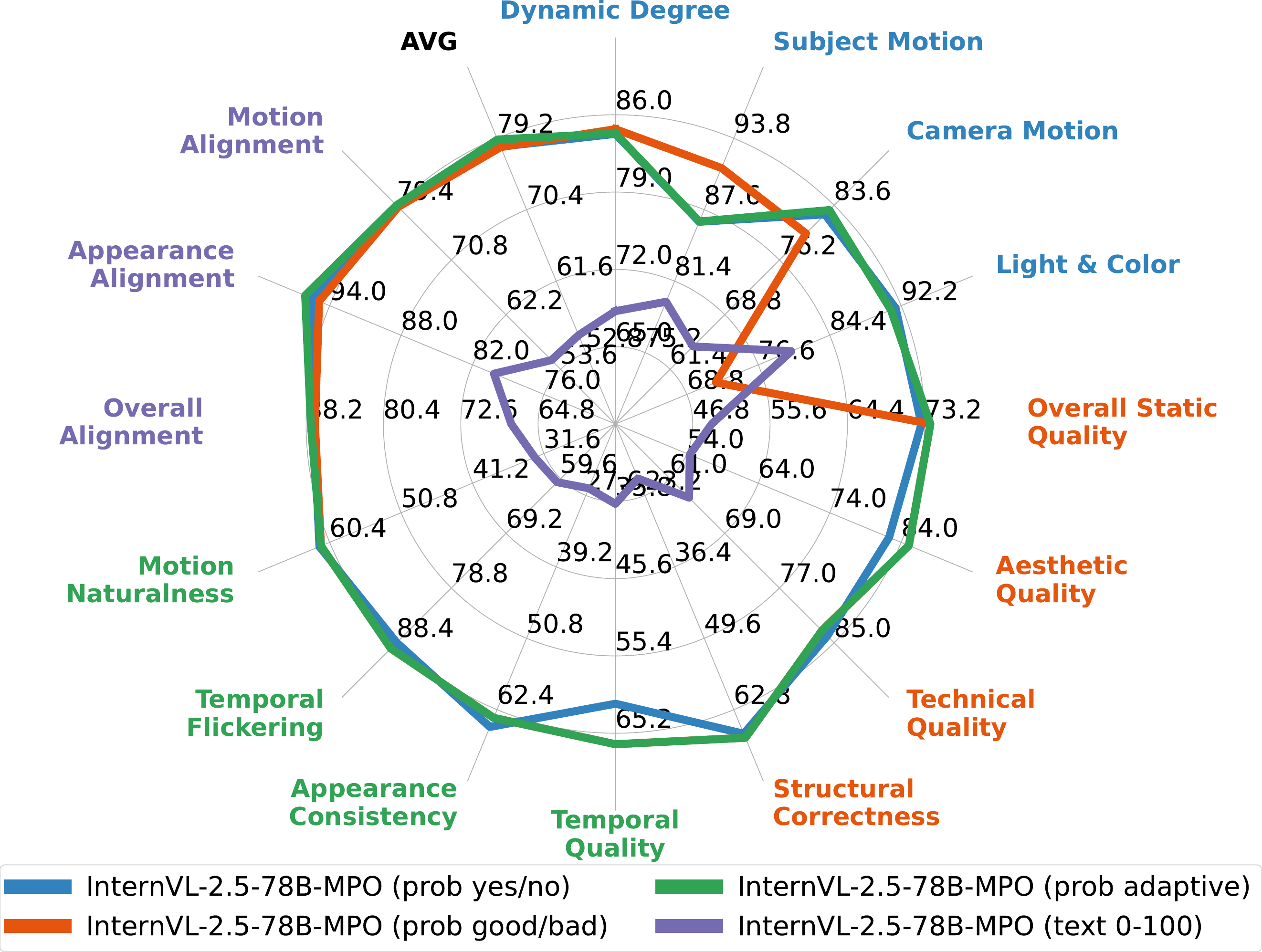}}
  \caption{Results of different scoring strategies for single video rating with Qwen2-VL-72B, InternVL-2.5-8B-MPO and InternVL-2.5-78B-MPO.}
  \label{fig:app_scoring_strategy}
\end{figure*}

In \S \ref{sec:effect_scoring_strategy}, we analyze the effect of different scoring strategies with Qwen2-VL-7B. Here, we extended the investigation to three additional models: Qwen2-VL-72B, InternVL-2.5-8B-MPO, and InternVL-2.5-78B. The results are presented in Fig.~\ref{fig:app_scoring_strategy}. Our findings reveal consistent patterns across all models. Similar to Qwen2-VL-7B, the \textit{good/bad} and \textit{yes/no} scoring strategies demonstrate distinct strengths in different evaluation aspects. Direct score generation through MLLMs consistently yields suboptimal results across all models. However, compared with Qwen2-VL-7B, the advantage of adaptive scoring token is less pronounced in Qwen2-VL-72B, InternVL-2.5-8B-MPO and InternVL-2.5-78B. This suggests that these three models are more robust to the variation of scoring tokens. These findings confirm that using \textit{yes/no} as unified scoring token is simple yet effective.

\subsection{Negative Prompt versus Positive Prompt}
\label{app:neg_prompt}
By default, we prompt MLLMs to assess whether a video is “good” in a specific aspect. As an alternative, we can also frame the question in a negative form—asking whether a video is “bad” in that aspect. To examine the impact of this change, we modify the evaluation prompts accordingly and invert the positive and negative token assignments when computing single video ratings in Eq.~\ref{eq:single_video_rating}.

We then evaluate the Qwen2-VL-7B model using both positive and negative prompts across four aspects. As shown in Table~\ref{tab:neg_pos_prompt}, the results reveal clear differences across dimensions. For \textit{Static Quality} and \textit{Temporal Quality}, the performance gap between positive and negative prompts is relatively small. In contrast, for \textit{Dynamic Degree} and \textit{Video-Text Alignment}, positive prompts yield substantially higher performance. These findings suggest that positive prompts provide more stable and reliable evaluations. Therefore, we adopt positive prompts as the default configuration in our experiments.

\subsection{Adapting Single Ratings for Pairwise Comparison}
\label{app:s2p_results}
While our previous experiments involve direct pairwise video comparisons by simultaneously feeding two videos into MLLMs, we now explore an alternative approach. This method first independently rates each video using Eq. \ref{eq:single_video_rating} and then converts these individual ratings into a four-way selection from set $\mathbf{O}$. The detailed conversion methodology is discussed in Appendix \ref{app:s2p_adaptation_method}.

As shown in Fig.~\ref{fig:exp_qwen2vl_s2p}, adaptation from single video rating brings substantial improvement to Qwen2-VL-7B compared to directly performing video pair comparison. However, Qwen2-VL-72B shows only minimal improvements under the same approach. Similar patterns were observed with InternVL-2.5-MPO, as shown in Appendix \ref{app:s2p_results}. We hypothesize that smaller-scale models, like the 7B variant, have inherently weaker pairwise comparison capabilities, and adapting from single video rating bypasses this limitation. Based on these findings, we recommend implementing the single-to-pairwise adaptation strategy particularly for MLLMs with limited pairwise comparison capabilities.

\begin{figure*}[t]
\centering
\includegraphics[width=0.5\textwidth]{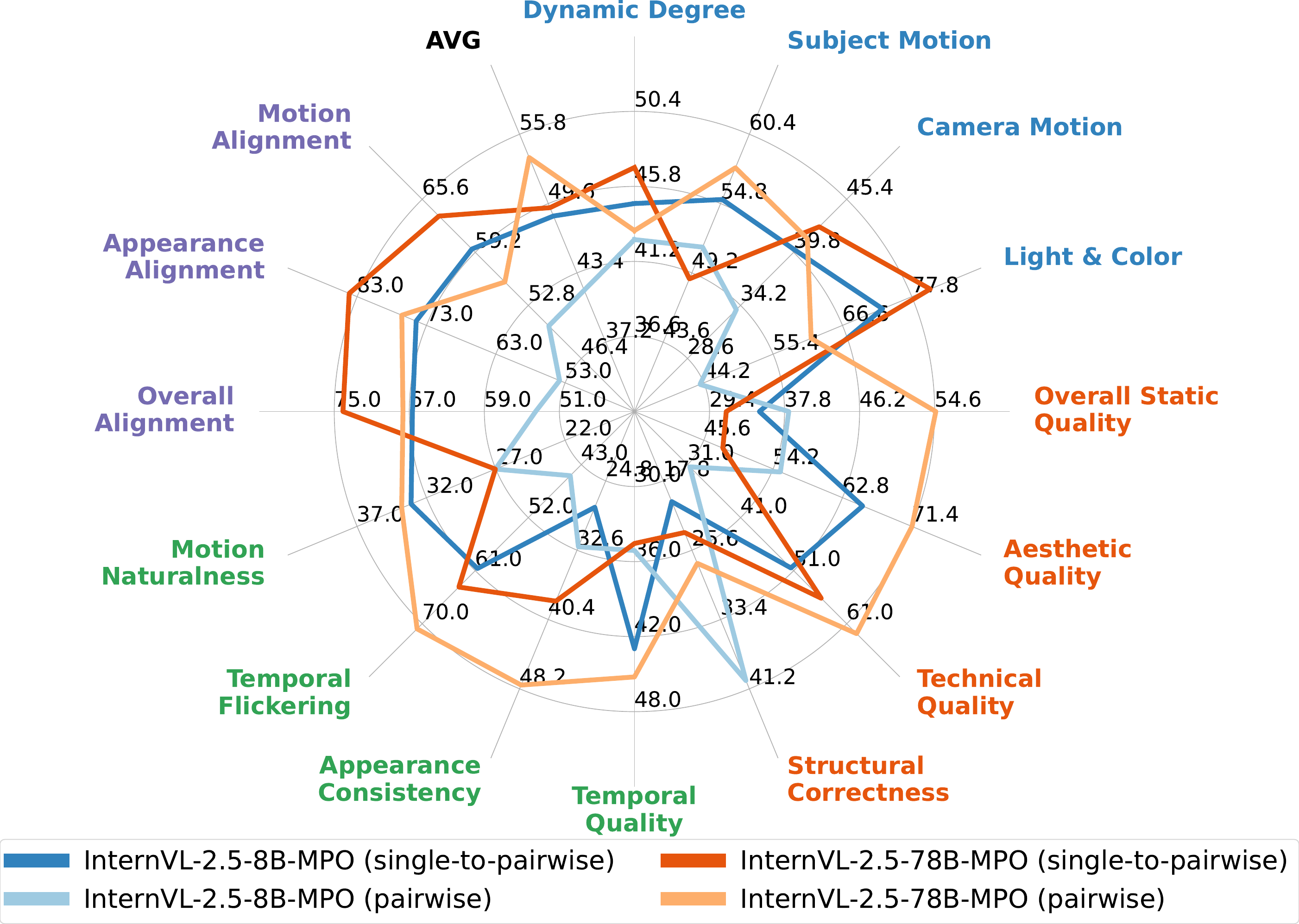}
\caption{Results of direct video pair comparison versus adapting single video rating to pairwise comparison, using InternVL-2.5-MPO.}
\label{fig:app_internvl2.5_s2p}
\end{figure*}

The effectiveness of adapting single video ratings for video pair comparisons using InternVL-2.5 models is illustrated in Fig.~\ref{fig:app_internvl2.5_s2p}. The results aligns with our observations with Qwen2-VL (shown in Fig.~\ref{fig:exp_qwen2vl_s2p}), revealing a model size-dependent pattern. While this adaptation strategy shows no benefit when applied to the 78B-scale model, it yields substantial improvements for the 8B-scale model.

\subsection{Effect of $\alpha$ and $\beta$}
\label{app:effect_alpha_beta}
We have chosen default values of $\alpha=0.4$ and $\beta=0.8$ to represent the thresholds for categorizing videos as ``bad'' and ``good'', respectively. To investigate the impact of these two parameters on single video rating evaluation, we conduct a sensitivity analysis. Specifically, we report the performance of four MLLMs across different values of $\alpha$ and $\beta$. As shown in Tab.~\ref{tab:effect_alpha_beta_a} and Tab.~\ref{tab:effect_alpha_beta_b}, adjusting these thresholds reveals notable trends in performance: (1) Lowering $\beta$ (i.e., making the criteria for ``good'' more lenient) leads to increased agreement between MLLMs and humans in the ``same good'' category, resulting in higher performance.
(2) Increasing $\alpha$ (i.e., making the criteria for ``bad'' stricter) also improves performance by enhancing alignment in the ``same bad'' category. However, despite the observed variations in performance as these parameters change, it is important to note that the relative performance of different MLLMs remains stable.

\subsection{Case Studies}
\label{app:case_study}
Tab.~\ref{tab:example_subject_motion}, \ref{tab:example_camera_motion}, \ref{tab:example_light_change}, \ref{tab:example_technical_quality}, \ref{tab:example_aesthetic_quality}, \ref{tab:example_structural_correctness}, \ref{tab:example_appearance_consistency}, \ref{tab:example_flickering}, \ref{tab:example_motion_naturalness}, \ref{tab:example_appearance_fine}, \ref{tab:example_motion_fine} showcase \datasetname data examples of different subaspects, along with the evaluation results by InternVL-2.5-8B-MPO and InternVL-2.5-78B-MPO.

\begin{table}[t]
    \centering
    \caption{Summary of MLLMs evaluated in the experiments.}
    \resizebox{\textwidth}{!}{$
    \begin{tabular}{lcccc}
         \toprule
         Model & Model Size & LLM & Vision Encoder &  Frames per Video \\
         \midrule
         Video-LLaVA \cite{Video-LLaVA}        & 7B  & \texttt{Vicuna-1.5} \cite{Vicuna}  & \texttt{LanguageBind} \cite{languagebind} &  8   \\
         LongVA-7B-DPO \cite{longva} & 7B & \texttt{Qwen2} \cite{qwen2} & \texttt{CLIP-ViT-L-336px} \cite{CLIP} & 16 \\
         ShareGPT4Video \cite{ShareGPT4Video} & 8B & \texttt{LLaMA3} \cite{llama3} & \texttt{CLIP-ViT-L-336px} \cite{CLIP} & 16 \\
         VideoLLaMA2.1 \cite{videollama2} & 7B & \texttt{Qwen2} \cite{qwen2} & \texttt{SigLip-400M} \cite{siglip} & 16 \\
         mPLUG-Owl3-7B \cite{mPLUG-Owl3} & 7B & \texttt{Qwen2} \cite{qwen2} & \texttt{SigLip-400M} \cite{siglip} & 16 \\
         VideoChat2-Mistral \cite{MVBench} & 7B & \texttt{Mistral} \cite{mistral} & \texttt{UMT-L/16} \cite{UMT} & 16 \\
         MiniCPM-V-2.6 \cite{minicpm-v} & 7B & \texttt{Qwen2} \cite{qwen2} & \texttt{SigLip-400M} \cite{siglip} & 16 \\
         LLaVA-OneVision-7B \cite{llavaonevision} & 7B & \texttt{Qwen2} \cite{qwen2} & \texttt{SigLip-400M} \cite{siglip} & 12,16 \\
         LLaVA-OneVision-72B \cite{llavaonevision} & 72B & \texttt{Qwen2} \cite{qwen2} & \texttt{SigLip-400M} \cite{siglip} & 12,16 \\
         LLaVA-Video-7B \cite{llava-video} & 7B & \texttt{Qwen2} \cite{qwen2} & \texttt{SigLip-400M} \cite{siglip} & 12,16 \\
         LLaVA-Video-72B \cite{llava-video} & 72B & \texttt{Qwen2} \cite{qwen2} & \texttt{SigLip-400M} \cite{siglip} & 12,16 \\
         Qwen2-VL-7B \cite{qwen2vl} & 7B & - & - & 12,16 \\
         Qwen2-VL-72B \cite{qwen2vl} & 72B & - & - & 12,16 \\
         InternVL-2.5-8B \cite{internvl2.5} & 8B & \texttt{InternLM2.5} \cite{internlm2.5} & \texttt{InternViT} \cite{internvl} & 12,16 \\
         InternVL-2.5-78B \cite{internvl2.5} & 78B & \texttt{InternLM2.5} \cite{internlm2.5} & \texttt{InternViT} \cite{internvl} & 12,16 \\
         GPT-4o-2024-08-06 \cite{gpt4o} & - & - & - & 12,16 \\
         Seed1.5-VL \cite{seed1.5-vl} & 20B Act. & Seed1.5-LLM & Seed-ViT & 12,16 \\
         Gemini2.5-Flash \cite{gemini-2.5} & - & - & - & 12 \\
         \bottomrule
    \end{tabular}
    $}
    \label{tab:mllms}
\end{table}
\begin{table}[t]
    \centering
    \caption{Association between VideoScore and VBench aspects with \datasetname aspects.}
    \resizebox{\textwidth}{!}{$
    \begin{tabular}{lccc}
         \toprule
         Original Aspect & Definition & Evaluation Model & \datasetname Aspect \\
         \midrule
         \textcolor{lightgray}{\textit{VideoScore}} \\
         Visual Quality &  {\makecell{\textit{the quality of the video in terms of clearness,}\\ \textit{resolution, brightness, and color}}} & VideoScore-v1.1 & {\makecell{Overall Static Quality\\Aesthetic Quality\\Technical Quality}} \\ \\
         Temporal Consistency &  \textit{the consistency of objects or humans in video} & VideoScore-v1.1 & Appearance Consistency \\ \\
         Dynamic Degree &  \textit{the degree of dynamic changes} & VideoScore-v1.1 & Overall Dynamic Degree \\ \\
         Text-to-Video Alignment &  \textit{the alignment between the text prompt and the video content} & VideoScore-v1.1 & Overall Alignment \\ \\
         Factual Consistency &  {\makecell{\textit{the consistency of the video content with} \\ \textit{the common-sense and factual knowledge}}} & VideoScore-v1.1 & {\makecell{Structural Correctness\\Motion Naturalness}} \\
         \midrule
         \textcolor{lightgray}{\textit{VBench}} \\
         Overall Consistency &  \textit{overall video-text consistency} & ViCLIP \cite{viclip} & Overall Alignment \\ \\
         Motion Smoothness &  {\makecell{\textit{whether the motion in the generated video is smooth}\\ \textit{and follows the physical law of the real world}}} & AMT \cite{amt} & Motion Naturalness \\ \\
         Aesthetic Quality &  {\makecell{\textit{reflect aesthetic aspects such as the layout,} \\ \textit{the richness and harmony of colors, the photo-realism,} \\ \textit{naturalness, and artistic quality of the video frames}}} &  {\makecell{LAION aesthetic predictor\\ \cite{laion_aesthetic} }} & Aesthetic Quality \\ \\ \\
         Imaging Quality &  {\makecell{\textit{refers to the distortion (e.g., over-exposure, noise, blur)} \\ \textit{presented in the generated frames}}} & MUSIQ \cite{MUSIQ} & Technical Quality \\ \\
         Dynamic Degree &  \textit{the degree of dynamics (i.e., whether it contains large motions)} & RAFT \cite{raft} & Overall Dynamic Degree \\ \\
         Subject Consistency &  {\makecell{\textit{whether the subject appearance remains}\\ \textit{consistent throughout the whole video}}} & DINO \cite{DINO} & Appearance Consistency \\ \\
         \bottomrule
    \end{tabular}
    $}
    \label{tab:aspect_map}
\end{table}
\begin{figure*}[t]
\centering
\includegraphics[width=0.7\textwidth]{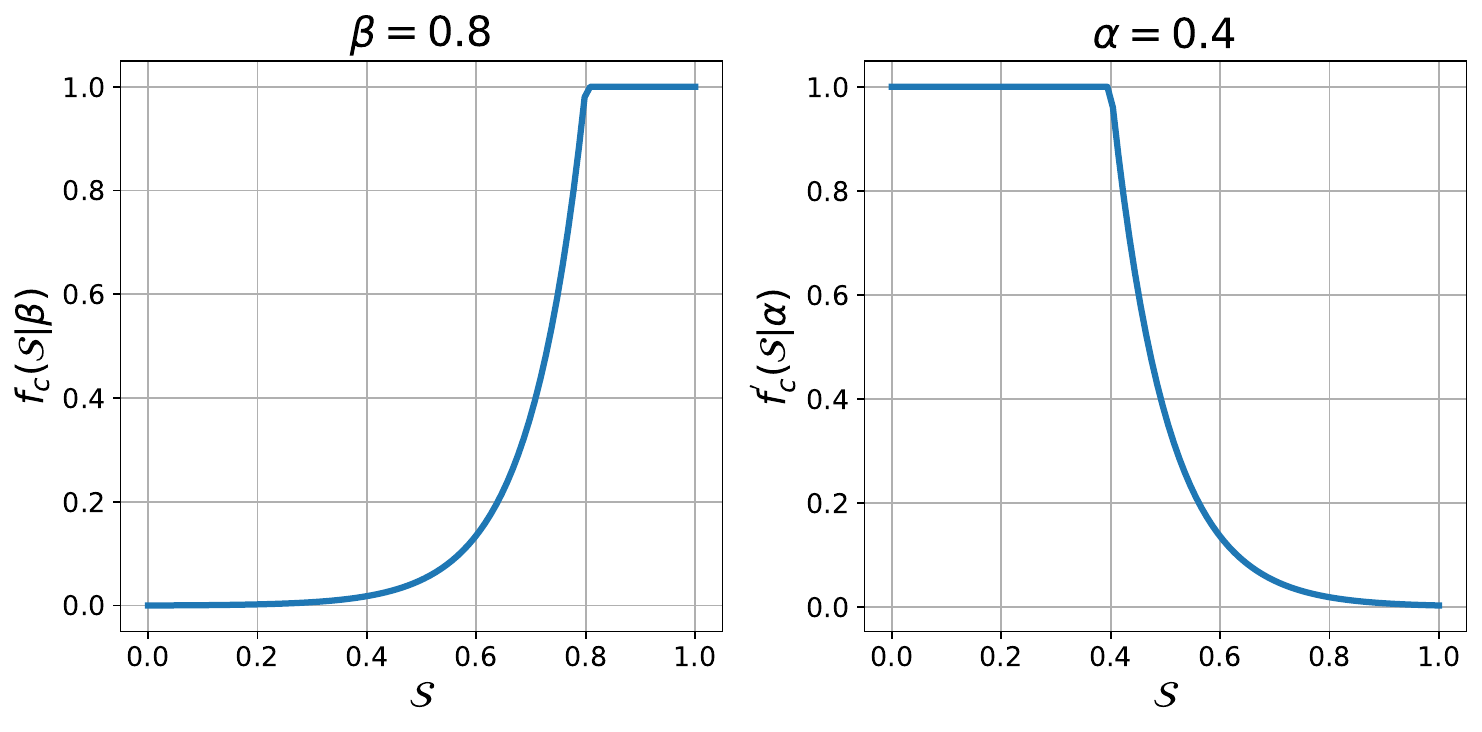}
\caption{Functions $f_{c}(\mathcal{S}|\beta), f'_{c}(\mathcal{S}|\alpha)$ used in the evaluation criteria of single video rating.}
\label{fig:exp_decay_func}
\end{figure*}

\begin{table*}[t]
    \centering
    \begin{minipage}{0.48\linewidth}
        \centering
        \caption{$\mathcal{A}^{single}$ results with varying values of $\beta$ when $\alpha$ is set to 0.4.}
        \begin{tabular}{lccc}
             \toprule
              & \textbf{$\beta$=0.7} & \textbf{$\beta$=0.8} & \textbf{$\beta$=0.9} \\
             \midrule
             Qwen2-VL-2B	& 61.2 & 58.3 & 55.9 \\
             LongVA-DPO-7B & 64.1 & 61.6 & 57.6 \\
             Qwen2-VL-7B	  & 74.4 & 70.9	& 66.3 \\
             Qwen2-VL-72B  & 77.3 & 75.4	& 72.5 \\
             \bottomrule
        \end{tabular}
        \label{tab:effect_alpha_beta_a}
    \end{minipage}
    \hfill
    \begin{minipage}{0.48\linewidth}
        \centering
        \caption{$\mathcal{A}^{single}$ results with varying values of $\alpha$ when $\beta$ is set to 0.8.}
        \begin{tabular}{lccc}
             \toprule
              & \textbf{$\alpha$=0.3} & \textbf{$\alpha$=0.4} & \textbf{$\alpha$=0.5} \\
             \midrule
             Qwen2-VL-2B	& 56.7 & 58.3 &	62.8 \\
             LongVA-DPO-7B & 60.4 & 61.6 & 63.0 \\
             Qwen2-VL-7B	  & 69.6 & 70.9 & 72.5 \\
             Qwen2-VL-72B  & 74.7 & 75.4	& 76.2 \\
             \bottomrule
        \end{tabular}
        \label{tab:effect_alpha_beta_b}
    \end{minipage}
\end{table*}

\begin{table}[t]
    \centering
    \caption{Comparison of positive and negative prompts in terms of $\mathcal{A}^{single}$ with Qwen2-VL-7B as the backbone model.}
    \resizebox{\textwidth}{!}{$
    \begin{tabular}{lccccc}
         \toprule
         \textbf{Prompt} & \textbf{Dynamic Degree}	& \textbf{Static Quality}	& \textbf{Temporal Quality} & \textbf{Video-Text Alignment} & \textbf{AVG} \\
         \midrule
         Positive  & \textbf{84.6} & \textbf{64.6} & 51.1 & \textbf{85.4} & \textbf{70.9}\\
         Negative  & 66.0 & 62.7 & \textbf{55.0} & 16.9 & 50.9\\
         \bottomrule
    \end{tabular}
    $}
    \label{tab:neg_pos_prompt}
\end{table}
\begin{table*}[t]
    \centering
    \caption{Data example of \textit{Subject Motion} and corresponding evaluations by InternVL-2.5-78B-MPO and InternVL-2.5-8B-MPO.}
    \resizebox{0.85\linewidth}{!}{
    \begin{tcolorbox}    
        \textbf{Video 1:}\\
        \includegraphics[width=1.\textwidth]{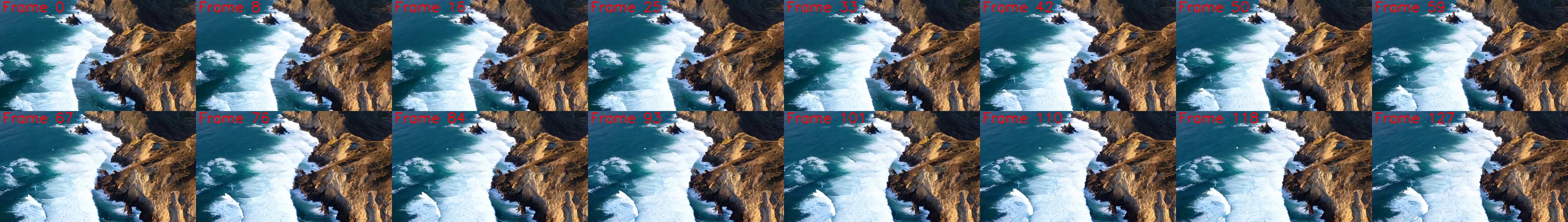} \\
        \textbf{Video 2:}\\
        \includegraphics[width=1.\textwidth]{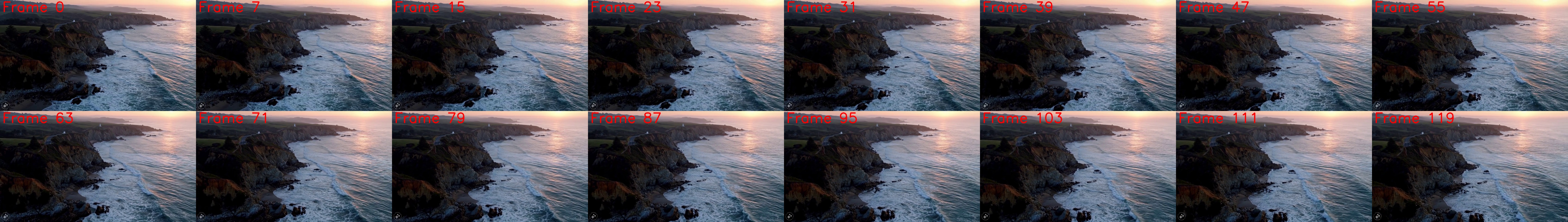} \\
        {\color[HTML]{CC0000} Aspect:} Subject Motion (Dynamic Degree)\\
        {\color[HTML]{00CC00} Human Preference:} same bad\\
        
        \textbf{Single Video Rating}\\
        {\color[HTML]{0000EF} InternVL-2.5-78B} \hspace{2cm} {\color[HTML]{0000EF} InternVL-2.5-8B}\\
        - Video 1: 0.076 \hspace{2.4cm} - Video 1: 0.085\\
        - Video 2: 0.0028 \hspace{2.22cm} - Video 2: 0.047\\
        - $\mathcal{A}^{\text{sinlge}}=1$ \hspace{2.98cm} - $\mathcal{A}^{\text{sinlge}}=1$\\ \\
        
        \textbf{Video Pair Comparison}\\
        {\color[HTML]{0000EF} InternVL-2.5-78B} \hspace{2cm} {\color[HTML]{0000EF} InternVL-2.5-8B} \\
        - Model Preference: same bad \hspace{0.7cm} - Model Preference: Video 2\\
        - Accuracy = \checkmark \hspace{2.88cm}  - Accuracy = \crossmark\\
        
    \end{tcolorbox}
    }
    \label{tab:example_subject_motion}
\end{table*}

\begin{table*}[t]
    \centering
    \caption{Data example of \textit{Camera Motion} and evaluations by InternVL-2.5-78B-MPO and InternVL-2.5-8B-MPO.}
    \resizebox{0.85\linewidth}{!}{
    \begin{tcolorbox}    
        \textbf{Video 1:}\\
        \includegraphics[width=1.\textwidth]{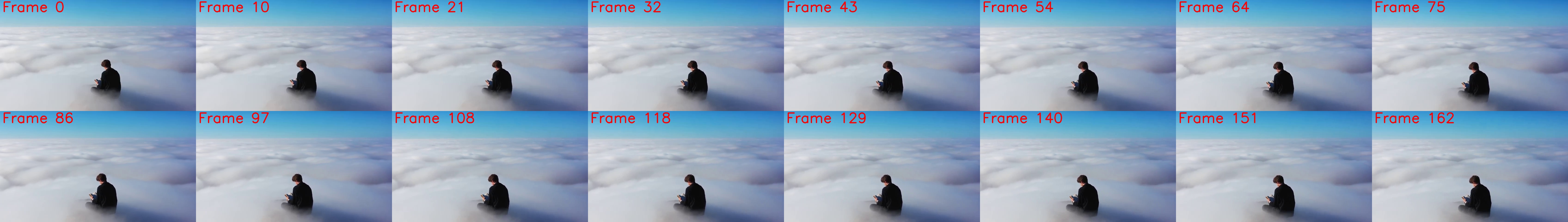} \\
        \textbf{Video 2:}\\
        \includegraphics[width=1.\textwidth]{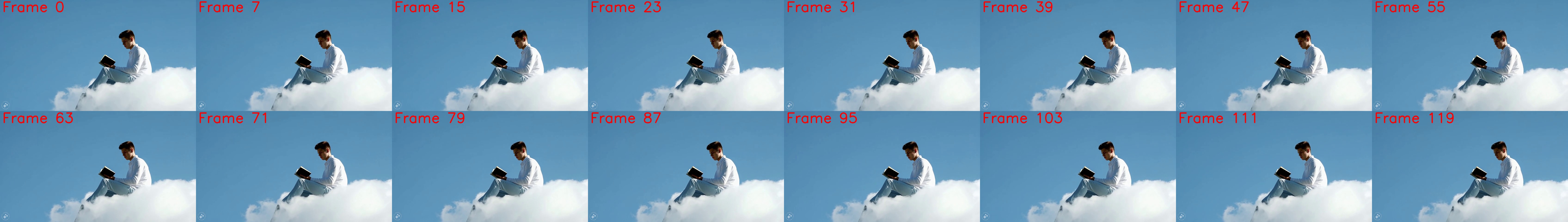} \\
        {\color[HTML]{CC0000} Aspect:} Camera Motion (Dynamic Degree)\\
        {\color[HTML]{00CC00} Human Preference:} Video 1 is better\\
        
        \textbf{Single Video Rating}\\
        {\color[HTML]{0000EF} InternVL-2.5-78B} \hspace{2cm} {\color[HTML]{0000EF} InternVL-2.5-8B}\\
        - Video 1: 1.2$\times$10$^{-6}$ \hspace{2cm} - Video 1: 1.5$\times$10$^{-3}$\\
        - Video 2: 8.8$\times$10$^{-8}$ \hspace{2cm} - Video 2: 3.4$\times$10$^{-4}$\\
        - $\mathcal{A}^{\text{sinlge}}=1$ \hspace{3.22cm} - $\mathcal{A}^{\text{sinlge}}=1$\\ \\
        
        \textbf{Video Pair Comparison}\\
        {\color[HTML]{0000EF} InternVL-2.5-78B} \hspace{2cm} {\color[HTML]{0000EF} InternVL-2.5-8B} \\
        - Model Preference: same bad \hspace{0.7cm} - Model Preference: same bad\\
        - Accuracy = \crossmark \hspace{2.88cm}  - Accuracy = \crossmark\\
    \end{tcolorbox}
    }
    \label{tab:example_camera_motion}
\end{table*}

\begin{table*}[t]
    \centering
    \caption{Data example of \textit{Light Change} and evaluations by InternVL-2.5-78B-MPO and InternVL-2.5-8B-MPO.}
    \resizebox{0.85\linewidth}{!}{
    \begin{tcolorbox}    
        \textbf{Video 1:}\\
        \includegraphics[width=1.\textwidth]{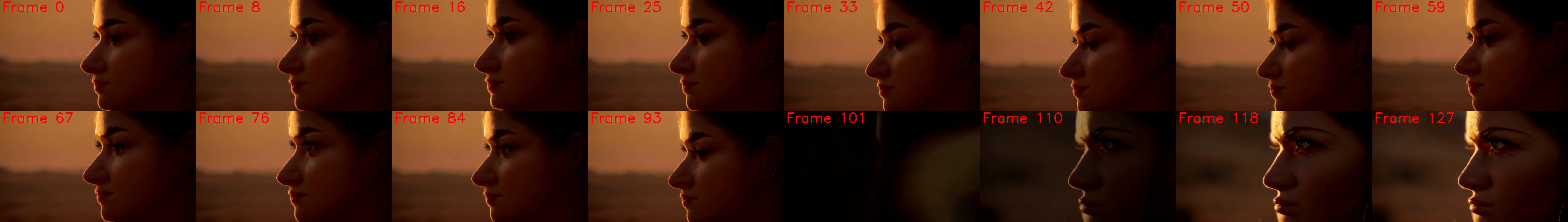} \\
        \textbf{Video 2:}\\
        \includegraphics[width=1.\textwidth]{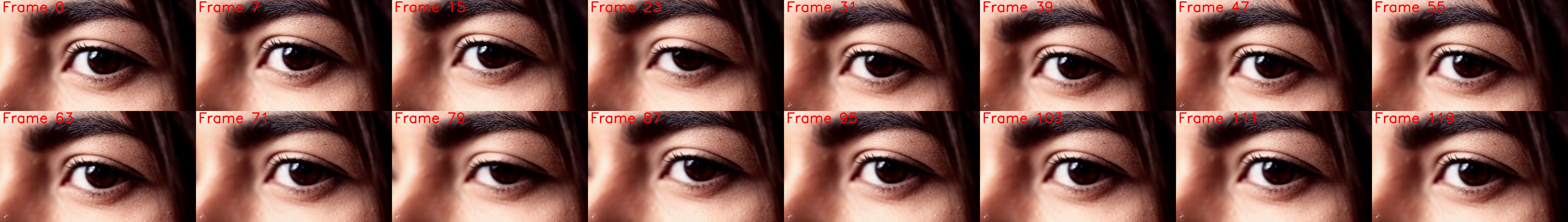} \\
        {\color[HTML]{CC0000} Aspect:} Light Change (Dynamic Degree)\\
        {\color[HTML]{00CC00} Human Preference:} Video 1 is better\\
        
        \textbf{Single Video Rating}\\
        {\color[HTML]{0000EF} InternVL-2.5-78B} \hspace{2cm} {\color[HTML]{0000EF} InternVL-2.5-8B}\\
        - Video 1: 0.531 \hspace{2.4cm} - Video 1: 0.201\\
        - Video 2: 0.004 \hspace{2.4cm} - Video 2: 0.047\\
        - $\mathcal{A}^{\text{sinlge}}=1$ \hspace{2.98cm} - $\mathcal{A}^{\text{sinlge}}=1$\\ \\
        
        \textbf{Video Pair Comparison}\\
        {\color[HTML]{0000EF} InternVL-2.5-78B} \hspace{2cm} {\color[HTML]{0000EF} InternVL-2.5-8B} \\
        - Model Preference: Video 1 \hspace{0.7cm} - Model Preference: Video 1\\
        - Accuracy = \checkmark \hspace{2.6cm}  - Accuracy = \checkmark\\
    \end{tcolorbox}
    }
    \label{tab:example_light_change}
\end{table*}

\begin{table*}[t]
    \centering
    \caption{Data example of \textit{Technical Quality} and evaluations by InternVL-2.5-78B-MPO and InternVL-2.5-8B-MPO.}
    \resizebox{0.85\linewidth}{!}{
    \begin{tcolorbox}    
        \textbf{Video 1:}\\
        \includegraphics[width=1.\textwidth]{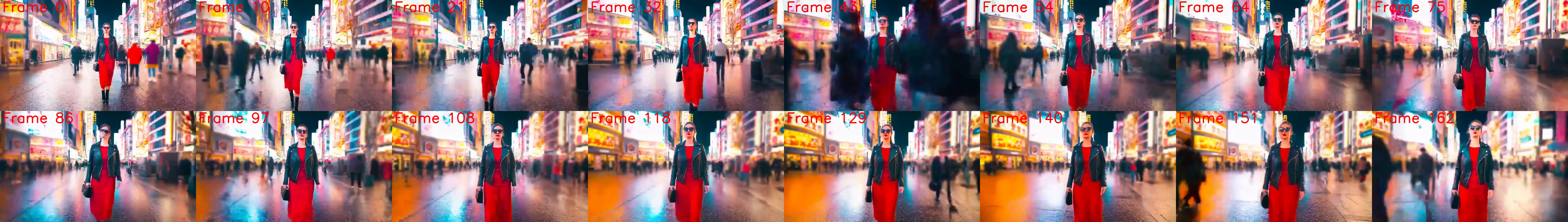} \\
        \textbf{Video 2:}\\
        \includegraphics[width=1.\textwidth]{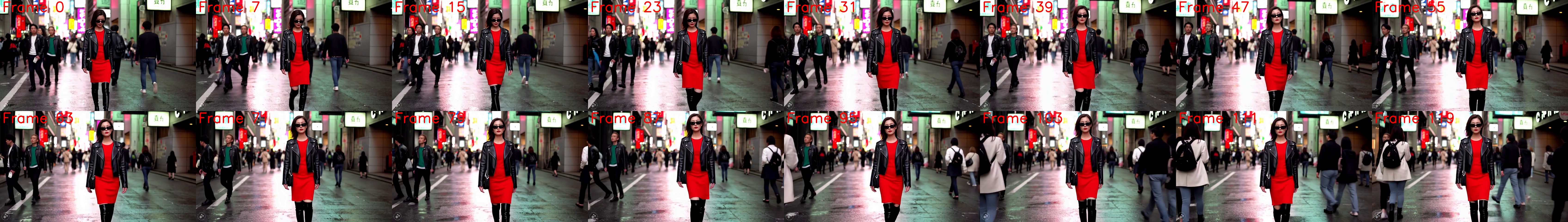} \\
        {\color[HTML]{CC0000} Aspect:} Technical Quality (Static Quality)\\
        {\color[HTML]{00CC00} Human Preference:} Video 2 is better\\
        
        \textbf{Single Video Rating}\\
        {\color[HTML]{0000EF} InternVL-2.5-78B} \hspace{2cm} {\color[HTML]{0000EF} InternVL-2.5-8B}\\
        - Video 1: 0.500 \hspace{2.4cm} - Video 1: 0.651\\
        - Video 2: 0.999 \hspace{2.4cm} - Video 2: 0.940\\
        - $\mathcal{A}^{\text{sinlge}}=1$ \hspace{2.98cm} - $\mathcal{A}^{\text{sinlge}}=1$\\ \\
        
        \textbf{Video Pair Comparison}\\
        {\color[HTML]{0000EF} InternVL-2.5-78B} \hspace{2cm} {\color[HTML]{0000EF} InternVL-2.5-8B} \\
        - Model Preference: Video 2 \hspace{0.7cm} - Model Preference: Video 2\\
        - Accuracy = \checkmark \hspace{2.6cm}  - Accuracy = \checkmark\\
    \end{tcolorbox}
    }
    \label{tab:example_technical_quality}
\end{table*}

\begin{table*}[t]
    \centering
    \caption{Data example of \textit{Aesthetic Quality} and evaluations by InternVL-2.5-78B-MPO and InternVL-2.5-8B-MPO.}
    \resizebox{0.85\linewidth}{!}{
    \begin{tcolorbox}    
        \textbf{Video 1:}\\
        \includegraphics[width=1.\textwidth]{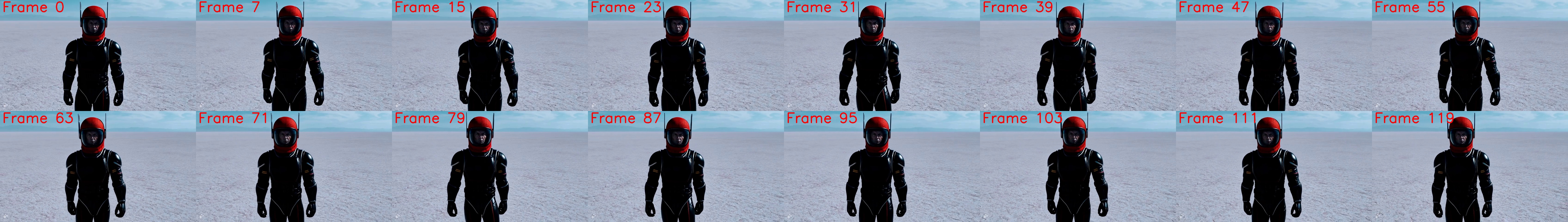} \\
        \textbf{Video 2:}\\
        \includegraphics[width=1.\textwidth]{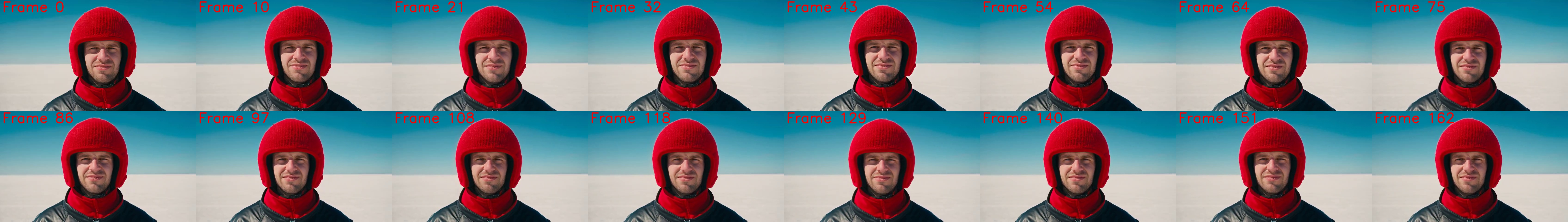} \\
        {\color[HTML]{CC0000} Aspect:} Aesthetic Quality (Static Quality)\\
        {\color[HTML]{00CC00} Human Preference:} same good\\
        
        \textbf{Single Video Rating}\\
        {\color[HTML]{0000EF} InternVL-2.5-78B} \hspace{2cm} {\color[HTML]{0000EF} InternVL-2.5-8B}\\
        - Video 1: 0.986 \hspace{2.4cm} - Video 1: 0.731\\
        - Video 2: 0.991 \hspace{2.4cm} - Video 2: 0.755\\
        - $\mathcal{A}^{\text{sinlge}}=1$ \hspace{2.98cm} - $\mathcal{A}^{\text{sinlge}}=0.319$\\ \\
        
        \textbf{Video Pair Comparison}\\
        {\color[HTML]{0000EF} InternVL-2.5-78B} \hspace{2cm} {\color[HTML]{0000EF} InternVL-2.5-8B} \\
        - Model Preference: same good \hspace{0.24cm} - Model Preference: Video 2\\
        - Accuracy = \checkmark \hspace{2.6cm}  - Accuracy = \crossmark\\
    \end{tcolorbox}
    }
    \label{tab:example_aesthetic_quality}
\end{table*}

\begin{table*}[t]
    \centering
    \caption{Data example of \textit{Structural Correctness} and evaluations by InternVL-2.5-78B-MPO and InternVL-2.5-8B-MPO.}
    \resizebox{0.85\linewidth}{!}{
    \begin{tcolorbox}    
        \textbf{Video 1:}\\
        \includegraphics[width=1.\textwidth]{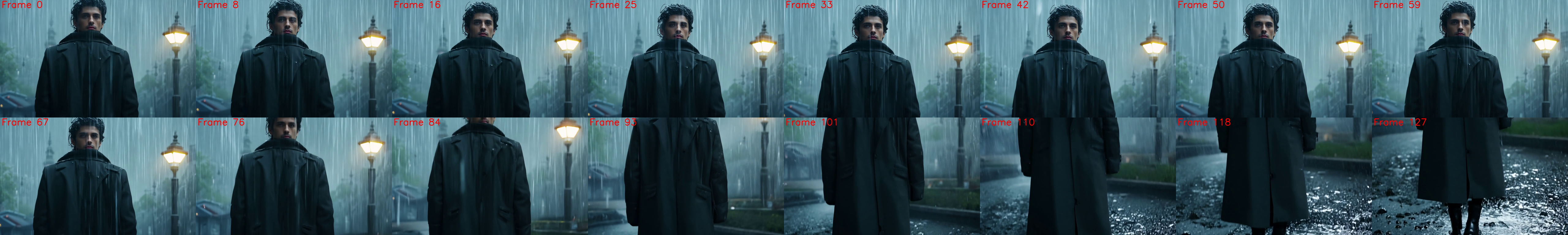} \\
        \textbf{Video 2:}\\
        \includegraphics[width=1.\textwidth]{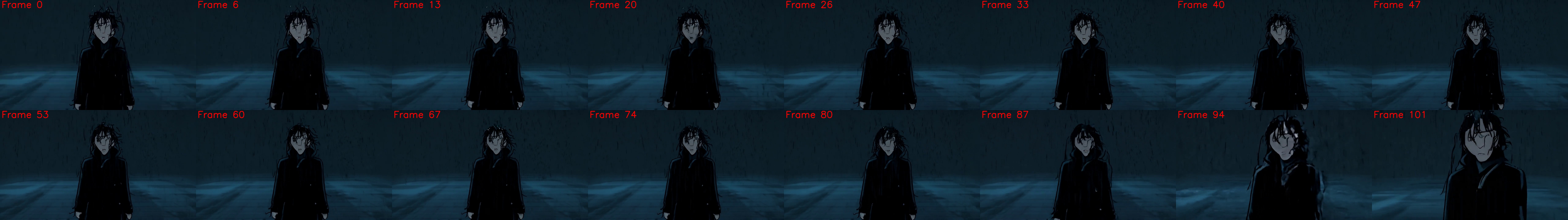} \\
        {\color[HTML]{CC0000} Aspect:} Structural Correctness (Static Quality)\\
        {\color[HTML]{00CC00} Human Preference:} same bad\\
        
        \textbf{Single Video Rating}\\
        {\color[HTML]{0000EF} InternVL-2.5-78B} \hspace{1.8cm} {\color[HTML]{0000EF} InternVL-2.5-8B}\\
        - Video 1: 0.999 \hspace{2.1cm} - Video 1: 0.971\\
        - Video 2: 0.971 \hspace{2.1cm} - Video 2: 0.915\\
        - $\mathcal{A}^{\text{sinlge}}=8.6\times10^{-6}$ \hspace{1.2cm} - $\mathcal{A}^{\text{sinlge}}=1.9\times10^{-5}$\\ \\
        
        \textbf{Video Pair Comparison}\\
        {\color[HTML]{0000EF} InternVL-2.5-78B} \hspace{2cm} {\color[HTML]{0000EF} InternVL-2.5-8B} \\
        - Model Preference: Video 1 \hspace{0.6cm} - Model Preference: Video 1\\
        - Accuracy = \crossmark \hspace{2.6cm}  - Accuracy = \crossmark\\
    \end{tcolorbox}
    }
    \label{tab:example_structural_correctness}
\end{table*}

\begin{table*}[t]
    \centering
    \caption{Data example of \textit{Appearance Consistency} and evaluations by InternVL-2.5-78B-MPO and InternVL-2.5-8B-MPO.}
    \resizebox{0.8\linewidth}{!}{
    \begin{tcolorbox}    
        \textbf{Video 1:}\\
        \includegraphics[width=1.\textwidth]{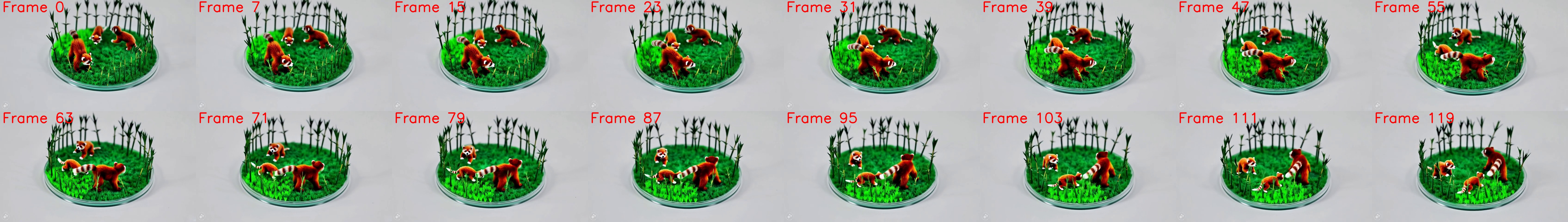} \\
        \textbf{Video 2:}\\
        \includegraphics[width=1.\textwidth]{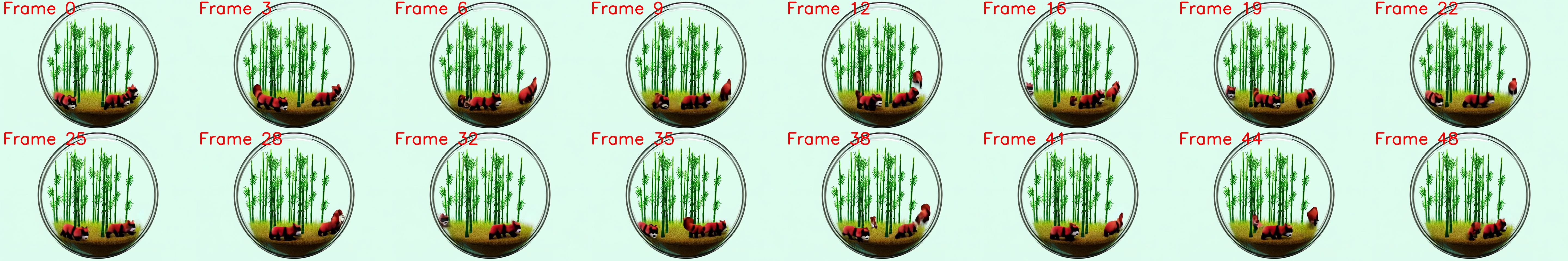} \\
        {\color[HTML]{CC0000} Aspect:} Appearance Consistency (Static Quality)\\
        {\color[HTML]{00CC00} Human Preference:} Video 1 is better\\
        
        \textbf{Single Video Rating}\\
        {\color[HTML]{0000EF} InternVL-2.5-78B} \hspace{1.8cm} {\color[HTML]{0000EF} InternVL-2.5-8B}\\
        - Video 1: 0.990 \hspace{2.1cm} - Video 1: 0.971\\
        - Video 2: 0.893 \hspace{2.1cm} - Video 2: 0.980\\
        - $\mathcal{A}^{\text{sinlge}}=1$ \hspace{2.7cm} - $\mathcal{A}^{\text{sinlge}}=0$\\ \\
        
        \textbf{Video Pair Comparison}\\
        {\color[HTML]{0000EF} InternVL-2.5-78B} \hspace{2cm} {\color[HTML]{0000EF} InternVL-2.5-8B} \\
        - Model Preference: Video 1 \hspace{0.6cm} - Model Preference: Video 2\\
        - Accuracy = \checkmark \hspace{2.5cm}  - Accuracy = \crossmark\\
    \end{tcolorbox}
    }
    \label{tab:example_appearance_consistency}
\end{table*}

\begin{table*}[t]
    \centering
    \caption{Data example of \textit{Flickering} and evaluations by InternVL-2.5-78B-MPO and InternVL-2.5-8B-MPO.}
    \resizebox{0.8\linewidth}{!}{
    \begin{tcolorbox}    
        \textbf{Video 1:}\\
        \includegraphics[width=1.\textwidth]{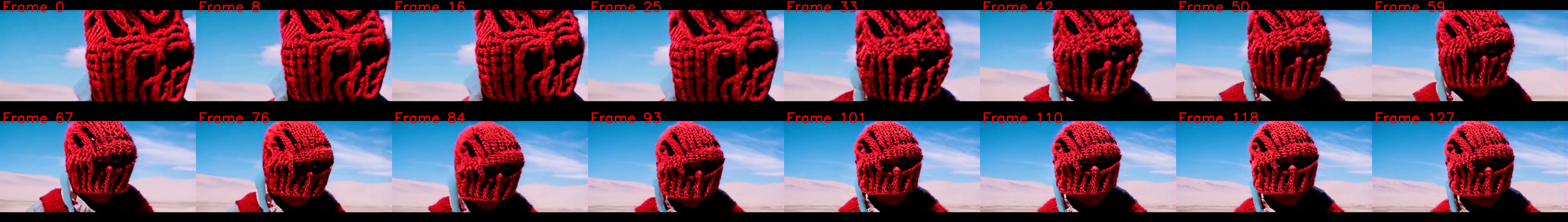} \\
        \textbf{Video 2:}\\
        \includegraphics[width=1.\textwidth]{figures/mochi_00002.jpg} \\
        {\color[HTML]{CC0000} Aspect:} Temporal Flickering (Temporal Quality)\\
        {\color[HTML]{00CC00} Human Preference:} Video 2 is better\\
        
        \textbf{Single Video Rating}\\
        {\color[HTML]{0000EF} InternVL-2.5-78B} \hspace{1.8cm} {\color[HTML]{0000EF} InternVL-2.5-8B}\\
        - Video 1: 0.852 \hspace{2.1cm} - Video 1: 0.731\\
        - Video 2: 0.998 \hspace{2.1cm} - Video 2: 0.905\\
        - $\mathcal{A}^{\text{sinlge}}=1$ \hspace{2.7cm} - $\mathcal{A}^{\text{sinlge}}=1$\\ \\
        
        \textbf{Video Pair Comparison}\\
        {\color[HTML]{0000EF} InternVL-2.5-78B} \hspace{2cm} {\color[HTML]{0000EF} InternVL-2.5-8B} \\
        - Model Preference: Video 2 \hspace{0.6cm} - Model Preference: Video 2\\
        - Accuracy = \checkmark \hspace{2.5cm}  - Accuracy = \checkmark\\
    \end{tcolorbox}
    }
    \label{tab:example_flickering}
\end{table*}

\begin{table*}[t]
    \centering
    \caption{Data example of \textit{Motion Naturalness} and evaluations by InternVL-2.5-78B-MPO and InternVL-2.5-8B-MPO.}
    \resizebox{0.8\linewidth}{!}{
    \begin{tcolorbox}    
        \textbf{Video 1:}\\
        \includegraphics[width=1.\textwidth]{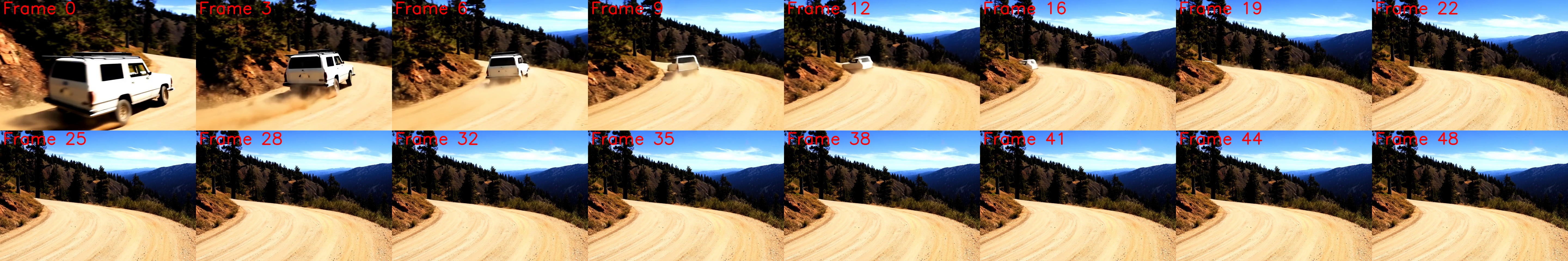} \\
        \textbf{Video 2:}\\
        \includegraphics[width=1.\textwidth]{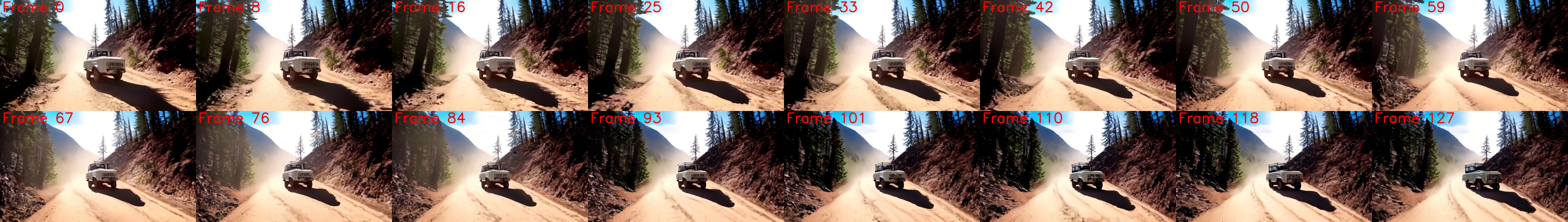} \\
        {\color[HTML]{CC0000} Aspect:} Motion Naturalness (Temporal Quality)\\
        {\color[HTML]{00CC00} Human Preference:} Video 1 is better\\
        
        \textbf{Single Video Rating}\\
        {\color[HTML]{0000EF} InternVL-2.5-78B} \hspace{1.8cm} {\color[HTML]{0000EF} InternVL-2.5-8B}\\
        - Video 1: 0.924 \hspace{2.1cm} - Video 1: 0.731\\
        - Video 2: 0.986 \hspace{2.1cm} - Video 2: 0.798\\
        - $\mathcal{A}^{\text{sinlge}}=0$ \hspace{2.7cm} - $\mathcal{A}^{\text{sinlge}}=0$\\ \\
        
        \textbf{Video Pair Comparison}\\
        {\color[HTML]{0000EF} InternVL-2.5-78B} \hspace{2cm} {\color[HTML]{0000EF} InternVL-2.5-8B} \\
        - Model Preference: Video 2 \hspace{0.6cm} - Model Preference: Video 2\\
        - Accuracy = \crossmark \hspace{2.6cm}  - Accuracy = \crossmark\\
    \end{tcolorbox}
    }
    \label{tab:example_motion_naturalness}
\end{table*}

\begin{table*}[t]
    \centering
    \caption{Data example of \textit{Appearance Fine} and evaluations by InternVL-2.5-78B-MPO and InternVL-2.5-8B-MPO.}
    \resizebox{0.85\linewidth}{!}{
    \begin{tcolorbox}    
        \textbf{Video 1:}\\
        \includegraphics[width=1.\textwidth]{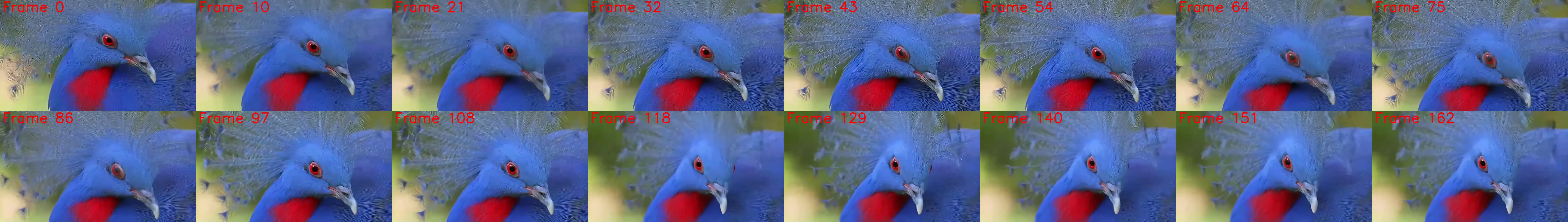} \\
        \textbf{Video 2:}\\
        \includegraphics[width=1.\textwidth]{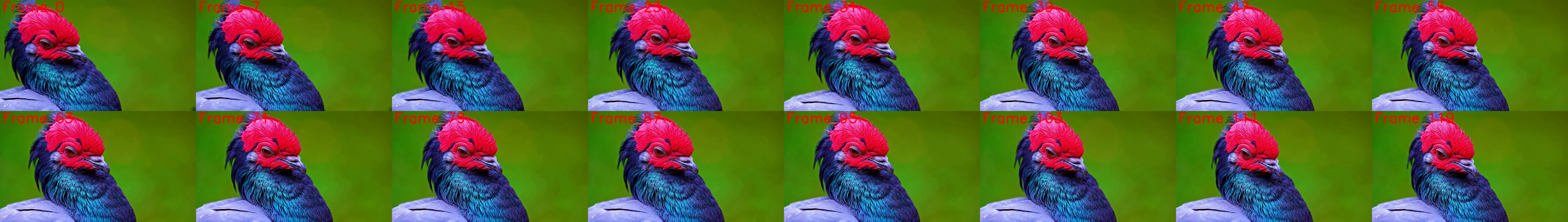} \\
        {\color[HTML]{CC0000} Aspect:} Appearance Alignment (Video-Text Alignment)\\
        {\color[HTML]{00CC00} Human Preference:} Video 1 is better\\
        
        \textbf{Single Video Rating}\\
        {\color[HTML]{0000EF} InternVL-2.5-78B} \hspace{1.8cm} {\color[HTML]{0000EF} InternVL-2.5-8B}\\
        - Video 1: 0.999 \hspace{2.1cm} - Video 1: 0.984\\
        - Video 2: 0.009 \hspace{2.1cm} - Video 2: 0.755\\
        - $\mathcal{A}^{\text{sinlge}}=1$ \hspace{2.7cm} - $\mathcal{A}^{\text{sinlge}}=1$\\ \\
        
        \textbf{Video Pair Comparison}\\
        {\color[HTML]{0000EF} InternVL-2.5-78B} \hspace{2cm} {\color[HTML]{0000EF} InternVL-2.5-8B} \\
        - Model Preference: Video 1 \hspace{0.6cm} - Model Preference: Video 1\\
        - Accuracy = \checkmark \hspace{2.5cm}  - Accuracy = \checkmark\\
    \end{tcolorbox}
    }
    \label{tab:example_appearance_fine}
\end{table*}

\begin{table*}[t]
    \centering
    \caption{Data example of \textit{Motion Fine} and evaluations by InternVL-2.5-78B-MPO and InternVL-2.5-8B-MPO.}
    \resizebox{0.85\linewidth}{!}{
    \begin{tcolorbox}    
        \textbf{Video 1:}\\
        \includegraphics[width=1.\textwidth]{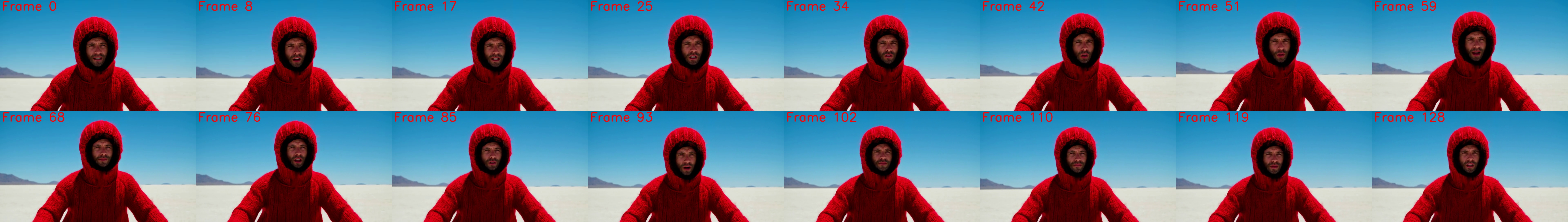} \\
        \textbf{Video 2:}\\
        \includegraphics[width=1.\textwidth]{figures/mochi_00002.jpg} \\
        {\color[HTML]{CC0000} Aspect:} Motion Alignment (Video-Text Alignment)\\
        {\color[HTML]{00CC00} Human Preference:} Video 1 is better\\
        
        \textbf{Single Video Rating}\\
        {\color[HTML]{0000EF} InternVL-2.5-78B} \hspace{1.8cm} {\color[HTML]{0000EF} InternVL-2.5-8B}\\
        - Video 1: 0.269 \hspace{2.1cm} - Video 1: 0.076\\
        - Video 2: 0.245 \hspace{2.1cm} - Video 2: 0.268\\
        - $\mathcal{A}^{\text{sinlge}}=1$ \hspace{2.6cm} - $\mathcal{A}^{\text{sinlge}}=0$\\ \\
        
        \textbf{Video Pair Comparison}\\
        {\color[HTML]{0000EF} InternVL-2.5-78B} \hspace{2cm} {\color[HTML]{0000EF} InternVL-2.5-8B} \\
        - Model Preference: same good \hspace{0.2cm} - Model Preference: Video 2\\
        - Accuracy = \crossmark \hspace{2.6cm}  - Accuracy = \crossmark\\
    \end{tcolorbox}
    }
    \label{tab:example_motion_fine}
\end{table*}

\end{document}